\documentclass[10pt,twocolumn,letterpaper,dvipsnames,table]{article}

\usepackage{iccv}              

\definecolor{lightorange}{RGB}{247,203,153}
\definecolor{lightyellow}{RGB}{255,255,153}
\definecolor{lightred}{RGB}{240,151,153}

\usepackage{tikz}
\usetikzlibrary{tikzmark,calc}

\usepackage{multicol}
\usepackage{algorithm}
\usepackage{algpseudocode}
\usepackage{algorithmicx}
\usepackage{amsmath}
\usepackage{amsthm}
\usepackage{mathtools}
\usepackage{subcaption}
\usepackage{pgfplots}
\usepackage{svg}
\usepackage{comment}
\usepackage{graphicx}
\usepackage{adjustbox}

\usepackage{dblfloatfix}

\usepackage{multirow}
\usepackage{soul} 

\usepackage{wrapfig}
\usepackage{tcolorbox}
\usepackage[accsupp]{axessibility}

\newtheorem{proposition}{Proposition}
\newcommand{\innerproduct}[2]{\left\langle #1, #2 \right\rangle}

\definecolor{lightred}{RGB}{249, 65, 68}
\definecolor{lightorange}{RGB}{248, 150, 30}
\definecolor{lightyellow}{RGB}{249, 199, 79}

\newcommand{\ccr}[0]{\cellcolor{red!10}}
\newcommand{\ccg}[0]{\cellcolor{gray!20}}

\definecolor{lightblue}{RGB}{173, 216, 230}

\newcommand{\lightbluehl}[1]{\sethlcolor{cyan!20}{\hl{#1}}}

\usepackage{pifont}

\newcommand{\cmark}{\textcolor{green}{\ding{51}}}
\newcommand{\xmark}{\textcolor{red}{\ding{55}}}

\tcbset{
    highlightstyle/.style={
        colback=cyan!20, 
        colframe=cyan!20, 
        arc=2pt, 
        boxrule=0pt, 
        boxsep=01pt, 
        left=1pt, right=1pt, top=1pt, bottom=1pt, 
        baseline=0pt, 
    }
}

\definecolor{iccvblue}{rgb}{0.21,0.49,0.74}
\usepackage[pagebackref,breaklinks,colorlinks,allcolors=iccvblue]{hyperref}

\def\paperID{9469} 
\def\confName{ICCV}
\def\confYear{2025}

\title{FedMeNF: Privacy-Preserving Federated Meta-Learning for Neural Fields}

\author{
    Junhyeog Yun \quad Minui Hong \quad Gunhee Kim \\
    Seoul National University \\
    {\tt\small \{junhyeog, alsdml123, gunhee\}@snu.ac.kr} \\
    {\tt\small \href{https://github.com/junhyeog/FedMeNF}{https://github.com/junhyeog/FedMeNF}}
}

\begin{document}
\maketitle
\begin{abstract}
Neural fields provide a memory-efficient representation of data, which can effectively handle diverse modalities and large-scale data.
However, learning to map neural fields often requires large amounts of training data and computations, which can be limited to resource-constrained edge devices.
One approach to tackle this limitation is to leverage Federated Meta-Learning (FML), but traditional FML approaches suffer from privacy leakage.
To address these issues, we introduce a novel FML approach called FedMeNF.
FedMeNF utilizes a new privacy-preserving loss function that regulates privacy leakage in the local meta-optimization. This enables the local meta-learner to optimize quickly and efficiently without retaining the client's private data.
Our experiments demonstrate that FedMeNF achieves fast optimization speed and robust reconstruction performance, even with few-shot or non-IID data across diverse data modalities, while preserving client data privacy.
    
\end{abstract}

\section{Introduction}
\label{sec:intro}

\begin{figure*}[ht]
    \centering
    \begin{minipage}{0.30\textwidth}
        \centering
        \includegraphics[width=\linewidth]{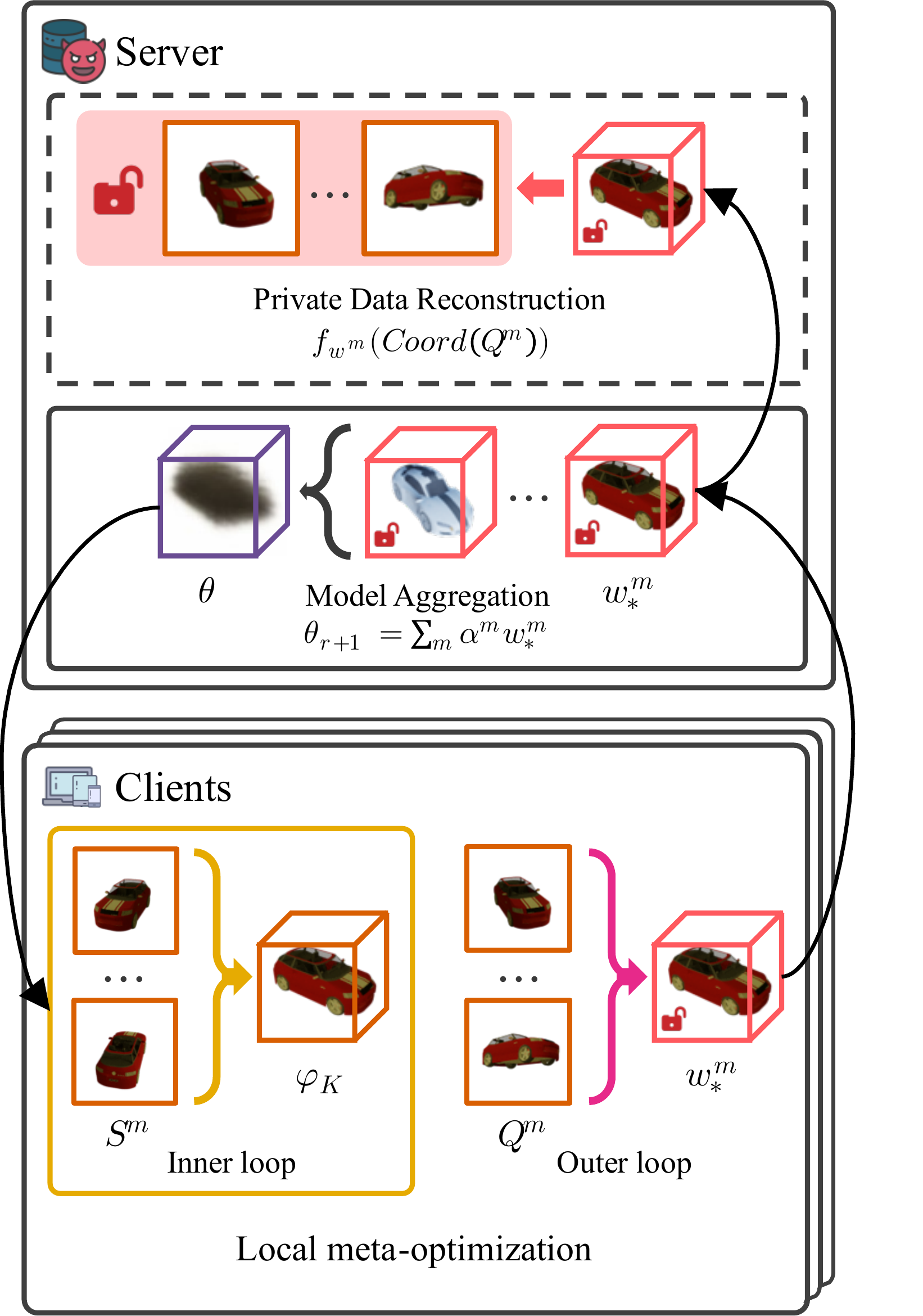}
        \subcaption{FML with a single task per client.}
        \label{fig:main_1}
    \end{minipage}
    \hfill
    \begin{minipage}{0.37\textwidth}
        \centering
        \setlength{\tabcolsep}{1.5pt}
        \small
        \begin{tabular}{lccc}
            \toprule
            \multicolumn{1}{c}{\bf Method} 
            & \multicolumn{1}{c}{\textit{Local}}
            & \multicolumn{1}{c}{FML}
            & \multicolumn{1}{c}{\textbf{Ours}} \\
            \midrule
            Fast optimization   & \xmark  & \cmark  & \cmark  \\
            Few-shot adaptation & \xmark  & \cmark  & \cmark  \\
            Privacy preservation  & \cmark  & \xmark  & \cmark  \\
            \bottomrule
        \end{tabular}
        
        \subcaption{
            Comparison of FedMeNF with \textit{Local} and FML in terms of fast or few-shot optimization on a new task, and privacy preservation.
        }
        \label{tab:compare}
        
        \vspace{1.05em}

        \centering
        \includegraphics[width=0.71\linewidth]{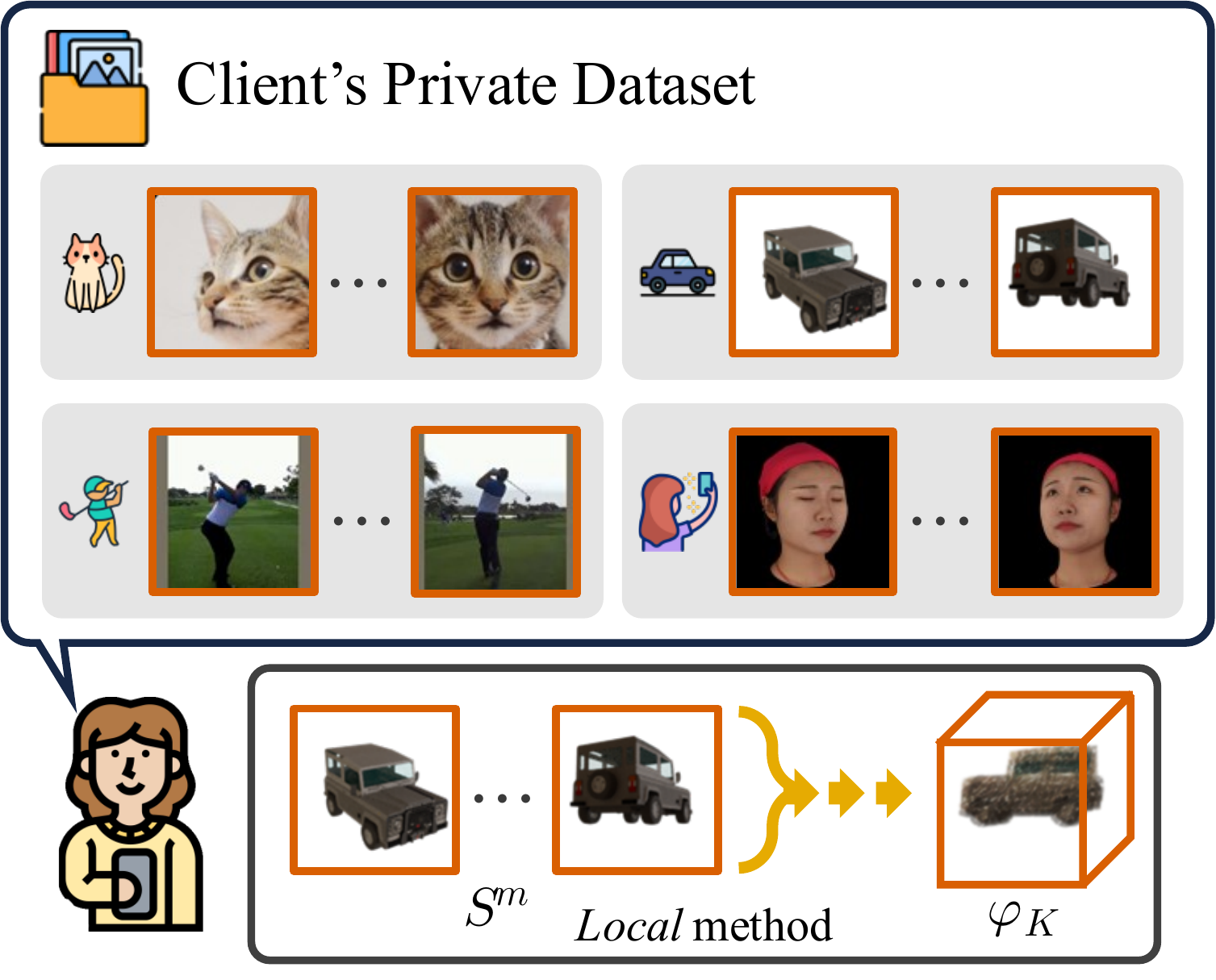}
        \subcaption{
            The client has photos of only one pet, car, body, or face, and aims to optimize a neural field for each instance.
        }
        \label{fig:main_2}
    \end{minipage}
    \hfill
    \begin{minipage}{0.30\textwidth}
        \centering
        \includegraphics[width=\linewidth]{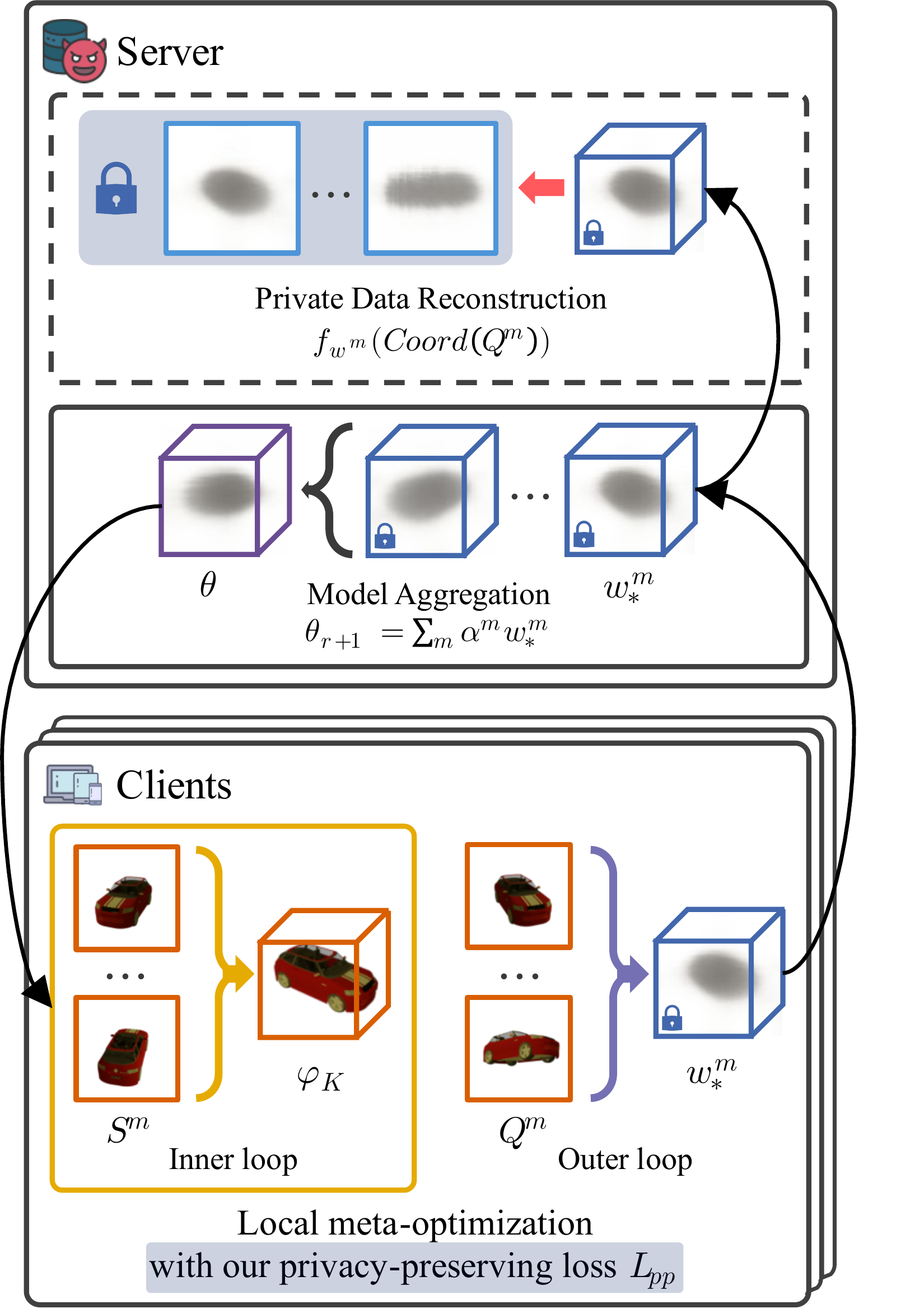}
        \subcaption{FedMeNF with a single task per client.}
        \label{fig:main_3}
    \end{minipage}

    \vspace{2pt}

    \pgfdeclareimage[height=0.6cm]{data}{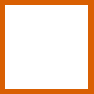}
    \pgfdeclareimage[height=0.3cm]{inner}{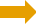}
    \pgfdeclareimage[height=0.3cm]{outer}{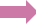}
    \pgfdeclareimage[height=0.7cm]{local}{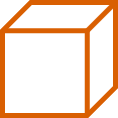}
    \pgfdeclareimage[height=0.7cm]{leaked}{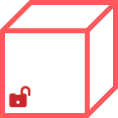}
    \pgfdeclareimage[height=0.7cm]{perserving}{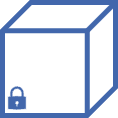}

    \begin{tikzpicture}
    \node[draw, rounded corners, dashed, inner sep=2pt, align=center] (bigbox) at (0,0) {
        \fontsize{8}{11}\selectfont
        \begin{tikzpicture}
            \node at (0,0) {\pgfuseimage{data}};
            \node[right] at (0.4, 0) {Private data};
            
            \node at (2.6, 0) {\pgfuseimage{local}};
            \node[right] at (3, 0) {Local neural fields};
            
            \node at (6.0, 0) {\pgfuseimage{leaked}};
            \node[right] at (6.4, 0) {Privacy-leaked meta-learner};

            \node at (10.5,0) {\pgfuseimage{perserving}};
            \node[right] at (10.9, 0) {Privacy-preserving meta-learner};

        \end{tikzpicture}
    };
    \end{tikzpicture}

    \vspace{-4pt}
        
    \caption{
        \textbf{Overview of our approach.}
        In each communication round, the server sends the global meta-learner $\theta$ to clients, each of which uses its private data to train a local meta-learner $w$ based on the global meta-learner.
        The trained local meta-learners are then sent back to the server to update the global meta-learner.
        (a) In existing federated meta-learning (FML) approaches, the local meta-learner contains privacy-sensitive information when a client has only a single task, as shown in (c). The server can infer the client's private data from the shared local meta-learner.
        (d) In contrast, our FedMeNF ensures that the local meta-learner only learns how to efficiently optimize NFs while preserving privacy-sensitive details.
        (b, c) \textit{Local} refers to using only the client's data without sharing any information with the server.
        }
    \vspace{-5pt}
    \label{fig:motivation}
\end{figure*}
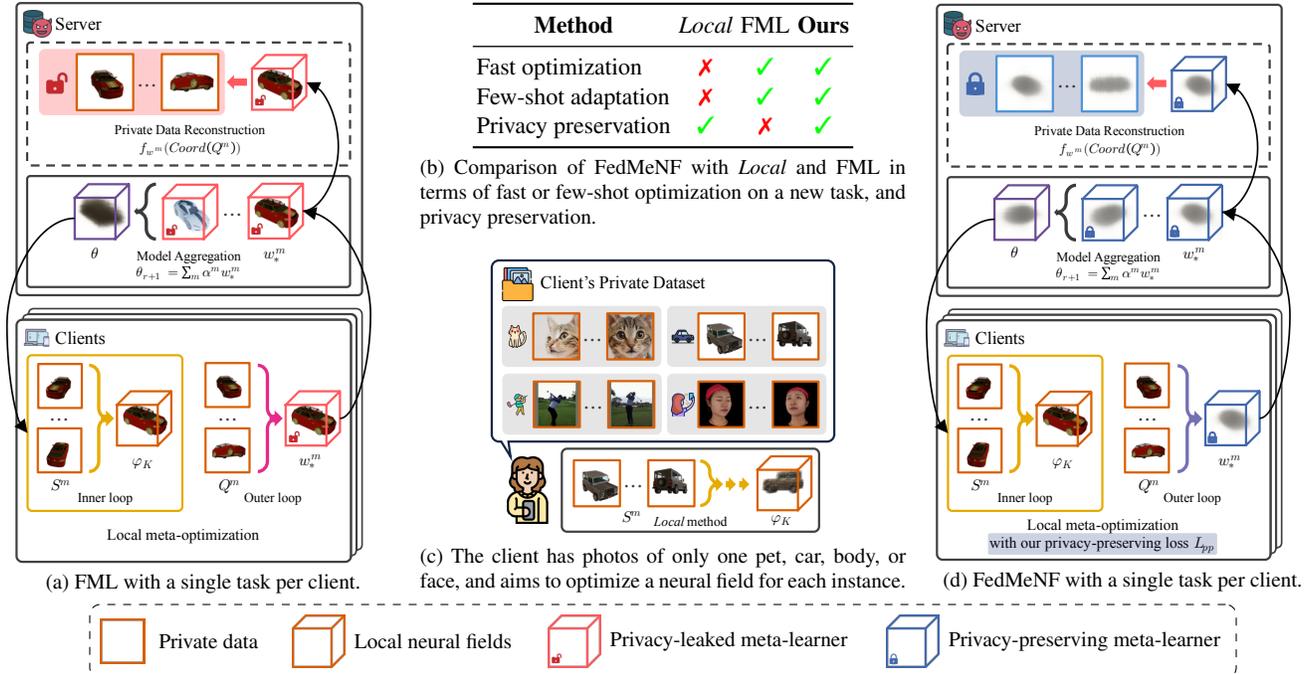

Neural Fields (NFs), also known as Implicit Neural Representations (INRs), use a deep neural network to approximate continuous signals~\cite{park2019deepsdf, tancik2020fourier, saragadam2022miner, tewari2020state, coombes2014nfbook}.
They enable memory-efficient representation and fast processing of data~\cite{dupont2022functa}.
Thanks to this, NFs effectively handle both large-scale data and diverse modalities, including audio, images, videos, and 3D objects~\cite{liang2023avnerf, su2022inras, chen2021learning, skorokhodov2021adversarial, sen2022vinr, chen2022videoinr, kim2022scalable, zhuang2022mofanerf, niemeyer2020differentiable}.

However, optimizing NFs is often slow and requires large amounts of data, which can be a serious issue on resource-constrained edge devices.
Consider a practical scenario: a user takes a few photos of an object using their phone and wants to get a high-fidelity 3D model of it.
Training an NF from scratch with such limited data and resources would be time-consuming and likely yield poor results.
One approach to address this issue is meta-learning~\cite{finn2017maml,onfirst}, which aims to train a meta-learner that can quickly adapt to new tasks with only a few samples.
Using the meta-learner, the user can quickly optimize a high-performing NF for the new object with just a few images.

Nevertheless, meta-learning itself requires a large and diverse set of tasks to learn an effective initialization or optimization strategy.
Unfortunately, in many practical scenarios, users only have data for one or a few tasks.
For example, most users typically take photos of only a single instance of privacy-sensitive objects such as their face, body, pet, or car.
To enable meta-learning in such scenarios, we can leverage Federated Meta-Learning (FML)~\cite{fedmeta, fallah202perfedavg, jiang2019perfed}, which collaboratively trains a global meta-learner using private data from numerous distributed clients.
FML builds upon Federated Learning (FL)~\cite{fedavg, imteaj2021fedsurveyiot, duan2024openfedsurvey, li2022fedsurvey, aledhari2020fedsurveyaccess}, a paradigm designed to train a global model while ensuring client data remains on local devices to preserve privacy.

Yet, applying FML to NFs introduces a critical challenge that fundamentally undermines the privacy guarantee of FL.
The privacy protection in traditional FL relies on the assumption that sharing trained model parameters, rather than the raw data itself, is safe.
This assumption completely breaks down for NFs for two main reasons.
First, in our scenario, each client has data for only a single or a few task instances (\textit{e.g.}, a single image or 3D object).
If even one client has only a single task instance, the local meta-optimization is like standard supervised training on that specific task~\cite{huang2022provable}.
Consequently, the trained local meta-learner functions as an optimized NF for that private data.
Second, NFs are, in essence, a compressed representation of the data and inherently encapsulate it.
Thus, even if only the NF parameters are shared, a malicious server can reconstruct the private data.
Due to this inherent privacy leakage issue, existing studies often limit FL applications for NFs to public data such as landmarks~\cite{tasneem2024decentnerfs} or cityscapes~\cite{holden2023fednerf, zhang2024fednerf2, suzuki2024fed3dgs}, where privacy is less of a concern.
For these reasons, naively applying FML methods to private data results in severe privacy issues, as the server can reconstruct the private data from the shared local meta-learner.

To address this critical challenge, we propose FedMeNF (\textbf{Fed}erated \textbf{Me}ta-Learning for \textbf{N}eural \textbf{F}ields), a novel FML framework designed to collaboratively train a global NF meta-learner while preserving privacy as overviewed in \cref{fig:motivation}, which then empowers each client to rapidly optimize a high-performance NF even with a small amount of new private data.
FedMeNF can be utilized in various privacy-sensitive applications.
For example, suppose that a user buys a new car and takes a few pictures of it. We can then train the NF of the car with this small set of images using a global meta-learner trained with other car images in other devices.
As another example, with only a handful of facial images, users can employ the global facial NF meta-learner to generate images showing a range of facial expressions.
Also, they can upscale low-resolution photos to high-resolution or generate higher frame-rate videos from personal data, such as pet photos or golf swing videos. 

To quantify and control privacy leakage, we define a privacy metric, $\text{PSNR}_p$.
We theoretically analyze how $\text{PSNR}_p$ increases under existing FML approaches.
By regularizing the increase of $\text{PSNR}_p$, FedMeNF forces the local meta-learner to learn only how to optimize the NF faster with limited data without memorizing the private data.
We demonstrate that FedMeNF achieves performance comparable to state-of-the-art methods while better preserving privacy across diverse data modalities, including images, videos, and neural radiance fields (NeRFs).
Additionally, FedMeNF shows robust performance even when local data are very few or highly diverse among clients.

Our contributions can be outlined as follows:

\begin{enumerate}

    \item This is the first study on FL for NFs on private data.
    \item
    We theoretically and empirically show how privacy leakage occurs during the FML for NFs.
    \item
    We propose FedMeNF that preserves the privacy of local data with minimal impact on optimization speed and reconstruction quality.
    \item
    We conduct comprehensive experiments on FedMeNF across various data modalities, private data sizes, and levels of data diversity, outperforming baseline methods.
\end{enumerate}

\section{Preliminaries}

\noindent \textbf{Coordinate-based Neural Fields.}
Coordinate-based neural fields~\cite{mildenhall2021nerf, sitzmann2020siren, tancik2020fourier,chen2022transinr,dupont2022functa,agaram2023canonical,tack2024gradncp} represent continuous functions that map spatial coordinates to signal values such as color or density. The objective for each client $m$ is to find the optimal NF parameter $\varphi_i^m$ that best represents its $i$-th local task data $T_i^m$. This is formulated as
\begin{equation}
    \varphi_i^m = \arg \min_{\varphi} \sum_{(\mathbf{x}_j, \mathbf{y}_j) \in T_i^m} L\left( f_{\varphi}(\mathbf{x}_j), \mathbf{y}_j \right),
    \label{eq:coordinate_neural_field_objective}
\end{equation}
where $f_{\varphi}$ is the forward map parameterized by $\varphi$, $\mathbf{x}_j$ represents the input coordinates (\textit{e.g.}, spatial positions), and $\mathbf{y}_j$ denotes the target values (\textit{e.g.}, pixel colors). $L$ is a suitable loss function such as mean squared error (MSE).

\noindent \textbf{Federated Learning.}
We consider a centralized federated learning (FL) environment \cite{fedavg,abdulrahman2020centralfedsurvey}, where clients do not share their data with either the server or other clients and communicate the model parameters exclusively with the centralized server.
The centralized FL  aims to optimize the global model that minimizes a given objective function as 
\begin{equation}
    \arg \min_{\theta} \sum_{m=0}^{M-1} \alpha^m L(\theta, D^m), \text{ where }
    \alpha^m = \frac{|D^m|}{\Sigma_i |D^i|}.
    \label{eq:fedavg_objevtive}
\end{equation}
$\theta$ is the parameter of the global meta-learner and $M$ is the number of clients. $D^m$ is the local dataset of client $m$. $\alpha^m$ is the proportion of the number of data samples of client $m$.
$L(\theta, D^m)$ represents the loss of model $\theta$ on a dataset $D^m$.
This objective ensures that the global model $\theta$ is optimized across all clients by weighting the contribution of each client based on its local dataset size.

\noindent \textbf{Federated Meta-learning.}
Federated meta-learning~\cite{fedmeta, fallah202perfedavg, jiang2019perfed, chen2022pflbench} is a variant of FL where the client trains the local model using a meta-learning approach with the objective:
\begin{align}
        &\arg \min_{\theta} \sum_{m=0}^{M-1} \alpha^m L(\varphi^m, Q^m), \\
        &\text{where} \ \varphi^m = \arg \min_\theta L(\theta, S^m).
        \label{eq:fedmeta_objective}
\end{align}
$S^m$ and $Q^m$ are the support set and query set of the client $m$.
Each client $m$ first trains the local model $\varphi^m$ on the support set $S^m$ (inner loop).
Then, starting from the global model $\theta$, the local meta-learner minimizes the loss of $\varphi^m$ on the query set $Q^m$ (outer loop).
This local meta-optimization enables the local meta-learner to adapt quickly to new tasks.
The trained local meta-learner is shared with the server, which aggregates the local meta-learners received from participating clients to update the global meta-learner.

\noindent \textbf{Differential Privacy.} 
Differential privacy (DP) is one of the most popular privacy frameworks with rich theoretical guarantees~\cite{abadi2016deepdp,wei2020federateddp,jeffrey2020dpmeta,bun2016concentrated,dwork2016concentrated,hayes2023bounding}.
It guarantees that when a single entry in the database changes, the change in the probability distribution of the algorithm outputs is bounded by predetermined factors $\epsilon$ and $\delta$~\cite{dwork2006dp}.
The privacy risk can be regulated by controlling these factors.
However, traditional DP methods are designed for large datasets; yet neural fields often capture just a single signal.
This causes ambiguity in defining privacy leakage for individual training samples, as well as a significant trade-off between privacy preservation and reconstruction performance.

\section{Problem Statement}
In this work, we assume a scenario where multiple users collaboratively train a global neural field meta-learner using federated meta-learning.
Each user has data about a personal object stored locally on their device.
For instance, each user holds a few images of his or her own car on their device.
Our goal is to optimize neural fields despite limited data and low computational resources on local devices.
Thus, optimizing a specific neural field becomes a local task.
The local task data are divided into a support set and a query set, each containing images of the object.
During training, the support set is used to train the neural field, while the query set is utilized to optimize the local meta-learner.
In the testing phase, the support set enables test-time optimization, and the query set is used to assess its performance.
The detailed FML process is described in \cref{fig:main_1} and \cref{alg:algo}, excluding the \lightbluehl{highlighted parts} which will be discussed in \cref{sec:approach}.

\subsection{First-order Approximation of Meta-gradients}

During the local meta-optimization, the local meta-learner $w$ is updated for each outer step $i$, where task data $T_i = \{ S_i, Q_i \}$ are sampled from the client's training dataset $D_{\text{train}}$.
Then, the local NF $\varphi$, initialized by $w_i$, are optimized with the support set $S_i$ (\cref{eq:inner_step}), and the local meta-learner $w_i$ are optimized with the query set $Q_i$ (\cref{eq:outer_step}).

\begin{proposition}
    \label{prop:g_maml}
    Let
        $g_{M} = \nabla_{w_i}L(\varphi_K, B_K)$ and
        $g_k = \nabla_{\varphi_0}L(\varphi_0, B_k)$.
    Then, the first-order approximation of the meta-gradient $g_{M}$ becomes
    \begin{align}
        \label{eq:g_maml}
        g_{M} \approx \ g_K - \lambda_{i} \mathcal{I}_{K}, \ 
        \text{where} \ \mathcal{I}_{K} = \sum_{k=0}^{K-1} \nabla_{\varphi_0}\innerproduct{g_K}{g_k}.
    \end{align}
\end{proposition}
$K$ is the \# of inner steps, $B_{0...K-1}$ are minibatches of $S_i$, $B_K$ is a minibatch of $Q_i$, and $\lambda_{i}$ is an inner loop learning rate.
The proof is provided in \cref{sec:proof_prop_1}.

\subsection{\texorpdfstring{Fast Optimization via $\mathcal{I}_K$}{}}
\label{sec:fast_optim_inner_product}
According to \cref{prop:g_maml}, the meta-gradient $g_{M}$ can be expressed with $g_K$ and the inner product term $\mathcal{I}_{K}$, which is the sum of inner products between $g_K$ and $g_0,\cdots, g_{K-1}$. This term maximizes (i) the alignment of the gradients from the query set ($g_K$) with those from the support set ($g_0,\cdots, g_{K-1}$), and (ii) the alignment of the gradients computed on different minibatches within the support set.
This enables the local meta-learner (i) to be more generalized and (ii) to quickly optimize the neural field for a new task.

\section{Privacy Leakage in Local Meta-Learning}
\label{sec:privacy_metric}

We define $\text{PSNR}_p$, an intuitive privacy metric to quantify the privacy leakage in FML for neural fields:
\begin{equation}
    \label{eq:def_privacy_metric}
    \text{PSNR}_p = \text{PSNR}(Q^m, f_w(\text{Coord}(Q^m))),
\end{equation}
\noindent where $Q^m$ is the query set of client $m$, $\text{Coord}(Q^m)$ is the coordinates of $Q^m$, and $f_w$ is the forward map parameterized by the local meta-learner $w$.
$\text{PSNR}_p$ assesses how well the server can reconstruct $Q^m$ with the shared local meta-learner $w$.
It is a simple variant of PSNR that measures similarity between images reconstructed with $w$ on the server and the client's local images.
A better reconstruction indicates a greater violation of the client's data privacy, meaning that a larger $\text{PSNR}_p$ indicates more privacy leakage.

We define our main privacy metric based on PSNR for two reasons.
First, neural fields can handle various types of signals such as vision, speech, or sensor data, so does the PSNR.
Second, PSNR is one of the most popular metrics for evaluating reconstruction attacks in FL~\cite{gao2021atsprivacy, yue2023gradient, hong2024fedavp}, as well as reconstruction tasks~\cite{chen2022transinr,kim2023ipc}.
As variants, we also define SSIM$_{p}$ and LPIPS$_{p}$ as privacy metrics to assess structural and perceptual similarities \cite{wang2004ssim,zhang2018lpips}; it can overcome the limitation of PSNR$_{p}$ that is based on only pixel-wise MSE. Please refer to \cref{sec:additional_metrics} for more details.

\subsection{Privacy Leakage via $g_K$}

Since $\text{PSNR}_p$ is defined by $10 \log_{10} (R / L(w, Q^m))$ where $R$ is a constant, it becomes larger as the MSE loss $L(w, Q^m)$ is smaller.
We analyze how $L(w, Q^m)$ changes during the client's local meta-optimization, which helps understand how $\text{PSNR}_p$ evolves.

\begin{proposition}
    \label{prop:privacy_metric}
    Let
        $\Delta L_{i+1} = L(w_{i+1}, B_K) - L(w_i, B_K)$.
    The first-order approximation of $\Delta L_{i+1}$ is then 
    \begin{align}
        \label{eq:privacy_metric}
        \Delta L_{i+1} \approx - \lambda_{o} \cdot {g_K}^2 \leq 0.
    \end{align}
\end{proposition}
The proof is provided in \cref{sec:proof_prop_2}. Note that $Q^m$ is replaced by $B_K$, a minibatch of the $Q^m$.
$\Delta L_{i+1}$ denotes the change in the loss of the local meta-learner $w$ on the client's query set.
According to \cref{prop:privacy_metric}, $\Delta L_{i+1}$ is always negative, which means that $L(w, Q^m)$ decreases with each outer loop iteration.
Since $\text{PSNR}_p$ is inversely proportional to $L(w, Q^m)$, $\text{PSNR}_p$ increases every outer loop iteration.
Consequently, to prevent the increase of $\text{PSNR}_p$, we regularize $\Delta L_{i+1}$  to be close to zero, thereby mitigating the privacy leakage in the local meta-optimization.

\section{Approach}
\label{sec:approach}

We reveal that federated meta-learning (FML) approaches suffer from privacy leakage as their local meta-gradients increase the privacy metric.
To address this issue, we propose FedMeNF, a novel FML framework for neural fields that ensures privacy-preserving local meta-optimization.
\mbox{FedMeNF} employs a privacy-preserving loss function $L_{pp}$, designed to regulate the increase in the privacy metric.
Please refer to \cref{alg:algo,fig:main_3} for its workflow.

\begin{algorithm}[t]
    \caption{FedMeNF (see \cref{sec:notation} for notation)}
    \label{alg:algo}
    
    \begin{algorithmic}[0]    \State \textbf{Input:} (1) $R$: \# of communication rounds, (2) $N$: \# of clients, (3) $M$: \# of participants, (4) $\lambda_{i}$: inner loop learning rate, (5) $\lambda_{o}$: outer loop learning rate, (6) $\alpha^m$: weight proportional to the client $m$'s dataset size (see \cref{eq:fedavg_objevtive}).

    \vspace{3pt}
    
    \State Initialize the global meta-learner $\theta_{0}$

    \For {each communication round $r=0,...,R-1$}        \State Sample $M$ clients from $1,...,N$ clients
        \For{$m=1,...,M$} 
            \State $w_*^m \leftarrow$ \textsc{LocalUpdate}($\theta_r$)
        \EndFor
        \State $\theta_{r+1} \leftarrow  \Sigma_m \alpha^m w_*^m$ \Comment{Server aggregation}
    \EndFor
    \end{algorithmic}
    
    \begin{algorithmic}[0]    \Procedure{LocalUpdate}{$\theta$}

    \vspace{-1.1em}
    \begin{tikzpicture}[remember picture, overlay]
        \draw[draw=cyan!20, rounded corners=2pt, fill=cyan!20]
            ([xshift=-0.1em,yshift=0.8em]$(pic cs:a1)$)
            rectangle
            ([xshift=0.1em,yshift=-0.3em]$(pic cs:a2)$);
    \end{tikzpicture}

    \State Initialize the local meta-learner $w_0 \leftarrow \theta$
    
    \For{each step $i = 0, ..., E-1$} \Comment{Outer loop}
        \State Sample task data $T_i = \{S_i, Q_i\}$ from $D_{\text{train}}$
        \State $\varphi_0 \leftarrow w_i$ \Comment{Initialize the neural field for $T_i$}
        
        \For{each  step $k = 0,...,K-1$} \Comment{Inner loop}
            \State Sample batch $B_k$ from the support set $S_i$
            \vspace{-10pt}
            \State \begin{equation}
                \varphi_{k+1} \leftarrow \varphi_k - \lambda_{i} \nabla_{\varphi_k} L(\varphi_k, B_k)
                \label{eq:inner_step}
                \vspace{-3pt}
            \end{equation}
        \EndFor

        \State Sample batch $B_K$ from the query set $Q_i$
        \State \tikzmark{a1}Select the regularization coefficient $\gamma$\tikzmark{a2}
        
        \vspace{-1.1em}
        \begin{tikzpicture}[remember picture, overlay]
            \draw[draw=cyan!20, rounded corners=2pt, fill=cyan!20]
                ([xshift=-0.0em,yshift=0.8em]$(pic cs:b1)$)
                rectangle
                ([xshift=0.0em,yshift=-0.3em]$(pic cs:b2)$);
        \end{tikzpicture}
    
        \State $w_{i+1} \hspace{-1pt}\leftarrow \hspace{-1pt} w_{i} - \lambda_{o} \nabla_{w_i}(L(\varphi_K, B_K) \tikzmark{b1}- \gamma L(w_i, B_K)\tikzmark{b2})$
        \vspace{-20pt}
        \State \begin{equation}
            \hspace{115pt} 
                \label{eq:outer_step}
        \end{equation} \vspace{-25pt}
    \EndFor

    \State Send $w_E$
    \EndProcedure
    
    \end{algorithmic}
\end{algorithm}

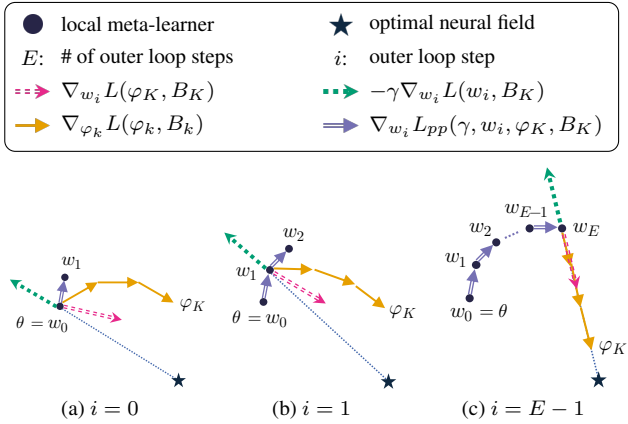
\begin{figure}[t]
    \centering

    \pgfdeclareimage[height=0.4cm]{inner}{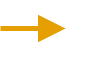}
    \pgfdeclareimage[height=0.4cm]{outer}{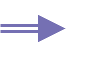}
    \pgfdeclareimage[height=0.4cm]{neg}{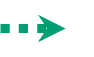}
    \pgfdeclareimage[height=0.4cm]{pos}{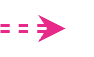}
    \pgfdeclareimage[height=0.25cm]{params}{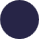}
    \pgfdeclareimage[height=0.3cm]{optimal}{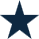}

    \begin{tikzpicture}
            
        \node[draw, rounded corners, solid, inner sep=2pt, align=center] (bigbox) at (0,0) {
            \fontsize{8}{11}\selectfont
            \begin{tikzpicture}

            \node at (-0.1, 0.0) {\pgfuseimage{params}};
            \node[right] at (0.2, 0.0) {local meta-learner};
            \node at (4.0, 0.0) {\pgfuseimage{optimal}};
            \node[right] at (4.3, 0.0) {optimal neural field};
    
            \node at (-0.1, -0.45) {$E$:};
            \node[right] at (0.2, -0.45)  {\# of outer loop steps};
            \node at (4.0, -0.45) {$i$:};
            \node[right] at (4.3, -0.45)  {outer loop step};
            
            \node at (0.0, -0.9) {\pgfuseimage{pos}};
            \node[right] at (0.2, -0.9) {$\nabla_{w_i} L(\varphi_K, B_K)$};            \node at (4.10, -0.9) {\pgfuseimage{neg}};
            \node[right] at (4.3,-0.9) {$- \gamma \nabla_{w_i} L(w_i, B_K)$};            
            \node at (0.0, -1.35) {\pgfuseimage{inner}};
            \node[right] at (0.2, -1.35) {$\nabla_{\varphi_k} L(\varphi_k, B_k)$};            \node at (4.10, -1.35) {\pgfuseimage{outer}};
            \node[right] at (4.3, -1.35) {$\nabla_{w_i} L_{pp}(\gamma, w_i, \varphi_K, B_K)$};

        \end{tikzpicture}
        };
    \end{tikzpicture}
    
    \begin{subfigure}[b]{0.325\linewidth}
        \centering
        \includegraphics[trim={0cm 0cm 0cm 0cm}, clip, width=1.1\linewidth]{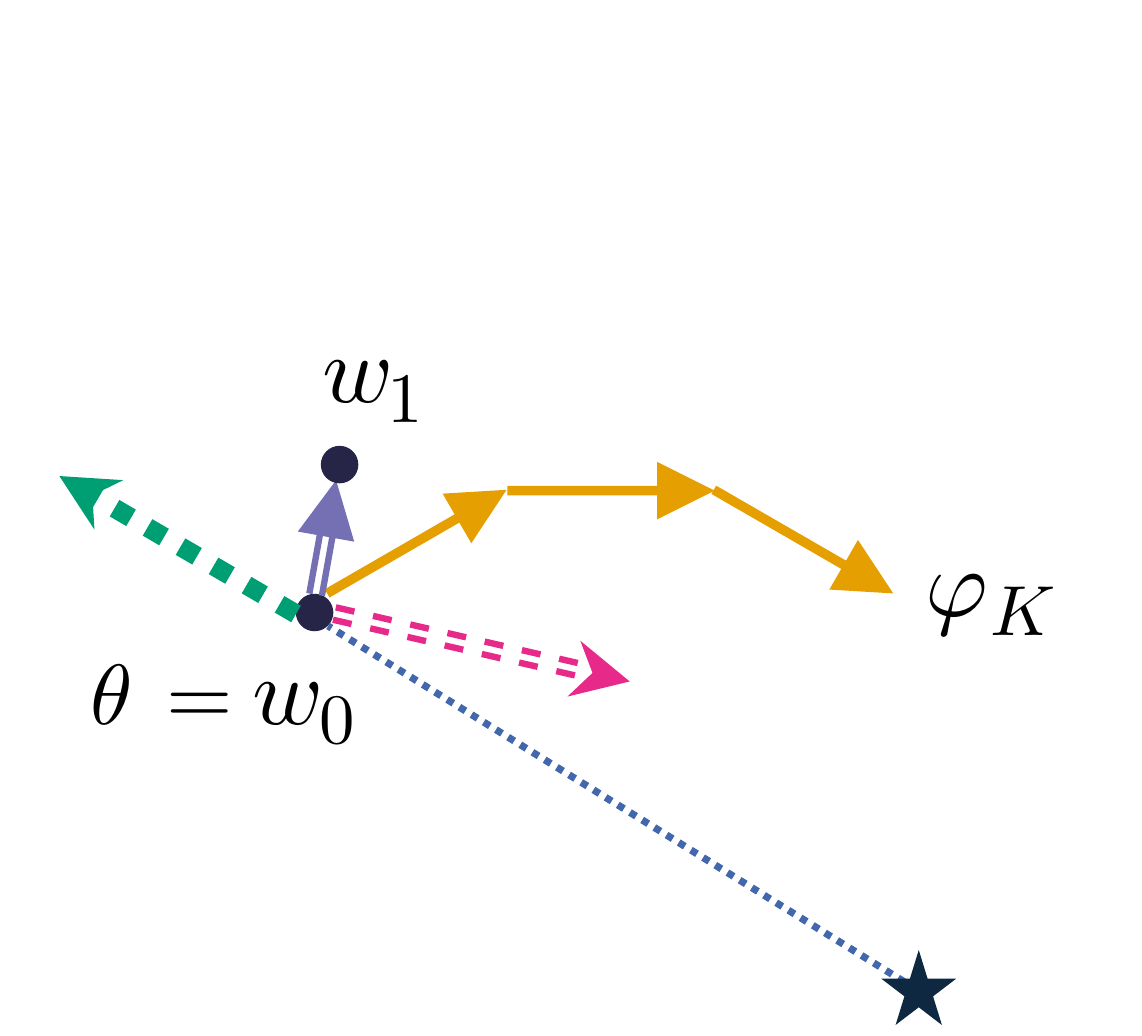}
        \subcaption{$i=0$}
        \label{fig:ours_outer_1}
    \end{subfigure}
    \hfill
    \begin{subfigure}[b]{0.325\linewidth}
        \centering
        \includegraphics[trim={0cm 0cm 0cm 0cm}, clip, width=1.1\linewidth]{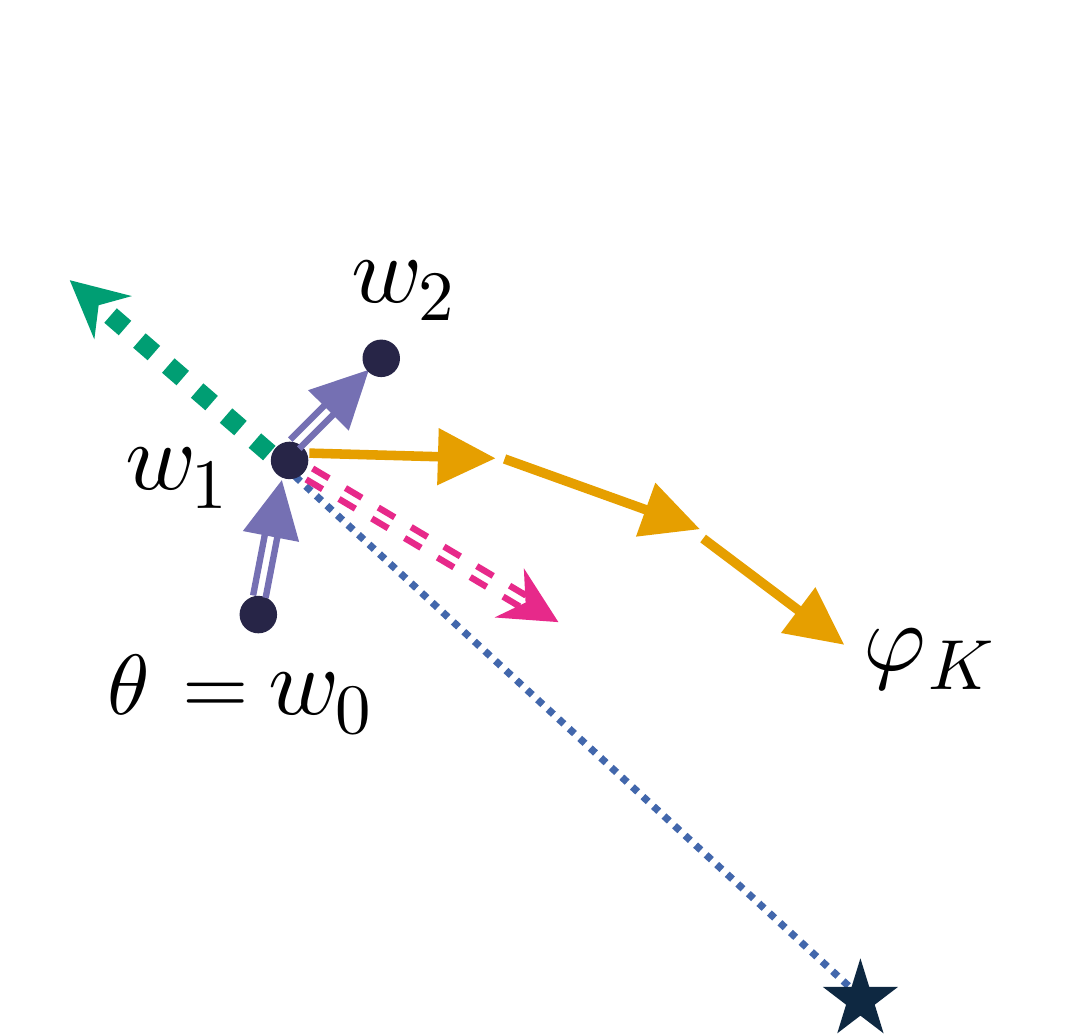}
        \subcaption{$i=1$}
        \label{fig:ours_outer_2}
    \end{subfigure}
    \hfill
    \begin{subfigure}[b]{0.325\linewidth}
        \centering
        \includegraphics[trim={0cm 0cm 0cm 0cm}, clip, width=1.1\linewidth]{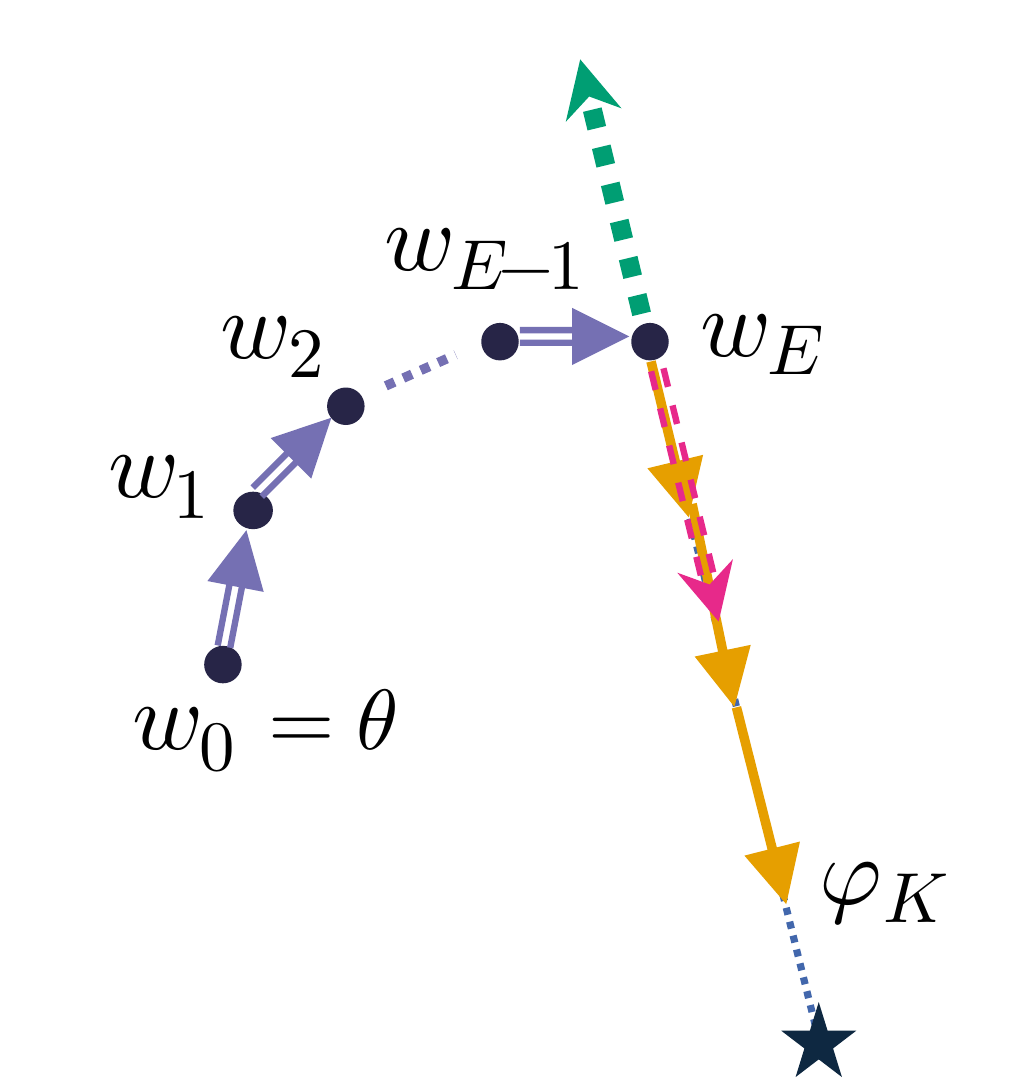}
        \subcaption{$i=E-1$}
        \label{fig:ours_outer_e}
    \end{subfigure}
    \caption{
        Illustration of FedMeNF's local meta-optimization in a client.
        The local meta-learner $w$ does not get closer to the optimal neural field and maintains the initial gap (the distance between the global meta-learner $\theta$ ($=w_0$) and the optimal neural field).
        Nevertheless, as meta-optimization progresses, gradients are being aligned, and the local meta-learner can find $w_E$ from which the optimal neural field is reached with a few gradient steps. 
    }
    \vspace{-0.2em}
    \label{fig:ours_illustration}
\end{figure}

\subsection{The Privacy-Preserving Loss}
\label{sec:l_pp}

\Cref{prop:g_maml}--\ref{prop:privacy_metric} theoretically show how client data privacy is leaked in federated meta-learning.
Now, we introduce a privacy-preserving loss $L_{pp}$ as follows, ensuring that it regularizes the $g_K$ term that increases $\text{PSNR}_p$:
\begin{equation}
    \label{eq:pp_loss}
        L_{pp}(\gamma, w_i, \varphi_K, B_K) = L(\varphi_K, B_K) - \gamma L(w_i, B_K),
\end{equation}
where $\gamma$ is a regularization coefficient that determines the portion of the $g_K$.
In \cref{prop:menf} below, we can obtain the privacy-preserving local meta-gradient $g_{pp}$ using $L_{pp}$.
\begin{proposition}
    \label{prop:menf}
    Let
    $g_{pp} = \nabla_{w_i} L_{pp}(\gamma, w_i, \varphi_K, B_K)$.
    Then, the first-order approximation of $g_{pp}$ and $\Delta L_{i+1}$ become
    \begin{align}
        \label{eq:menf}
        g_{pp} &\approx (1-\gamma) \cdot g_K - \lambda_{i} \mathcal{I}_{K}, \\
        \label{eq:privacy_metric_and_gamma}
        \Delta L_{i+1} &\approx - \lambda_{o} (1 - \gamma) (g_K)^2 \leq 0.
    \end{align}
\end{proposition}
Note $\mathcal{I}_{K}$ is defined in \cref{eq:g_maml}.
At $\gamma=1$, the $g_K$ term, which causes privacy leakage, is completely removed from $g_{pp}$.
This means that the local meta-learner no longer contains information about the client's local dataset and learns only fast optimization by maximizing the inner products between gradients across minibatches.
Each client can choose a different $\gamma$ that aligns with its desired privacy guarantee.We will empirically show that as $\gamma$ increases, the privacy leakage from the client's local dataset decreases in \cref{sec:exp_privacy_preservation}.

There are two ways to set $\gamma$: as a fixed hyperparameter or adjusted dynamically.
While we use the former approach as our default, we also propose the following adaptive method.
Similar to how $\epsilon$ defines the privacy boundary in Differential Privacy, we introduce a privacy budget $\zeta$, limiting the total magnitude of the loss change defined in \cref{eq:privacy_metric_and_gamma}:
\begin{gather}
    \lvert \Delta L_{i+1} \rvert \cdot R \cdot E \cdot M / N \leq \zeta, \\
    \gamma = \min(\max(1 - N \zeta / (REM\lambda_{o} (g_K)^2), 0), 1).
\end{gather}

\subsection{Privacy-Preserving Local Meta-Optimization}
\Cref{fig:ours_illustration} illustrates FedMeNF's local meta-optimization.
In existing FML approaches, each outer loop iteration brings the local meta-learner $w_i$ closer to the optimal neural field for the client’s private data.
Sharing the optimized local meta-learner $w_E$ with the server leads to significant privacy leakage, as the server can directly exploit $w_E$ to reconstruct the original private data.
In contrast, FedMeNF explicitly prevents the $w_i$ from converging to the optimal neural field, focusing solely on gradient alignment.
Thus, FedMeNF trains a local meta-learner that retains minimal private data while still enabling efficient optimization.

\begin{figure}[!t]
    \centering
    \includegraphics[width=0.95\linewidth]{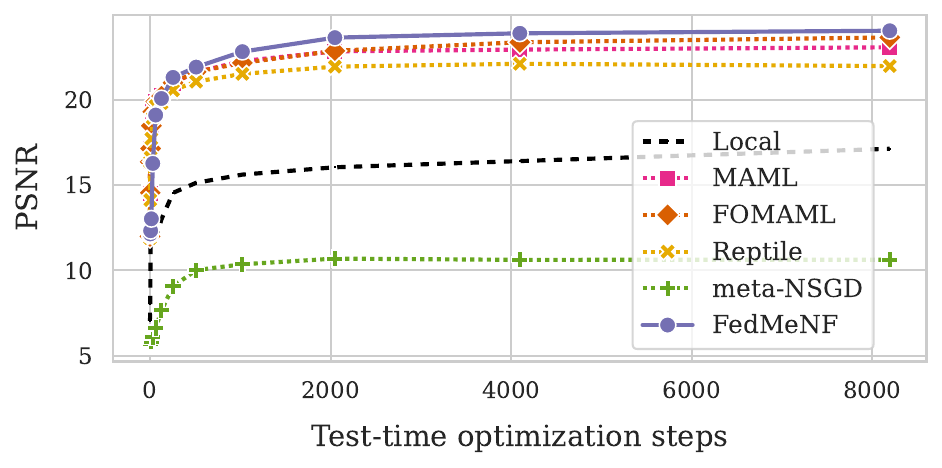}
    \vspace{-8pt}
    \caption{Test-time optimization of the baselines and our FedMeNF ($\gamma = 0.75$) using FedAvg~\cite{fedavg} on the Cars dataset~\cite{chang2015shapenet}.}
    \label{fig:plot_per_round}
\end{figure}

\begin{figure}[t]
    \centering
    \begin{subfigure}[b]{0.435\linewidth}
        \centering
        \includegraphics[width=\linewidth]{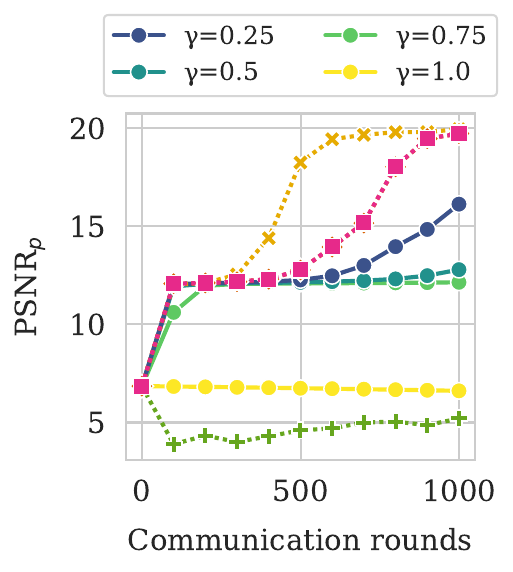}
        \subcaption{$\text{PSNR}_p$ / Rounds}
        \label{fig:plot_psnr_p_per_round}
    \end{subfigure}
    \hspace{-5pt}
    \hfill
    \begin{subfigure}[b]{0.565\linewidth}
        \centering
        \includegraphics[width=\linewidth]{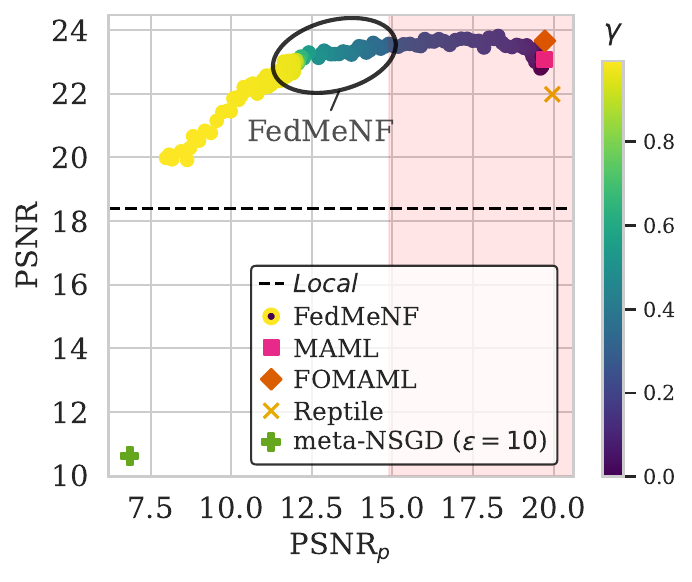}
        \subcaption{PSNR / $\text{PSNR}_p$}
        \label{fig:plot_gamma_scatter}
    \end{subfigure}
    
    \vspace{-5pt}
    \caption{
        Comparison of privacy preservation between baselines and our FedMeNF with different $\gamma$ values using FedAvg~\cite{fedavg} on the Cars dataset~\cite{chang2015shapenet}.
        Baselines like MAML, which excel in reconstruction (PSNR$\uparrow$), exhibit high $\text{PSNR}_p$ value, while methods like Reptile, which have lower $\text{PSNR}_p$ value, perform poorly in reconstruction (PSNR$\downarrow$).
        FedMeNF effectively adjusts $\gamma$ to maintain an acceptable $\text{PSNR}_p$ value while achieving a comparable PSNR.
    }
    \label{fig:plot_privacy_perservation}
\end{figure}

\begin{table*}[t]
    \centering
    \small
    \resizebox{1.0\linewidth}{!}{

        \begin{tabular}{p{0.8cm}l|ccc|ccc|ccc|ccc}
            \toprule
            
            \multicolumn{2}{c}{\bf{Modality}} &
            \multicolumn{3}{c}{Image} &
            \multicolumn{3}{c}{Video} & 
            \multicolumn{6}{c}{3D (NeRF)} \\
            
            \multicolumn{2}{c}{\bf{Dataset}} &
            \multicolumn{3}{c}{PetFace~\cite{shinoda2025petface}} &
            \multicolumn{3}{c}{GolfDB~\cite{mcnally2019golfdb}} &
            \multicolumn{3}{c}{Cars~\cite{chang2015shapenet}} &
            \multicolumn{3}{c}{FaceScape~\cite{zhu2023facescape, yang2020facescape}} \\
            
            \cmidrule(lr){3-5} \cmidrule(lr){6-8} \cmidrule(lr){9-11} \cmidrule(lr){12-14}
            
            \multicolumn{2}{c}{\bf{Method \textbackslash \space Metric}} 
            & \multicolumn{1}{c}{$\text{PSNR}_p$} & \multicolumn{1}{c}{PSNR} & \multicolumn{1}{c}{$\Delta$(↑)}
            & \multicolumn{1}{c}{$\text{PSNR}_p$} & \multicolumn{1}{c}{PSNR} & \multicolumn{1}{c}{$\Delta$(↑)}
            & \multicolumn{1}{c}{$\text{PSNR}_p$} & \multicolumn{1}{c}{PSNR} & \multicolumn{1}{c}{$\Delta$(↑)}
            & \multicolumn{1}{c}{$\text{PSNR}_p$} & \multicolumn{1}{c}{PSNR} & \multicolumn{1}{c}{$\Delta$(↑)}
            \\
            
            \midrule
            \textit{Local} &  & - & 22.29  & - & - & 26.92 & - & - & 17.13  & - & - & 23.67  & - \\
            \midrule[0.25pt]
            FedAvg
            & + MAML          & \ccr 16.57 &\ccr  27.39   &\ccr  10.82 &\ccr  21.21  &\ccr  29.68  &\ccr  8.47  &\ccr 19.73  &\ccr 23.08  &\ccr  3.35 &\ccr 21.31  &\ccr 28.59 &\ccr  7.28  \\
            & + FOMAML        & \ccr 18.52 &\ccr  23.15   &\ccr  4.63 &\ccr  20.82  &\ccr  28.57  &\ccr  7.75  &\ccr 19.73  &\ccr 23.66  &\ccr  3.93 &\ccr 21.24  &\ccr 28.65 &\ccr  7.41  \\
            & + Reptile       & \ccr 17.39 &\ccr  22.52   &\ccr  5.13  &\ccr  19.89  &\ccr  27.22  &\ccr  7.33  &\ccr 19.96    &\ccr 21.98  &\ccr 2.02&   \ccr 21.92 &\ccr  28.24 & \ccr 6.32 \\
            & + meta-NSGD     &\ccg 12.49 &\ccg 5.15    & \ccg -7.34 & \ccg 10.96  &\ccg   4.85  &\ccg -6.11  &\ccg  6.85  &\ccg 10.62  &\ccg  3.77 &\ccg  7.72  &\ccg 11.29 &\ccg  3.57  \\
            & + \textbf{Ours} & \textbf{14.77} &\textbf{ 27   }   & \textbf{12.23} &      \textbf{17.31}  &      \textbf{28.89}  &     \textbf{11.58}  &     \textbf{12.15}  &     \textbf{24.05}  &     \textbf{11.9 } &     \textbf{15.16}  &     \textbf{27.88} &      \textbf{12.72} \\
            \midrule[0.25pt]
            FedProx
            & + MAML          &\ccr  16.58 &\ccr  27.49   &\ccr  10.91 & \ccr 20.93  & \ccr 29.26  &\ccr  8.33  &\ccr 19.67  &\ccr 22.94  &\ccr  3.27  &\ccr 21.1  &\ccr 28.54  &\ccr  7.44 \\
            & + FOMAML        &\ccr  18.53 &\ccr  24.32   &\ccr   5.79 & \ccr 21.21  & \ccr 28.76  &\ccr  7.55  &\ccr 19.67  &\ccr 23.73  &\ccr  4.06  &\ccr 21.02 &\ccr 28.52  &\ccr  7.5  \\
            & + Reptile       &\ccr  17.37 &\ccr  22.5    &\ccr  5.13  & \ccr 19.79  & \ccr 27.21  &\ccr  7.42  &\ccr 19.95    &\ccr 22.11  &\ccr 2.16& \ccr 21.89 &  \ccr 28.21  & \ccr 6.32 \\
            & + meta-NSGD     &\ccg 12.49 &\ccg 5.15    & \ccg -7.34 & \ccg 10.96  &\ccg   4.84  &\ccg -6.12  &\ccg  6.85  &\ccg 10.91  &\ccg  4.06  &\ccg  7.72 &\ccg 11.29  &\ccg  3.57 \\
            & + \textbf{Ours} & \textbf{14.44} &\textbf{ 27.08}   & \textbf{12.64} &      \textbf{15.68}  &      \textbf{28.96}  &     \textbf{13.28}  &     \textbf{12.14}  &     \textbf{23.98}  &     \textbf{11.84}  &    \textbf{ 14.09} &     \textbf{27.54}  &    \textbf{13.45} \\
            \midrule[0.25pt]
            Scaffold
            & + MAML          &\ccr  16.71   &\ccr  27.66  &\ccr  10.95     & \ccr 21.3   & \ccr 29.11  &\ccr  7.81  &\ccr 19.07  &\ccr 24.21  &\ccr  5.14  &\ccr 21.09 &\ccr 28.51  &\ccr  7.42 \\
            & + FOMAML        &\ccr  18.55   &\ccr  23.98  &\ccr   5.43     & \ccr 21.29  & \ccr 28.82  &\ccr  7.53  &\ccr 19.07  &\ccr 24.47  &\ccr  5.4   &\ccr 21.01 &\ccr 28.51  &\ccr  7.5  \\
            & + Reptile       &\ccr  17.37   &\ccr  22.47  &\ccr  5.1       & \ccr 20     & \ccr 27.23  &\ccr  7.23  &\ccr 19.81    &\ccr 22.6   &\ccr 2.79&     \ccr 21.79 &  \ccr 28.14  &  \ccr 6.35 \\
            & + meta-NSGD     &\ccg 12.49   &\ccg 5.13   &\ccg  -7.36     & \ccg 10.96  &\ccg   4.84  &\ccg -6.12  &\ccg  6.85  &\ccg 10.63  &\ccg 3.78   &\ccg  7.72 &\ccg 11.29  &\ccg  3.57 \\
            & + \textbf{Ours} & \textbf{14.93}   &\textbf{ 27.19}  & \textbf{12.26}     &      \textbf{16.21}  &      \textbf{29.23}  &     \textbf{13.02}  &     \textbf{14.14}  &     \textbf{24.34}  &     \textbf{9.94}  &     \textbf{13.94} &     \textbf{27.53}  &     \textbf{13.59} \\
            \midrule[0.25pt]
            FedNova
            & + MAML          & \ccr 16.52 & \ccr 27.51   &\ccr  10.99 & \ccr 21.47  & \ccr 29.64  &\ccr  8.17  &\ccr 19.72  &\ccr 23.63  &\ccr 3.91   &\ccr 21.37 &\ccr 28.59  &\ccr  7.22 \\
            & + FOMAML        & \ccr 18.5  &\ccr  24.1    &\ccr   5.6  & \ccr 20.12  & \ccr 28.69  &\ccr  8.57  &\ccr 19.72  &\ccr 23.27  &\ccr 3.55   &\ccr 21.3  &\ccr 28.62  &\ccr  7.32 \\
            & + Reptile       & \ccr 17.38 & \ccr 22.52   &\ccr  5.14  & \ccr 19.93  & \ccr 27.26  &\ccr  7.33  &\ccr 19.96    &\ccr 22.7   &\ccr 2.74&   \ccr 21.98 &  \ccr 28.17  &  \ccr 6.19 \\
            & + meta-NSGD     &\ccg 12.49 &\ccg 5.14    & \ccg -7.35 & \ccg 10.96  &\ccg   4.84  &\ccg -6.12  &\ccg  6.85  &\ccg 10.6   &\ccg  3.75  &\ccg  7.72 &\ccg 11.29  &\ccg  3.57 \\
            & + \textbf{Ours} & \textbf{14.94} &\textbf{ 27.15}   & \textbf{12.21} &      \textbf{15.71}  &      \textbf{28.98}  &     \textbf{13.27}  &    \textbf{ 12.14}  &     \textbf{24.12}  &     \textbf{11.98}  &     \textbf{15.45} &     \textbf{27.86}  &     \textbf{12.41} \\
            \midrule[0.25pt]
            FedExP
            & + MAML          &\ccr  16.57 &\ccr  27.39    &\ccr  10.82 & \ccr 21.18  & \ccr 29.58  &\ccr  8.4   &\ccr 19.81  &\ccr 22.87  &\ccr  3.06  &\ccr 21.33 &\ccr 27.64  &\ccr  6.31 \\
            & + FOMAML        &\ccr  18.52 &\ccr  23.15    &\ccr   4.63 & \ccr 20.82  & \ccr 28.6   &\ccr  7.78  &\ccr 19.81  &\ccr 22.81  &\ccr  3     &\ccr 21.26 &\ccr 27.66  &\ccr  6.4  \\
            & + Reptile       &\ccr  17.39 &\ccr  22.64    &\ccr  5.25 & \ccr 19.89  & \ccr 27.23  &\ccr  7.34 &\ccr 20.91    &\ccr 22.03  &\ccr 1.12 &     \ccr 21.93 &  \ccr  28.27   &  \ccr 6.35  \\
            & + meta-NSGD     &\ccg 12.49 &\ccg 5.15     &\ccg  -7.34 & \ccg 10.96  &\ccg   4.85  &\ccg -6.11  &\ccg  6.85  &\ccg 10.61  &\ccg  3.76  &\ccg  7.72 &\ccg 11.29  &\ccg  3.57 \\
            & + \textbf{Ours} & \textbf{14.65} &\textbf{ 26.86}    & \textbf{12.21} &      \textbf{15.38}  &      \textbf{28.94}  &     \textbf{13.56}  &     \textbf{12.17}  &     \textbf{24.05}  &     \textbf{11.88}  &     \textbf{13.69} &     \textbf{26.12}  &     \textbf{12.43} \\
            \midrule[0.25pt]
            FedACG
            & + MAML          &\ccr  16.63 &\ccr  27.5  &\ccr  10.87    & \ccr 21.09  & \ccr 29.51  &\ccr  8.42  &\ccr 19.9   &\ccr 22     &\ccr  2.1   &\ccr 22.28 &\ccr 28.02  &\ccr  5.74 \\
            & + FOMAML        &\ccr  18.48 &\ccr  24.04 & \ccr   5.56    & \ccr 21.12  & \ccr 28.7   &\ccr  7.58  &\ccr 19.9   &\ccr 21.95  &\ccr  2.05  &\ccr 22.24 &\ccr 28.03  &\ccr  5.79 \\
            & + Reptile       &\ccr  17.38 &\ccr  22.63 &\ccr  5.25     & \ccr 19.83  & \ccr 27.23  &\ccr  7.4  &\ccr 20.13    &\ccr 22.26  &\ccr 2.13 &   \ccr 22.07 & \ccr 28.32  &  \ccr 6.25 \\
            & + meta-NSGD     &\ccg 12.49 &\ccg 5.15  &\ccg  -7.34    & \ccg 10.96  &\ccg   4.84  &\ccg -6.12  &\ccg  6.85  &\ccg 10.59  &\ccg  3.74  &\ccg  7.72 &\ccg 11.29  &\ccg  3.57 \\
            & + \textbf{Ours} & \textbf{14.71} &\textbf{ 26.97} & \textbf{12.26}    &      \textbf{15.94}  &      \textbf{28.99}  &     \textbf{13.05}  &     \textbf{10.93}  &     \textbf{22.45}  &     \textbf{11.52}  &    \textbf{ 14.83} &     \textbf{27.46}  &     \textbf{12.63} \\
            
            \bottomrule
        \end{tabular}
        }
    \caption{
        Comparison of reconstruction (PSNR(↑)) and privacy ($\text{PSNR}_p$(↓)) quality across different modalities: images, videos, and 3D NeRF scenes.
        $\Delta$ is defined as $\text{PSNR} - \text{PSNR}_p$ where a larger value indicates a better trade-off between reconstruction quality and privacy protection.
        FedMeNF demonstrates a balanced performance by maintaining high PSNR for reconstruction and low $\text{PSNR}_p$ for privacy.
        The \sethlcolor{red!10}{\hl{red-highlighted cases}} indicate cases where $\text{PSNR}_p$ is high enough to qualitatively confirm privacy leakage,
        and the \sethlcolor{gray!20}{\hl{gray-highlighted cases}} indicate cases where the reconstruction performance is lower than that of the \textit{Local}.
    }
    \label{tab:main}
\end{table*}

\begin{figure*}[!t]
    \centering
    \captionsetup[subfigure]{font=scriptsize}
    \begin{subfigure}[b]{0.15\linewidth}
        \centering
        \includegraphics[trim={0.5cm 0.5cm 0.5cm 0.5cm}, clip, width=0.7\linewidth]{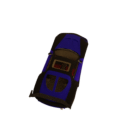}
        \subcaption{GT (PSNR)}
    \end{subfigure}
    \hfill
    \begin{subfigure}[b]{0.15\linewidth}
        \centering
        \includegraphics[trim={0.5cm 0.5cm 0.5cm 0.5cm}, clip, width=0.7\linewidth]{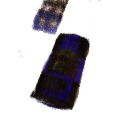}
        \subcaption{\textit{Local} (14.97)}
    \end{subfigure}
    \hfill
    \begin{subfigure}[b]{0.15\linewidth}
        \centering
        \includegraphics[trim={0.5cm 0.5cm 0.5cm 0.5cm}, clip, width=0.7\linewidth]{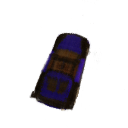}
        \subcaption{MAML (21.22)}
    \end{subfigure}
    \hfill
    \begin{subfigure}[b]{0.15\linewidth}
        \centering
        \includegraphics[trim={0.5cm 0.5cm 0.5cm 0.5cm}, clip, width=0.7\linewidth]{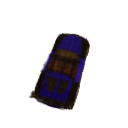}
        \subcaption{FOMAML ({21.44})}
    \end{subfigure}
    \hfill
    \begin{subfigure}[b]{0.15\linewidth}
        \centering
        \includegraphics[trim={0.5cm 0.5cm 0.5cm 0.5cm}, clip, width=0.7\linewidth]{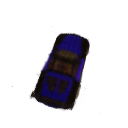}
        \subcaption{Reptile (21.18)}
    \end{subfigure}
    \hfill
    \begin{subfigure}[b]{0.15\linewidth}
        \centering
        \includegraphics[trim={0.5cm 0.5cm 0.5cm 0.5cm}, clip, width=0.7\linewidth]{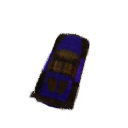}
        \subcaption{FedMeNF (\textbf{21.92})}
    \end{subfigure}

    \vspace{0.5em}
    \vfill
    \begin{subfigure}[b]{0.15\linewidth}
        \centering
        \includegraphics[trim={0.cm 0.cm 0.cm 0.cm}, clip, width=0.7\linewidth]{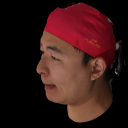}
        \subcaption{GT (PSNR)}
    \end{subfigure}
    \hfill
    \begin{subfigure}[b]{0.15\linewidth}
        \centering
        \includegraphics[trim={0.cm 0.cm 0.cm 0.cm}, clip, width=0.7\linewidth]{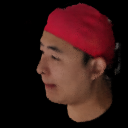}
        \subcaption{\textit{Local} (32.69)}
    \end{subfigure}
    \hfill
    \begin{subfigure}[b]{0.15\linewidth}
        \centering
        \includegraphics[trim={0.cm 0.cm 0.cm 0.cm}, clip, width=0.7\linewidth]{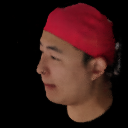}
        \subcaption{MAML (33.17)}
    \end{subfigure}
    \hfill
    \begin{subfigure}[b]{0.15\linewidth}
        \centering
        \includegraphics[trim={0.cm 0.cm 0.cm 0.cm}, clip, width=0.7\linewidth]{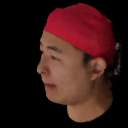}
        \subcaption{FOMAML ({33.26})}
    \end{subfigure}
    \hfill
    \begin{subfigure}[b]{0.15\linewidth}
        \centering
        \includegraphics[trim={0.cm 0.cm 0.cm 0.cm}, clip, width=0.7\linewidth]{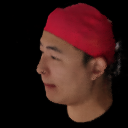}
        \subcaption{Reptile (32.28)}
    \end{subfigure}
    \hfill
    \begin{subfigure}[b]{0.15\linewidth}
        \centering
        \includegraphics[trim={0.cm 0.cm 0.cm 0.cm}, clip, width=0.7\linewidth]{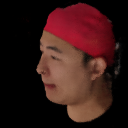}
        \subcaption{FedMeNF (\textbf{33.54})}
    \end{subfigure}
    \vspace{-0.5em}
    \caption{
        Qualitative results on the Cars~\cite{chang2015shapenet} and FaceScape~\cite{zhu2023facescape,yang2020facescape} datasets. Test-time optimization was performed for 8192 steps on the Cars dataset and 32768 steps on the FaceScape dataset.
    }
    \label{fig:car_face_qulaitative}
\end{figure*}

\section{Experiments}

\noindent \textbf{Setup.}
We assume a scenario where $N=50$ clients participate in the training. The training consists of 1000 communication rounds, with 5 participants randomly selected per round.
We employ the SIREN architecture \cite{sitzmann2020siren}  for image and video modalities, which consists of 6 layers with a hidden dimension of 128. For neural radiance fields (NeRF)~\cite{mildenhall2021nerf}, we adopt the model used in \cite{tancik2021learnedit, chen2022transinr}, which consists of 6 layers with a hidden dimension of 256.
Please refer to \cref{sec:implementation_details} for more implementation details.

\noindent \textbf{Datasets.}
We conduct experiments across the datasets of various modalities, including images, videos, and NeRFs.
To evaluate performance with limited local training data per client, we select scenarios in which each client has only one or very few tasks.
For images, we use the cat category from the PetFace dataset~\cite{shinoda2025petface}, assuming each client has images of a unique cat instance.
For videos, we use the GolfDB dataset~\cite{mcnally2019golfdb}, where each client has videos of only one person's golf swings.
For NeRFs, we use the Cars category from the ShapeNet dataset~\cite{chang2015shapenet} and the FaceScape dataset~\cite{zhu2023facescape,yang2020facescape} for human faces.
Please refer to \cref{sec:dataset_suppl} for detailed distribution information for each dataset.

\noindent \textbf{Baselines.}
We test a lower bound baseline referred to as \textit{Local}, using only local dataset without an FL approach.
We construct baselines by combining an FL algorithm and a meta-learning algorithm~\cite{fedmeta, fallah202perfedavg}.
For the FL algorithm, we use not only FedAvg~\cite{fedavg} as the most standard method, but also state-of-the-art methods including FedProx~\cite{fedprox}, Scaffold~\cite{scaffold}, FedNova~\cite{wang2020fednova}, FedExP~\cite{fedexp}, and FedACG~\cite{kim2024fedacg}.
For the meta-learning algorithm, we select widely used methods such as MAML~\cite{finn2017maml}, FOMAML~\cite{finn2017maml}, and Reptile~\cite{onfirst}, alongside HF-MAML~\cite{fallah2020hfmaml, fallah202perfedavg} and A-MAML~\cite{li2023amaml}.
We also include meta-NSGD~\cite{zhou2022metansgd}, the meta-learning algorithm with task-level differential privacy.

\subsection{Fast Optimization}
\Cref{fig:plot_per_round} illustrates the test-time optimization (TTO) performance of FedMeNF against baselines.
Compared to \textit{Local}, FML approaches reach higher PSNR more quickly.
FedMeNF achieves 40.4\% and 74.64\% higher PSNR than \textit{Local} with and without TTO, respectively.
FedMeNF also achieves a higher PSNR in just 39 steps, making it about 210x faster than \textit{Local}, which requires 8192 steps.

\subsection{Privacy Preservation}
\label{sec:exp_privacy_preservation}

\begin{figure}[t]
    \centering
    \captionsetup[subfigure]{font=scriptsize}
    \hfill
    \begin{subfigure}[b]{0.14\textwidth}
        \centering
        \includegraphics[width=0.9\linewidth]{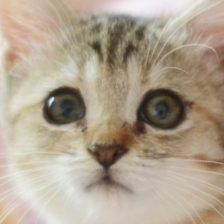}
        \subcaption{GT ($\text{PSNR}_p$)}
    \end{subfigure}
    \hfill
    \begin{subfigure}[b]{0.14\textwidth}
        \centering
        \includegraphics[width=0.9\linewidth]{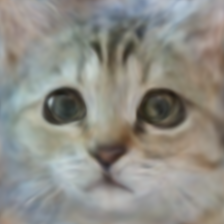}
        \subcaption{MAML (17.33)}
    \end{subfigure}
    \hfill
    \begin{subfigure}[b]{0.14\textwidth}
        \centering
        \includegraphics[width=0.9\linewidth]{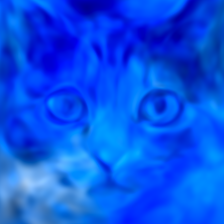}
        \subcaption{FedMeNF (\textbf{5.84})}
    \end{subfigure}
    \hfill
    
    \vfill
    
    \begin{subfigure}[b]{0.45\textwidth}
        \centering
        \includegraphics[width=\linewidth]{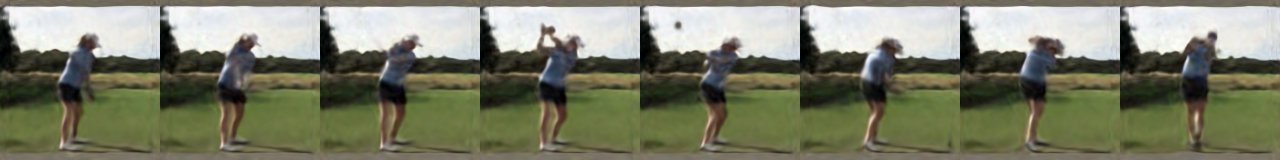}
        \subcaption{MAML ($\text{PSNR}_p = 23.43$)}
    \end{subfigure}
    \vfill
    \begin{subfigure}[b]{0.45\textwidth}
        \centering
        \includegraphics[width=\linewidth]{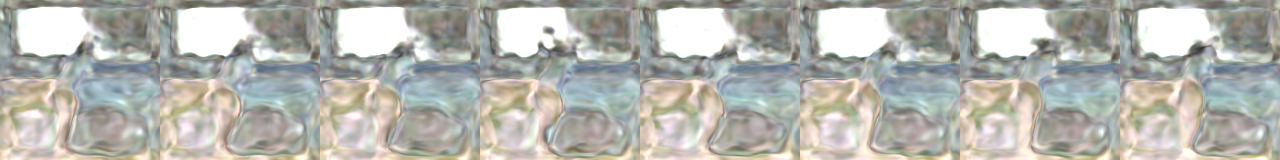}
        \subcaption{FedMeNF ($\text{PSNR}_p = \textbf{8.38}$)}
    \end{subfigure}
    
    \vfill

    \vspace{-8pt}
    \rule{\linewidth}{0.75pt}
    \vspace{-8pt}

    \hfill
    \begin{subfigure}[b]{0.14\textwidth}
        \centering
        \includegraphics[width=0.9\linewidth]{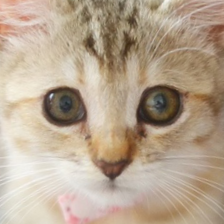}
        \subcaption{GT (PSNR)}
    \end{subfigure}
    \hfill
    \begin{subfigure}[b]{0.14\textwidth}
        \centering
        \includegraphics[width=0.9\linewidth]{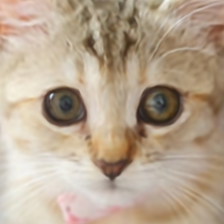}
        \subcaption{MAML (\textbf{35.74})}
    \end{subfigure}
    \hfill
    \begin{subfigure}[b]{0.14\textwidth}
        \centering
        \includegraphics[width=0.9\linewidth]{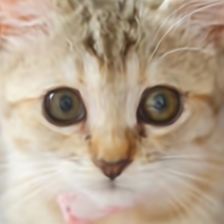}
        \subcaption{FedMeNF (35.16)}
    \end{subfigure}
    \hfill

    \vfill

    \begin{subfigure}[b]{0.45\textwidth}
        \centering
        \includegraphics[width=\linewidth]{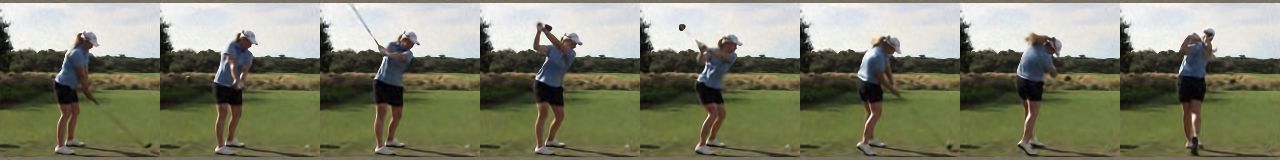}
        \subcaption{MAML ($\text{PSNR} = \text{\textbf{29.65}}$)}
    \end{subfigure}
    \vfill
    \begin{subfigure}[b]{0.45\textwidth}
        \centering
        \includegraphics[width=\linewidth]{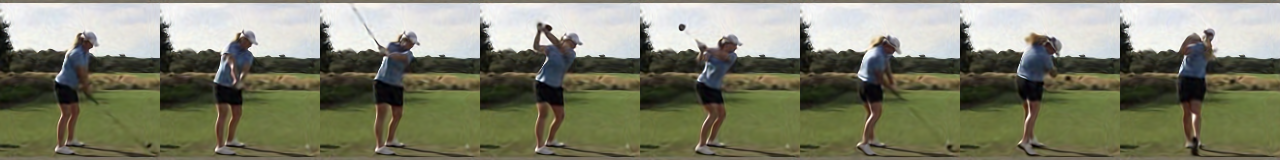}
        \subcaption{FedMeNF ($\text{PSNR} = 29.15$)}
    \end{subfigure}
    \vspace{-6pt}
    \caption{
        \textbf{Upper.} Reconstruction results of the client’s private image/video on the server: (b, d) using MAML and (c, e) using FedMeNF.
        \textbf{Lower.} Reconstruction results of a new private image/video on the client: (g, i) using MAML and (h, j) using FedMeNF.
        Our FedMeNF effectively enhances privacy as in (c, e) with minimal impact on the reconstruction quality as in (h, j).
    }
    \label{fig:tto0_psnr}
\end{figure}

\Cref{fig:plot_privacy_perservation} illustrates the privacy preservation performance across varying values of the regularization coefficient $\gamma$.
\cref{fig:plot_psnr_p_per_round} presents the privacy metric $\text{PSNR}_p$ over communication rounds.
FedMeNF achieves an 18.25\%, 35.17\%, 38.42\%, and 65.28\% improvement in the privacy metric $\text{PSNR}_p$ at $\gamma=0.25, 0.5$, and $0.75$, compared to the baseline.
\cref{fig:plot_gamma_scatter} shows the trade-off between privacy preservation ($\text{PSNR}_p$↓) and reconstruction performance (PSNR↑).
Meta-NSGD, which utilizes the differential privacy mechanism, achieves lower $\text{PSNR}_p$ values, indicating stronger privacy protection.
However, it suffers from significant degradation in reconstruction performance.
In contrast, existing meta-learning methods like MAML achieve high PSNR at the cost of greater vulnerability to privacy leakage with their higher $\text{PSNR}_p$.
FedMeNF establishes an efficient frontier that balances privacy protection and reconstruction performance, effectively reducing $\text{PSNR}_p$ with minimal impact on PSNR.
Qualitative results are presented in \cref{fig:tto0_psnr,fig:car_face_qulaitative}.

As shown in \cref{tab:main}, FedMeNF achieves comparable PSNR while maintaining an acceptable $\text{PSNR}_p$ across diverse modalities, as well as across various FL algorithms.
FedMeNF also performs well on other metrics, such as SSIM and LPIPS, as detailed in \cref{sec:additional_metrics}.

\subsection{Justification on the Privacy Metrics}
To investigate the practical relevance of the privacy metrics in capturing privacy leakage, we simulate two standard privacy attacks on the Cars dataset~\cite{chang2015shapenet}, assuming an honest-but-curious server.
The target models for these attacks are defined as the final local meta-learners from each client in the last communication round.
The first is the Membership Inference Attack~\cite{shokri2017mia}, which determines if a given data sample was in the training set of a black-box model.
We assess how accurately the server can predict the membership of the client's local data using the client's local meta-learner as the black-box model.
The second is the Property Inference Attack~\cite{ganju2018pia}, which infers a property of a given model's training set.
We evaluate the server's accuracy in classifying the vehicle category of the client's car using the client's local meta-learner.
The setup is detailed in \cref{sec:suppl:privacy_attack_scenarios}.
\Cref{tab:rebuttal-main} shows that worse privacy metrics correlate with greater vulnerability to privacy attacks.

\begin{figure}[t]
    \centering
    \includegraphics[width=0.95\linewidth]{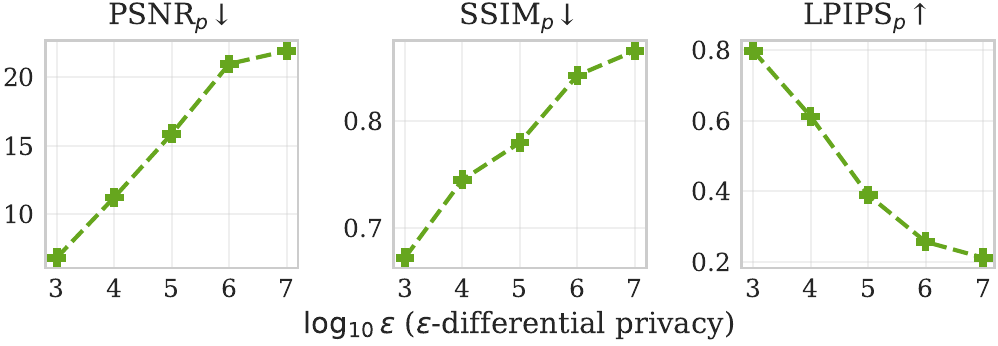}
    \vspace{-8pt}
    \caption{
        Correlation between $\epsilon$ and privacy metrics.
    }
    \label{fig:plot_epsilon}
\end{figure}

Moreover, we examine the correlation between the privacy metrics and $\epsilon$ of the differential privacy framework using meta-NSGD~\cite{zhou2022metansgd}.
\Cref{fig:plot_epsilon} shows that the privacy metrics degrade as $\epsilon$ increases, supporting their generalizability as a measure of privacy leakage.

\subsection{Generalization on Non-IID Settings}

To evaluate the generalization capability of FedMeNF on non-IID settings, we vary the data quantity per client using a Dirichlet distribution with parameter $\alpha$; for the NeRF scenes (where each client has one task), we vary the number of input views, while for the image and video modalities (where each task is a single image or video), we vary the number of tasks.
A larger $\alpha$ indicates that the data quantity per client is more uniform.
See \cref{sec:suppl:dataset_shapenet} for details.

\Cref{tab:main} presents the results for $\alpha = 5.0$, while \cref{tab:non-iid} shows the results for $\alpha = 10, 5.0, 1.0$.
For the baselines, a small $\alpha$ creates two problems.
First, local meta-optimization on a few views tends to result in over-fitted local models, while training on many views leads to well-trained ones.
In both cases, the server can reconstruct clients' private data with high fidelity, leading to privacy leakage.
Second, over-fitted models perform poorly on novel view synthesis, degrading the global meta-learner.
In contrast, FedMeNF is robust to this data imbalance because it only learns how to optimize rather than simply memorizing data.
This approach prevents over-fitting and thus achieves robust performance in both privacy preservation and novel view synthesis, regardless of the data quantity.

\begin{table}[t]
    \setlength{\tabcolsep}{2.5pt}

    \centering
    \small
    \resizebox{\linewidth}{!}{

        \begin{tabular}{l|cc|ccc|c}
            \toprule
            
            \multicolumn{1}{c}{\textbf{Metric}} 
            & \multicolumn{1}{c}{MIA} & \multicolumn{1}{c}{PIA} & \multicolumn{3}{c}{Privacy Metrics} & \multicolumn{1}{c}{NVS} \vspace{-2px}
            \\

            \cmidrule(lr){2-2} \cmidrule(lr){3-3} \cmidrule(lr){4-6} \cmidrule(lr){7-7} 
            
            \multicolumn{1}{c}{\bf{Method}} 
            & \multicolumn{1}{c}{Acc$\downarrow$} & \multicolumn{1}{c}{Acc$\downarrow$}
            & \multicolumn{1}{c}{$\text{PSNR}_p$$\downarrow$} & \multicolumn{1}{c}{$\text{SSIM}_p$$\downarrow$} & \multicolumn{1}{c}{$\text{LPIPS}_p$$\uparrow$} & \multicolumn{1}{c}{PSNR$\uparrow$} \vspace{-1px}
            \\
            
            \midrule
            MAML                          & 95.83 & 48.78 & 19.73 & 0.827 & 0.343 & 23.08 \\
            FOMAML                        & 97.22 & 48.78 & 19.73 & 0.827 & 0.343 & 23.66 \\
            Reptile   & 95.83 & 43.90 & 19.96 & 0.839 & 0.311 & 21.98 \\
            HF-MAML                       & 88.89 & 43.90 & 19.24 & 0.827 & 0.341 & 23.58 \\
            A-MAML                  & 95.83 & 43.90 & 19.64 & 0.826 & 0.343 & 23.13 \\
            \midrule[0.1pt]
            \textbf{Ours}                 & \textbf{55.56} & \textbf{29.27} & 12.15 & 0.798 & 0.537 & \textbf{24.05} \\
            \textbf{Ours} (adaptive)      & \textbf{55.56} & \textbf{29.27} & \textbf{12.09} & \textbf{0.797} & \textbf{0.538} & 23.19 \\
            \bottomrule
        \end{tabular}
    }
    \vspace{-4pt}
    \caption{
        Comparison of Membership Inference Attack (MIA) accuracy, Property Inference Attack (PIA) accuracy, Privacy Metrics, and Novel View Synthesis (NVS) quality.
    }
    \label{tab:rebuttal-main}
\end{table}

\subsection{Few-shot Novel View Synthesis}
We conduct experiments with varying average number of input views using FedAvg~\cite{fedavg} on the Cars~\cite{chang2015shapenet} dataset, and the results are shown in \cref{tab:few-view}.
For all methods, the view synthesis performance (PSNR) improves as the average number of input views increases.
FML approaches outperform the \textit{Local} across all cases.
Notably, FedMeNF achieves the highest PSNR with acceptable $\text{PSNR}_p$, regardless of the average number of input views. This confirms that FedMeNF performs well even in few-shot settings.

\begin{table}[!t]
    \setlength{\tabcolsep}{3.5pt}
    \centering
    \small
    \resizebox{\linewidth}{!}{

        \begin{tabular}{lcccccccc}
            \toprule

            \multicolumn{1}{c}{\textbf{Dirichlet $\alpha$}} & \multicolumn{2}{c}{$\alpha = 10$} & \multicolumn{2}{c}{$\alpha = 5.0$} & \multicolumn{2}{c}{$\alpha = 1.0$}  \vspace{-2px} \\
            
            \cmidrule(lr){2-3} \cmidrule(lr){4-5} \cmidrule(lr){6-7}
            
            \multicolumn{1}{c}{\bf{Method}} 
            & \multicolumn{1}{c}{$\text{PSNR}_p$} & \multicolumn{1}{c}{PSNR}
            & \multicolumn{1}{c}{$\text{PSNR}_p$} & \multicolumn{1}{c}{PSNR}
            & \multicolumn{1}{c}{$\text{PSNR}_p$} & \multicolumn{1}{c}{PSNR}  \vspace{-1px}
            \\
            
            \midrule
            MAML          & 19.62 & 23.33 & 19.73 & 23.08 & 20    & 21.24 \\
            FOMAML        & 19.62 & 23.1  & 19.73 & 23.66 & 20    & 21.25 \\
            Reptile       & 19.96 & 22.47 & 19.96 & 21.98 & 19.95 & 21.29 \\
            HF-MAML       & 19.18 & 23.49 & 19.24 & 23.58 & 18.84 & 22.2 \\
            A-MAML        & 19.57 & 23.29 & 19.64 & 23.13 & 19.75 & 21.44 \\
            \midrule[0.1pt]
            \textbf{Ours} & {12.13}  & {23.47} & {12.15} & \textbf{24.05} & \textbf{12.73} & \textbf{24.32} \\
            \textbf{Ours} (adaptive) & \textbf{12.1} & \textbf{23.52} & \textbf{12.09} & 23.19 & 13.73 & 20.06 \\
            \bottomrule
        \end{tabular}
    }
    \vspace{-4pt}
    \caption{
        Results of the novel view synthesis performance and privacy metric with different Dirichlet parameters $\alpha = [10, 5.0, 1.0]$.
    }
    \label{tab:non-iid}
\end{table}

\begin{table}[!t]
    \setlength{\tabcolsep}{3.5pt}
    \centering
    \small
    \resizebox{\linewidth}{!}{

        \begin{tabular}{lcccccccc}
            \toprule
            
            \multicolumn{1}{c}{\textbf{Avg. \# of views}} & \multicolumn{2}{c}{2-view} & \multicolumn{2}{c}{4-view} & \multicolumn{2}{c}{8-view} \vspace{-2px} \\
            
            \cmidrule(lr){2-3} \cmidrule(lr){4-5} \cmidrule(lr){6-7}
            
            \multicolumn{1}{c}{\bf{Method}} 
            & \multicolumn{1}{c}{$\text{PSNR}_p$} & \multicolumn{1}{c}{PSNR}
            & \multicolumn{1}{c}{$\text{PSNR}_p$} & \multicolumn{1}{c}{PSNR}
            & \multicolumn{1}{c}{$\text{PSNR}_p$} & \multicolumn{1}{c}{PSNR} \vspace{-1px}
            \\
            
            \midrule
            MAML          & 19.73 & 20.23 & 19.73 & 23.08 & 19.54 & 26.03 \\
            FOMAML        & 19.74 & 20.39 & 19.73 & 23.66 & 19.54 & 25.97 \\
            Reptile       & 18.59 & 20.8  & 19.96 & 21.98 & 20.34 & 26.3  \\
            HF-MAML       & 18.83 & 21.15 & 19.24 & 23.58 & 18.41 & 25.62 \\
            A-MAML        & 19.73 & 20.25 & 19.64 & 23.13 & 19.57 & 25.22 \\
            \midrule[0.1pt]
            \textbf{Ours} & \textbf{15.56} & \textbf{21.27} & {12.15} & \textbf{24.05} & \textbf{11.8} & \textbf{26.31} \\
            \textbf{Ours} (adaptive) & 15.84 & 20.21 & \textbf{12.09} & 23.19 & 12.03 & 25.82 \\

            \bottomrule
        \end{tabular}
    }
    \vspace{-4pt}
    \caption{
        Results of the novel view synthesis performance and privacy metric for 2, 4, and 8 input views per task.
    }
    \label{tab:few-view}
\end{table}

\section{Discussion}
Since FedMeNF is designed for distributed devices, we use a relatively simple NF model.
Although recent work~\cite{chen2022transinr, kim2023ipc, lee2024locality} has achieved generalizable NFs using large models like transformers, these methods are often impractical for FL with edge devices due to their high communication and computational demands. Future work should focus on scalable, generalizable NFs aligned with FL constraints.

\section{Conclusion}
We proposed FedMeNF, the first federated meta-learning framework explicitly designed for neural fields of private data. FedMeNF regulates the increase of our intuitive and widely applicable privacy metric to train the global neural fields meta-learner while preserving privacy.
Through experiments across diverse data modalities, including images, videos, and NeRFs, we demonstrate that FedMeNF effectively preserves client privacy while achieving rapid optimization speed and high reconstruction performance, even when local data are highly limited or non-IID.
A potential limitation of this study is that we used relatively simple neural field models and vision data only.
Future work will investigate generalizable, privacy-preserving neural fields for a broader range of personal data.
Measuring and preserving privacy in neural fields is a crucial and novel challenge which has received less attention. We believe this study will serve as an essential starting point for further research.

\section*{Acknowledgements}
This work was partially supported by the Institute of Information \& Communications Technology Planning \& Evaluation (IITP) grant funded by the Korea government (MSIT) (No. RS-2022-II220156, Fundamental research on continual meta-learning for quality enhancement of casual videos and their 3D metaverse transformation; No.~RS-2021-II211343, Artificial Intelligence Graduate School Program (Seoul National University)), National Research Foundation of Korea (NRF) grant funded by the Korea government (MSIT) (No.~2023R1A2C2005573), and Center for Applied Research in Artificial Intelligence (CARAI) grant funded by Defense Acquisition Program Administration (DAPA) and Agency for Defense Development (ADD) (UD230017TD). Gunhee Kim is the corresponding author.

{
    \small
    \bibliographystyle{ieeenat_fullname}
    \bibliography{main}

\begin{thebibliography}{71}
\providecommand{\natexlab}[1]{#1}
\providecommand{\url}[1]{\texttt{#1}}
\expandafter\ifx\csname urlstyle\endcsname\relax
  \providecommand{\doi}[1]{doi: #1}\else
  \providecommand{\doi}{doi: \begingroup \urlstyle{rm}\Url}\fi

\bibitem[Abadi et~al.(2016)Abadi, Chu, Goodfellow, McMahan, Mironov, Talwar, and Zhang]{abadi2016deepdp}
Martin Abadi, Andy Chu, Ian Goodfellow, H~Brendan McMahan, Ilya Mironov, Kunal Talwar, and Li Zhang.
\newblock Deep learning with differential privacy.
\newblock In \emph{ACM CCS}, 2016.

\bibitem[AbdulRahman et~al.(2020)AbdulRahman, Tout, Ould-Slimane, Mourad, Talhi, and Guizani]{abdulrahman2020centralfedsurvey}
Sawsan AbdulRahman, Hanine Tout, Hakima Ould-Slimane, Azzam Mourad, Chamseddine Talhi, and Mohsen Guizani.
\newblock A survey on federated learning: The journey from centralized to distributed on-site learning and beyond.
\newblock \emph{IEEE Internet of Things Journal}, 2020.

\bibitem[Agaram et~al.(2023)Agaram, Dewan, Sajnani, Poulenard, Krishna, and Sridhar]{agaram2023canonical}
Rohith Agaram, Shaurya Dewan, Rahul Sajnani, Adrien Poulenard, Madhava Krishna, and Srinath Sridhar.
\newblock Canonical fields: Self-supervised learning of pose-canonicalized neural fields.
\newblock In \emph{CVPR}, 2023.

\bibitem[Aledhari et~al.(2020)Aledhari, Razzak, Parizi, and Saeed]{aledhari2020fedsurveyaccess}
Mohammed Aledhari, Rehma Razzak, Reza~M Parizi, and Fahad Saeed.
\newblock Federated learning: A survey on enabling technologies, protocols, and applications.
\newblock \emph{IEEE Access}, 2020.

\bibitem[Bun and Steinke(2016)]{bun2016concentrated}
Mark Bun and Thomas Steinke.
\newblock Concentrated differential privacy: Simplifications, extensions, and lower bounds.
\newblock In \emph{Theory of Cryptography}, 2016.

\bibitem[Chang et~al.(2015)Chang, Funkhouser, Guibas, Hanrahan, Huang, Li, Savarese, Savva, Song, Su, et~al.]{chang2015shapenet}
Angel~X Chang, Thomas Funkhouser, Leonidas Guibas, Pat Hanrahan, Qixing Huang, Zimo Li, Silvio Savarese, Manolis Savva, Shuran Song, Hao Su, et~al.
\newblock Shapenet: An information-rich 3d model repository.
\newblock \emph{arXiv:1512.03012}, 2015.

\bibitem[Chen et~al.(2022{\natexlab{a}})Chen, Gao, Kuang, Li, and Ding]{chen2022pflbench}
Daoyuan Chen, Dawei Gao, Weirui Kuang, Yaliang Li, and Bolin Ding.
\newblock pfl-bench: A comprehensive benchmark for personalized federated learning.
\newblock \emph{NeurIPS}, 2022{\natexlab{a}}.

\bibitem[Chen et~al.(2018)Chen, Luo, Dong, Li, and He]{fedmeta}
Fei Chen, Mi Luo, Zhenhua Dong, Zhenguo Li, and Xiuqiang He.
\newblock Federated meta-learning with fast convergence and efficient communication.
\newblock \emph{arXiv:1802.07876}, 2018.

\bibitem[Chen and Wang(2022)]{chen2022transinr}
Yinbo Chen and Xiaolong Wang.
\newblock Transformers as meta-learners for implicit neural representations.
\newblock In \emph{ECCV}, 2022.

\bibitem[Chen et~al.(2021)Chen, Liu, and Wang]{chen2021learning}
Yinbo Chen, Sifei Liu, and Xiaolong Wang.
\newblock Learning continuous image representation with local implicit image function.
\newblock In \emph{CVPR}, 2021.

\bibitem[Chen et~al.(2022{\natexlab{b}})Chen, Chen, Liu, Xu, Goel, Wang, Shi, and Wang]{chen2022videoinr}
Zeyuan Chen, Yinbo Chen, Jingwen Liu, Xingqian Xu, Vidit Goel, Zhangyang Wang, Humphrey Shi, and Xiaolong Wang.
\newblock Videoinr: Learning video implicit neural representation for continuous space-time super-resolution.
\newblock In \emph{CVPR}, 2022{\natexlab{b}}.

\bibitem[Coombes et~al.(2014)Coombes, beim Graben, Potthast, and Wright]{coombes2014nfbook}
Stephen Coombes, Peter beim Graben, Roland Potthast, and James Wright.
\newblock \emph{Neural fields: theory and applications}.
\newblock Springer, 2014.

\bibitem[Duan et~al.(2024)Duan, Li, Jiang, and He]{duan2024openfedsurvey}
Moming Duan, Qinbin Li, Linshan Jiang, and Bingsheng He.
\newblock Towards open federated learning platforms: Survey and vision from technical and legal perspectives, 2024.

\bibitem[Dupont et~al.(2022)Dupont, Kim, Eslami, Rezende, and Rosenbaum]{dupont2022functa}
Emilien Dupont, Hyunjik Kim, SM Eslami, Danilo Rezende, and Dan Rosenbaum.
\newblock From data to functa: Your data point is a function and you can treat it like one.
\newblock \emph{arXiv:2201.12204}, 2022.

\bibitem[Dwork and Rothblum(2016)]{dwork2016concentrated}
Cynthia Dwork and Guy~N Rothblum.
\newblock Concentrated differential privacy.
\newblock \emph{arXiv:1603.01887}, 2016.

\bibitem[Dwork et~al.(2006)Dwork, Kenthapadi, McSherry, Mironov, and Naor]{dwork2006dp}
Cynthia Dwork, Krishnaram Kenthapadi, Frank McSherry, Ilya Mironov, and Moni Naor.
\newblock Our data, ourselves: Privacy via distributed noise generation.
\newblock In \emph{EUROCRYPT}, 2006.

\bibitem[Fallah et~al.(2020{\natexlab{a}})Fallah, Mokhtari, and Ozdaglar]{fallah2020hfmaml}
Alireza Fallah, Aryan Mokhtari, and Asuman Ozdaglar.
\newblock On the convergence theory of gradient-based model-agnostic meta-learning algorithms.
\newblock In \emph{AISTATS}, 2020{\natexlab{a}}.

\bibitem[Fallah et~al.(2020{\natexlab{b}})Fallah, Mokhtari, and Ozdaglar]{fallah202perfedavg}
Alireza Fallah, Aryan Mokhtari, and Asuman Ozdaglar.
\newblock Personalized federated learning with theoretical guarantees: A model-agnostic meta-learning approach.
\newblock \emph{NeurIPS}, 2020{\natexlab{b}}.

\bibitem[Finn et~al.(2017)Finn, Abbeel, and Levine]{finn2017maml}
Chelsea Finn, Pieter Abbeel, and Sergey Levine.
\newblock Model-agnostic meta-learning for fast adaptation of deep networks.
\newblock In \emph{ICML}, 2017.

\bibitem[Ganju et~al.(2018)Ganju, Wang, Yang, Gunter, and Borisov]{ganju2018pia}
Karan Ganju, Qi Wang, Wei Yang, Carl~A Gunter, and Nikita Borisov.
\newblock Property inference attacks on fully connected neural networks using permutation invariant representations.
\newblock In \emph{ACM CCS}, pages 619--633, 2018.

\bibitem[Gao et~al.(2021)Gao, Guo, Zhang, Qiu, Wen, and Liu]{gao2021atsprivacy}
Wei Gao, Shangwei Guo, Tianwei Zhang, Han Qiu, Yonggang Wen, and Yang Liu.
\newblock Privacy-preserving collaborative learning with automatic transformation search.
\newblock In \emph{CVPR}, 2021.

\bibitem[Gu et~al.(2023)Gu, Wang, and Yeung]{gu2023generalizable}
Jeffrey Gu, Kuan-Chieh Wang, and Serena Yeung.
\newblock Generalizable neural fields as partially observed neural processes.
\newblock In \emph{ICCV}, 2023.

\bibitem[Hayes et~al.(2023)Hayes, Balle, and Mahloujifar]{hayes2023bounding}
Jamie Hayes, Borja Balle, and Saeed Mahloujifar.
\newblock Bounding training data reconstruction in dp-sgd.
\newblock In \emph{NeurIPS}, 2023.

\bibitem[Holden et~al.(2023)Holden, Dayoub, Harvey, and Chin]{holden2023fednerf}
Lachlan Holden, Feras Dayoub, David Harvey, and Tat-Jun Chin.
\newblock Federated neural radiance fields.
\newblock \emph{arXiv:2305.01163}, 2023.

\bibitem[Hong et~al.(2024)Hong, Yun, Jeon, and Kim]{hong2024fedavp}
Minui Hong, Junhyeog Yun, Insu Jeon, and Gunhee Kim.
\newblock Fedavp: Augment local data via shared policy in federated learning.
\newblock \emph{NeurIPS}, 2024.

\bibitem[Huang et~al.(2022)Huang, Liang, and Huang]{huang2022provable}
Yu Huang, Yingbin Liang, and Longbo Huang.
\newblock Provable generalization of overparameterized meta-learning trained with sgd.
\newblock \emph{NeurIPS}, 2022.

\bibitem[Imteaj et~al.(2021)Imteaj, Thakker, Wang, Li, and Amini]{imteaj2021fedsurveyiot}
Ahmed Imteaj, Urmish Thakker, Shiqiang Wang, Jian Li, and M~Hadi Amini.
\newblock A survey on federated learning for resource-constrained iot devices.
\newblock \emph{IEEE Internet of Things Journal}, 2021.

\bibitem[Jhunjhunwala et~al.(2023)Jhunjhunwala, Wang, and Joshi]{fedexp}
Divyansh Jhunjhunwala, Shiqiang Wang, and Gauri Joshi.
\newblock Fedexp: Speeding up federated averaging via extrapolation.
\newblock \emph{arXiv:2301.09604}, 2023.

\bibitem[Jiang et~al.(2019)Jiang, Kone{\v{c}}n{\`y}, Rush, and Kannan]{jiang2019perfed}
Yihan Jiang, Jakub Kone{\v{c}}n{\`y}, Keith Rush, and Sreeram Kannan.
\newblock Improving federated learning personalization via model agnostic meta learning.
\newblock \emph{arXiv:1909.12488}, 2019.

\bibitem[Karimireddy et~al.(2020)Karimireddy, Kale, Mohri, Reddi, Stich, and Suresh]{scaffold}
Sai~Praneeth Karimireddy, Satyen Kale, Mehryar Mohri, Sashank Reddi, Sebastian Stich, and Ananda~Theertha Suresh.
\newblock Scaffold: Stochastic controlled averaging for federated learning.
\newblock In \emph{ICML}, 2020.

\bibitem[Kim et~al.(2023)Kim, Lee, Kim, Cho, and Han]{kim2023ipc}
Chiheon Kim, Doyup Lee, Saehoon Kim, Minsu Cho, and Wook-Shin Han.
\newblock Generalizable implicit neural representations via instance pattern composers.
\newblock In \emph{CVPR}, 2023.

\bibitem[Kim et~al.(2024)Kim, Kim, and Han]{kim2024fedacg}
Geeho Kim, Jinkyu Kim, and Bohyung Han.
\newblock Communication-efficient federated learning with accelerated client gradient.
\newblock In \emph{CVPR}, 2024.

\bibitem[Kim et~al.(2022)Kim, Yu, Lee, and Shin]{kim2022scalable}
Subin Kim, Sihyun Yu, Jaeho Lee, and Jinwoo Shin.
\newblock Scalable neural video representations with learnable positional features.
\newblock \emph{NeurIPS}, 2022.

\bibitem[Lee et~al.(2024)Lee, Kim, Cho, and HAN]{lee2024locality}
Doyup Lee, Chiheon Kim, Minsu Cho, and WOOK~SHIN HAN.
\newblock Locality-aware generalizable implicit neural representation.
\newblock \emph{NeurIPS}, 2024.

\bibitem[Li et~al.(2020)Li, Khodak, Caldas, and Talwalkar]{jeffrey2020dpmeta}
Jeffrey Li, Mikhail Khodak, Sebastian Caldas, and Ameet Talwalkar.
\newblock Differentially private meta-learning.
\newblock In \emph{ICLR}, 2020.

\bibitem[Li et~al.(2022)Li, Diao, Chen, and He]{li2022fedsurvey}
Qinbin Li, Yiqun Diao, Quan Chen, and Bingsheng He.
\newblock Federated learning on non-iid data silos: An experimental study.
\newblock In \emph{IEEE International Conference on Data Engineering}, 2022.

\bibitem[Li et~al.(2023)Li, Wang, Narayan, Kirby, and Zhe]{li2023amaml}
Shibo Li, Zheng Wang, Akil Narayan, Robert Kirby, and Shandian Zhe.
\newblock Meta-learning with adjoint methods.
\newblock In \emph{AISTATS}, pages 7239--7251. PMLR, 2023.

\bibitem[Li et~al.(2019)Li, Huang, Yang, Wang, and Zhang]{fedprox}
Xiang Li, Kaixuan Huang, Wenhao Yang, Shusen Wang, and Zhihua Zhang.
\newblock On the convergence of fedavg on non-iid data.
\newblock \emph{arXiv:1907.02189}, 2019.

\bibitem[Li et~al.(2017)Li, Zhou, Chen, and Li]{li2017metasgd}
Zhenguo Li, Fengwei Zhou, Fei Chen, and Hang Li.
\newblock Meta-sgd: Learning to learn quickly for few-shot learning.
\newblock \emph{arXiv:1707.09835}, 2017.

\bibitem[Liang et~al.(2023)Liang, Huang, Tian, Kumar, and Xu]{liang2023avnerf}
Susan Liang, Chao Huang, Yapeng Tian, Anurag Kumar, and Chenliang Xu.
\newblock Av-nerf: Learning neural fields for real-world audio-visual scene synthesis.
\newblock \emph{NeurIPS}, 2023.

\bibitem[McMahan et~al.(2017)McMahan, Moore, Ramage, Hampson, and y~Arcas]{fedavg}
Brendan McMahan, Eider Moore, Daniel Ramage, Seth Hampson, and Blaise~Aguera y Arcas.
\newblock Communication-efficient learning of deep networks from decentralized data.
\newblock In \emph{AISTATS}, 2017.

\bibitem[McNally et~al.(2019)McNally, Vats, Pinto, Dulhanty, McPhee, and Wong]{mcnally2019golfdb}
William McNally, Kanav Vats, Tyler Pinto, Chris Dulhanty, John McPhee, and Alexander Wong.
\newblock Golfdb: A video database for golf swing sequencing.
\newblock In \emph{CVPRW}, 2019.

\bibitem[Mildenhall et~al.(2021)Mildenhall, Srinivasan, Tancik, Barron, Ramamoorthi, and Ng]{mildenhall2021nerf}
Ben Mildenhall, Pratul~P Srinivasan, Matthew Tancik, Jonathan~T Barron, Ravi Ramamoorthi, and Ren Ng.
\newblock Nerf: Representing scenes as neural radiance fields for view synthesis.
\newblock \emph{Communications of the ACM}, 2021.

\bibitem[Nichol et~al.(2018)Nichol, Achiam, and Schulman]{onfirst}
Alex Nichol, Joshua Achiam, and John Schulman.
\newblock On first-order meta-learning algorithms.
\newblock \emph{arXiv:1803.02999}, 2018.

\bibitem[Niemeyer et~al.(2020)Niemeyer, Mescheder, Oechsle, and Geiger]{niemeyer2020differentiable}
Michael Niemeyer, Lars Mescheder, Michael Oechsle, and Andreas Geiger.
\newblock Differentiable volumetric rendering: Learning implicit 3d representations without 3d supervision.
\newblock In \emph{CVPR}, 2020.

\bibitem[Park et~al.(2019)Park, Florence, Straub, Newcombe, and Lovegrove]{park2019deepsdf}
Jeong~Joon Park, Peter Florence, Julian Straub, Richard Newcombe, and Steven Lovegrove.
\newblock Deepsdf: Learning continuous signed distance functions for shape representation.
\newblock In \emph{CVPR}, 2019.

\bibitem[Saragadam et~al.(2022)Saragadam, Tan, Balakrishnan, Baraniuk, and Veeraraghavan]{saragadam2022miner}
Vishwanath Saragadam, Jasper Tan, Guha Balakrishnan, Richard~G Baraniuk, and Ashok Veeraraghavan.
\newblock Miner: Multiscale implicit neural representation.
\newblock In \emph{ECCV}, 2022.

\bibitem[Sen et~al.(2022)Sen, Agarwal, Namboodiri, and Jawahar]{sen2022vinr}
Bipasha Sen, Aditya Agarwal, Vinay~P Namboodiri, and CV Jawahar.
\newblock Inr-v: A continuous representation space for video-based generative tasks.
\newblock \emph{arXiv:2210.16579}, 2022.

\bibitem[Shinoda and Shiohara(2025)]{shinoda2025petface}
Risa Shinoda and Kaede Shiohara.
\newblock Petface: A large-scale dataset and benchmark for animal identification.
\newblock In \emph{ECCV}, 2025.

\bibitem[Shokri et~al.(2017)Shokri, Stronati, Song, and Shmatikov]{shokri2017mia}
Reza Shokri, Marco Stronati, Congzheng Song, and Vitaly Shmatikov.
\newblock Membership inference attacks against machine learning models.
\newblock In \emph{IEEE Symposium on Security and Privacy}, pages 3--18. IEEE, 2017.

\bibitem[Sitzmann et~al.(2020)Sitzmann, Martel, Bergman, Lindell, and Wetzstein]{sitzmann2020siren}
Vincent Sitzmann, Julien Martel, Alexander Bergman, David Lindell, and Gordon Wetzstein.
\newblock Implicit neural representations with periodic activation functions.
\newblock \emph{NeurIPS}, 2020.

\bibitem[Skorokhodov et~al.(2021)Skorokhodov, Ignatyev, and Elhoseiny]{skorokhodov2021adversarial}
Ivan Skorokhodov, Savva Ignatyev, and Mohamed Elhoseiny.
\newblock Adversarial generation of continuous images.
\newblock In \emph{CVPR}, 2021.

\bibitem[Su et~al.(2022)Su, Chen, and Shlizerman]{su2022inras}
Kun Su, Mingfei Chen, and Eli Shlizerman.
\newblock Inras: Implicit neural representation for audio scenes.
\newblock \emph{NeurIPS}, 2022.

\bibitem[Suzuki(2023)]{suzuki2023fednerflarge}
Teppei Suzuki.
\newblock Federated learning for large-scale scene modeling with neural radiance fields.
\newblock \emph{arXiv:2309.06030}, 2023.

\bibitem[Suzuki(2024)]{suzuki2024fed3dgs}
Teppei Suzuki.
\newblock Fed3dgs: Scalable 3d gaussian splatting with federated learning.
\newblock \emph{arXiv:2403.11460}, 2024.

\bibitem[Tack et~al.(2024)Tack, Kim, Yu, Lee, Shin, and Schwarz]{tack2024gradncp}
Jihoon Tack, Subin Kim, Sihyun Yu, Jaeho Lee, Jinwoo Shin, and Jonathan~Richard Schwarz.
\newblock Learning large-scale neural fields via context pruned meta-learning.
\newblock \emph{NeurIPS}, 2024.

\bibitem[Tancik et~al.(2020)Tancik, Srinivasan, Mildenhall, Fridovich-Keil, Raghavan, Singhal, Ramamoorthi, Barron, and Ng]{tancik2020fourier}
Matthew Tancik, Pratul Srinivasan, Ben Mildenhall, Sara Fridovich-Keil, Nithin Raghavan, Utkarsh Singhal, Ravi Ramamoorthi, Jonathan Barron, and Ren Ng.
\newblock Fourier features let networks learn high frequency functions in low dimensional domains.
\newblock \emph{NeurIPS}, 2020.

\bibitem[Tancik et~al.(2021)Tancik, Mildenhall, Wang, Schmidt, Srinivasan, Barron, and Ng]{tancik2021learnedit}
Matthew Tancik, Ben Mildenhall, Terrance Wang, Divi Schmidt, Pratul~P Srinivasan, Jonathan~T Barron, and Ren Ng.
\newblock Learned initializations for optimizing coordinate-based neural representations.
\newblock In \emph{CVPR}, 2021.

\bibitem[Tasneem et~al.(2024)Tasneem, Dave, Singh, Tiwary, Vepakomma, Veeraraghavan, and Raskar]{tasneem2024decentnerfs}
Zaid Tasneem, Akshat Dave, Abhishek Singh, Kushagra Tiwary, Praneeth Vepakomma, Ashok Veeraraghavan, and Ramesh Raskar.
\newblock Decentnerfs: Decentralized neural radiance fields from crowdsourced images.
\newblock \emph{arXiv:2403.13199}, 2024.

\bibitem[Tewari et~al.(2020)Tewari, Fried, Thies, Sitzmann, Lombardi, Sunkavalli, Martin-Brualla, Simon, Saragih, Nie{\ss}ner, et~al.]{tewari2020state}
Ayush Tewari, Ohad Fried, Justus Thies, Vincent Sitzmann, Stephen Lombardi, Kalyan Sunkavalli, Ricardo Martin-Brualla, Tomas Simon, Jason Saragih, Matthias Nie{\ss}ner, et~al.
\newblock State of the art on neural rendering.
\newblock In \emph{Computer Graphics Forum}, 2020.

\bibitem[Wang et~al.(2020)Wang, Liu, Liang, Joshi, and Poor]{wang2020fednova}
Jianyu Wang, Qinghua Liu, Hao Liang, Gauri Joshi, and H~Vincent Poor.
\newblock Tackling the objective inconsistency problem in heterogeneous federated optimization.
\newblock \emph{NeurIPS}, 2020.

\bibitem[Wang et~al.(2004)Wang, Bovik, Sheikh, and Simoncelli]{wang2004ssim}
Zhou Wang, Alan~C Bovik, Hamid~R Sheikh, and Eero~P Simoncelli.
\newblock Image quality assessment: from error visibility to structural similarity.
\newblock \emph{IEEE Transactions on Image Processing}, 2004.

\bibitem[Wei et~al.(2020)Wei, Li, Ding, Ma, Yang, Farokhi, Jin, Quek, and Vincent~Poor]{wei2020federateddp}
Kang Wei, Jun Li, Ming Ding, Chuan Ma, Howard~H. Yang, Farhad Farokhi, Shi Jin, Tony Q.~S. Quek, and H. Vincent~Poor.
\newblock Federated learning with differential privacy: Algorithms and performance analysis.
\newblock \emph{IEEE Transactions on Information Forensics and Security}, 2020.

\bibitem[Xie et~al.(2022)Xie, Wang, Gao, Chen, Yao, Kuang, Li, Ding, and Zhou]{xie2022federatedscope}
Yuexiang Xie, Zhen Wang, Dawei Gao, Daoyuan Chen, Liuyi Yao, Weirui Kuang, Yaliang Li, Bolin Ding, and Jingren Zhou.
\newblock Federatedscope: A flexible federated learning platform for heterogeneity.
\newblock \emph{arXiv:2204.05011}, 2022.

\bibitem[Yang et~al.(2020)Yang, Zhu, Wang, Huang, Shen, Yang, and Cao]{yang2020facescape}
Haotian Yang, Hao Zhu, Yanru Wang, Mingkai Huang, Qiu Shen, Ruigang Yang, and Xun Cao.
\newblock Facescape: a large-scale high quality 3d face dataset and detailed riggable 3d face prediction.
\newblock In \emph{CVPR}, 2020.

\bibitem[Yue et~al.(2023)Yue, Jin, Wong, Baron, and Dai]{yue2023gradient}
Kai Yue, Richeng Jin, Chau-Wai Wong, Dror Baron, and Huaiyu Dai.
\newblock Gradient obfuscation gives a false sense of security in federated learning.
\newblock In \emph{USENIX Security Symposium}, 2023.

\bibitem[Zhang et~al.(2018)Zhang, Isola, Efros, Shechtman, and Wang]{zhang2018lpips}
Richard Zhang, Phillip Isola, Alexei~A Efros, Eli Shechtman, and Oliver Wang.
\newblock The unreasonable effectiveness of deep features as a perceptual metric.
\newblock In \emph{CVPR}, 2018.

\bibitem[Zhang and Shao(2024)]{zhang2024fednerf2}
Yintian Zhang and Ziyu Shao.
\newblock Federated neural radiance field for distributed intelligence.
\newblock \emph{arXiv:2406.10474}, 2024.

\bibitem[Zhou and Bassily(2022)]{zhou2022metansgd}
Xinyu Zhou and Raef Bassily.
\newblock Task-level differentially private meta learning.
\newblock \emph{NeurIPS}, 2022.

\bibitem[Zhu et~al.(2023)Zhu, Yang, Guo, Zhang, Wang, Huang, Wu, Shen, Yang, and Cao]{zhu2023facescape}
Hao Zhu, Haotian Yang, Longwei Guo, Yidi Zhang, Yanru Wang, Mingkai Huang, Menghua Wu, Qiu Shen, Ruigang Yang, and Xun Cao.
\newblock Facescape: 3d facial dataset and benchmark for single-view 3d face reconstruction.
\newblock \emph{IEEE TPAMI}, 2023.

\bibitem[Zhuang et~al.(2022)Zhuang, Zhu, Sun, and Cao]{zhuang2022mofanerf}
Yiyu Zhuang, Hao Zhu, Xusen Sun, and Xun Cao.
\newblock Mofanerf: Morphable facial neural radiance field.
\newblock In \emph{ECCV}, 2022.

\end{thebibliography}
}

\appendix
\clearpage
\setcounter{page}{1}
\maketitlesupplementary

\section{Notations}
\label{sec:notation}

\begin{table}[ht]
    \centering
    \small
        \begin{tabular}{c|p{0.7\linewidth}}
            \toprule
            \multicolumn{1}{c|}{Notation} &  \multicolumn{1}{c}{Description}\\
            \hline
            $N$ & Number of clients \\
            $M$ & Number of participating clients in each communication round \\
            $R$ & Number of communication rounds \\
            $E$ & Number of outer loop steps \\
            $K$ & Number of inner loop steps \\
            $\alpha^m$ & Weight proportional to the client $m$'s dataset size (see \cref{eq:fedavg_objevtive}) \\
            $\lambda_{i}$ & Inner loop learning rate \\
            $\lambda_{o}$ & Outer loop learning rate \\
            $T$ & Data within a task, \ie, an image / a video / images and camera poses of a 3D object. \\
            $S$ & Support set of the task data \\
            $Q$ & Query set of the task data \\
            $B$ & Minibatch of the data ($B_K$ is sampled from the query set $Q$, while the rest is sampled from the support set $S$) \\
            $\theta$ & Global model (meta-learner) parameters \\
            $w$ & Local model (meta-learner) parameters \\
            $\varphi$ & Neural field parameters\\
            $D_{\text{train}}$ & Client's local training dataset\\
            $\gamma$ & Regularization coefficient\\
            \bottomrule
        \end{tabular}
    \caption{
        Notations.
    }
    \label{tab:notation}
\end{table}

\section{Related Work}

\subsection{Federated Meta-Learning}
Traditional federated learning algorithms assume that decentralized local datasets all belong to the same task and aim to train a single global model.
However, when data is heterogeneous, or clients require different tasks, a single global model often fails to perform well for all clients.  

To address this challenge, federated meta-learning introduces the meta-learning approach to achieve personalization.
Federated meta-learning combines federated learning and meta-learning to train a global meta-learner using data from multiple clients.
For example, \cite{fedmeta} proposed a method that combines the traditional FedAvg~\cite{fedavg} algorithm with MAML~\cite{finn2017maml} and Meta-SGD~\cite{li2017metasgd}.
Similarly, \cite{fallah202perfedavg} introduced an approach that integrates FedAvg~\cite{fedavg} with HF-MAML~\cite{fallah2020hfmaml}.  

In our work, we constructed baselines by combining not only FedAvg~\cite{fedavg} but also state-of-the-art federated learning algorithms such as FedProx~\cite{fedprox}, Scaffold~\cite{scaffold}, FedNova~\cite{wang2020fednova}, FedExP~\cite{fedexp}, FedACG~\cite{kim2024fedacg} with meta-learning algorithms like MAML~\cite{finn2017maml}, FOMAML~\cite{finn2017maml}, Reptile~\cite{onfirst}, and meta-NSGD~\cite{zhou2022metansgd}.
We then evaluated the performance of our FedMeNF against these baselines. 

\subsection{Meta-Learning for Neural Fields}

Neural fields typically require training a separate neural network for each task (signal) and demand large amounts of data and computation. Additionally, the training process often converges slowly, which is a significant limitation~\cite{gu2023generalizable,tancik2021learnedit}. To address these challenges, recent research has focused on learning a generalized neural field that can handle multiple tasks efficiently.  

For example, \cite{tancik2021learnedit} used MAML~\cite{finn2017maml} and Reptile~\cite{onfirst} to train a meta-initialized neural field network. Instead of training neural fields for each task from scratch, they optimize the neural fields for each task from a meta-initialized neural field. This approach accelerates neural field optimization and achieves high performance even in few-shot scenarios.
Inspired by this, we include MAML~\cite{finn2017maml}, FOMAML~\cite{finn2017maml}, and Reptile~\cite{onfirst} as baselines and evaluate our method against them.
\cite{dupont2022functa} extended the work of \cite{tancik2021learnedit} by introducing task-specific modulation vectors.
During the inner loop, only the modulation vector is updated, while in the outer loop, the base network is updated separately from the modulation vector.
This method reduces memory requirements by storing a modulation vector for each task with a single base network, rather than all neural field parameters.
While our work focuses on achieving fast optimization and high performance with limited data, we did not include modulation-based meta-learning methods as baselines because they do not align directly with our objectives.  

Recent research has also explored hypernetwork-based approaches for training generalized neural fields.
\cite{chen2022transinr,kim2023ipc} employed transformers as hypernetworks to predict neural field parameters from task data, while \cite{gu2023generalizable} extended this approach using neural processes.
These hypernetwork-based methods have shown superior generalization compared to gradient-based meta-learning approaches like \cite{tancik2021learnedit}.
However, they come with the drawback of requiring much larger models.
Such large models are impractical in a federated learning setting where models are frequently communicated between the server and clients at every communication round.
For this reason, we excluded hypernetwork-based methods from our baselines.
In future work, we plan to explore ways to adapt these high-performing hypernetwork-based methods to federated learning environments, making them more efficient and scalable.

\subsection{Federated Learning for Neural Fields}
\label{sec:privacy_leakage_fl}

Federated learning methods for neural fields, such as those proposed in \cite{holden2023fednerf,zhang2024fednerf2,suzuki2024fed3dgs,suzuki2023fednerflarge}, train a global neural field using local data from multiple clients without sharing the raw data.
These methods aim to train a global neural field for a single scene collaboratively.
This approach differs from ours, which focuses on training a global meta-learner that can quickly adapt to diverse tasks with minimal data. 

In this scenario, all clients have a subset of the same task data.
For instance, if multiple clients have images of the same car, these methods aim to train a global neural field for that car using a federated learning approach.
The trained global neural field is shared with all clients, including the server, and can be used to render images from new viewpoints.
However, if the server or other clients render images using the same camera poses as those in a client's private dataset, they can reconstruct images that are very similar to the client's private images.
This violates the core principle of data privacy, which is central to federated learning.

We experimentally demonstrate this privacy leakage.
Using FedNeRF~\cite{holden2023fednerf}, we trained a global neural field for the Lego scene~\cite{mildenhall2021nerf} with varying numbers of total clients: 5, 10, and 50.
Our experimental setup follows \cite{holden2023fednerf} with two key differences: only five clients participated in training per round, and we assume the server has no access to any data and does not pre-train an initial NeRF.
The dataset includes 100 training images with a resolution of $400 \times 400$ evenly distributed among clients and 200 test images.

\begin{figure}[!h]
    \centering
    \captionsetup[subfigure]{font=scriptsize}
    \begin{subfigure}[b]{0.23\textwidth}
        \centering
        \includegraphics[width=\linewidth]{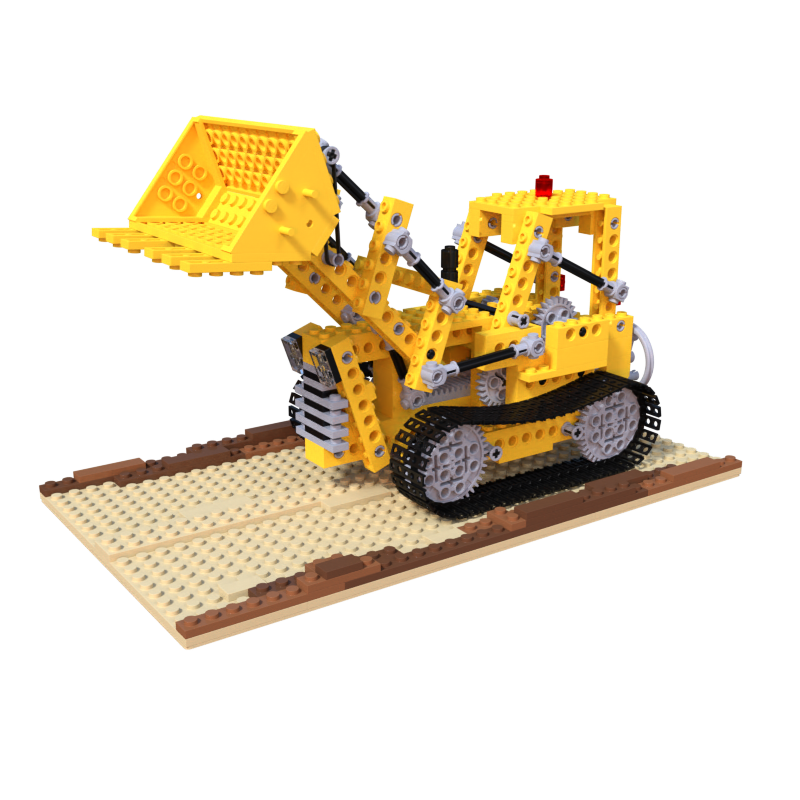}
        \vspace{-2.5em}
        \subcaption{Client's training image ($\text{PSNR}_p$)}
    \end{subfigure}
    \hfill
    \begin{subfigure}[b]{0.23\textwidth}
        \centering
        \includegraphics[width=\linewidth]{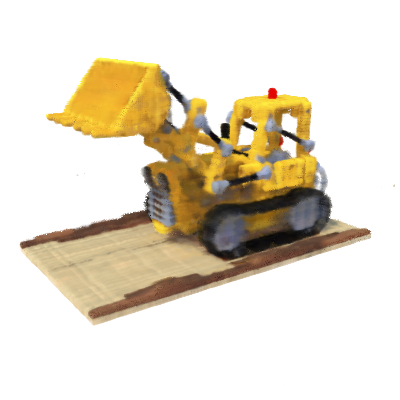}
        \vspace{-2.5em}
        \subcaption{Server-reconstructed image (23.46)}
    \end{subfigure}

    \caption{
        Qualitative results of reconstructing a client's private image on the server (\# of clients = 5).
        A significant privacy leak arises as the server can reconstruct images that are highly similar to the client's original images.
    }
    \label{fig:fednerf_qualitative_train}
\end{figure}

\begin{figure}[!h]
    \centering
    \captionsetup[subfigure]{font=scriptsize}
    \begin{subfigure}[b]{0.23\textwidth}
        \centering
        \includegraphics[width=\linewidth]{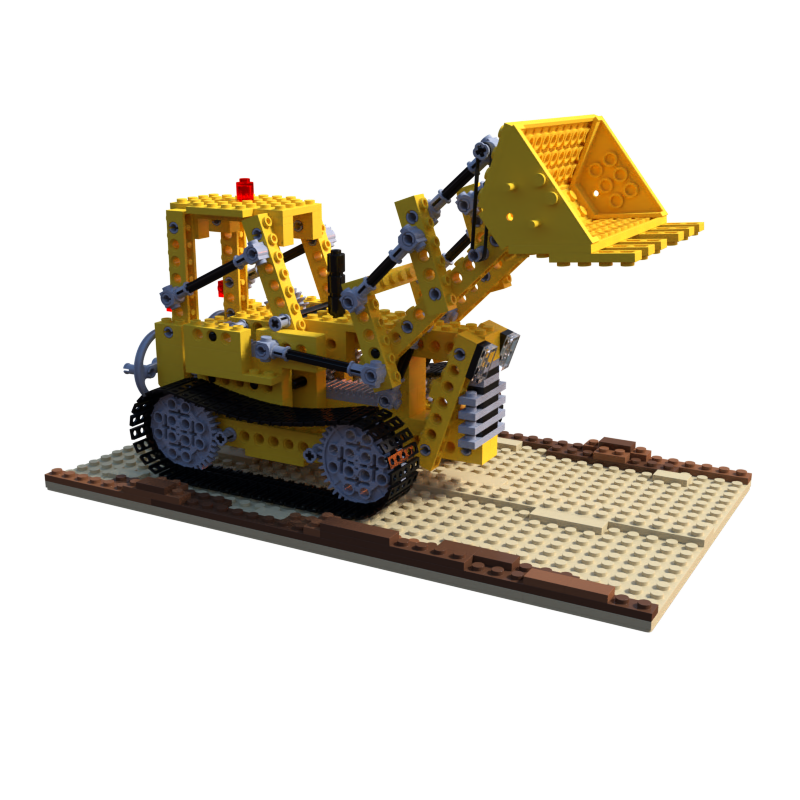}
        \vspace{-2.5em}
        \subcaption{GT image of novel view (PSNR)}
    \end{subfigure}
    \hfill
    \begin{subfigure}[b]{0.23\textwidth}
        \centering
        \includegraphics[width=\linewidth]{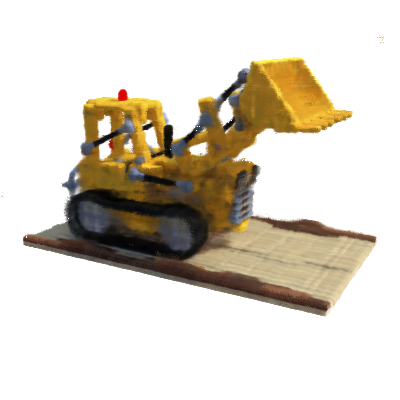}
        \vspace{-2.5em}
        \subcaption{Rendered image (23.53)}
    \end{subfigure}

    \caption{
        Qualitative results of novel view synthesis performed on the client with the trained global neural field (\# of clients = 5).
    }
    \label{fig:fednerf_qualitative_test}
\end{figure}

As shown in \cref{tab:fednerf,fig:fednerf_qualitative_train,fig:fednerf_qualitative_test}, the trained global neural field can render images that are highly similar to the private images held by clients.
These results show that we cannot train a global neural field for a single scene in a federated learning setup, which inherently carries a risk of privacy leakage.
This approach should only be used with non-private data, such as public scenes.

\begin{table}[!ht]
    \setlength{\tabcolsep}{1.5pt}
    \centering
    \small
    \resizebox{\linewidth}{!}{

        \begin{tabular}{ccccccc}
            \toprule
            \multicolumn{1}{c}{\bf{Dataset}} & \multicolumn{6}{c}{Lego~\cite{mildenhall2021nerf}} \\
            \cmidrule(lr){2-3} \cmidrule(lr){4-5} \cmidrule(lr){6-7}
            \multicolumn{1}{c}{\bf{\# of clients}} 
            & \multicolumn{1}{c}{$\text{PSNR}_p$(↓)} & \multicolumn{1}{c}{PSNR(↑)}
            & \multicolumn{1}{c}{$\text{SSIM}_p$(↓)} & \multicolumn{1}{c}{SSIM(↑)}
            & \multicolumn{1}{c}{$\text{LPIPS}_p$(↑)} & \multicolumn{1}{c}{LPIPS(↓)}
            \\
            
            \midrule
            5  &     24.00  & \bf{23.59} &     0.829  & \bf{0.827} &     0.173  & \bf{0.172} \\
            10 &     23.66  &     23.40  &     0.823  &     0.823  &     0.177  &     0.176  \\
            50 & \bf{22.48} &     22.03  & \bf{0.808} &     0.804  & \bf{0.188} &     0.191  \\
            \bottomrule
        \end{tabular}
    }
    \caption{
        Results of various reconstruction quality metrics (PSNR, SSIM, LPIPS) and privacy metrics ($\text{PSNR}_p$, $\text{SSIM}_p$, $\text{LPIPS}_p$) on the Lego~\cite{mildenhall2021nerf} dataset.
    }
    \label{tab:fednerf}
\end{table}

\section{Privacy Attack Scenarios}
\label{sec:suppl:privacy_attack_scenarios}

\subsection{Membership Inference Attack}
The objective of the Membership Inference Attack (MIA)~\cite{shokri2017mia} is for the server to determine whether a given data sample was in the training set of a target client's local meta-learner.
Therefore, a successful attack would allow the server to identify the owner of a car in the data sample.

\noindent \textbf{Setting.}Our attacker model is implemented as a simple convolutional network consisting of two convolution layers and two fully connected layers.
This model takes a pair of images as input: a target image and a synthesized image.
The synthesized image is rendered by a specific local meta-learner using the same view as the target image.
The attacker's task is to infer whether the target image belongs to the training set of the meta-learner used for synthesis.
The attacker model outputs a binary prediction indicating whether the target image was included in the training set of the local meta-learner used to generate the corresponding synthesized image.

\noindent \textbf{Dataset.} To train and evaluate the attacker, we partition the 50 clients into a group of 40 shadow clients and a held-out group of 10 target clients.
To construct the dataset for training and evaluating our attacker model, we partition the 50 clients into 40 shadow clients and 10 target clients. Each client holds a training set of four images of a unique car, totaling 200 unique target images (50 clients $\times$ 4 images/client).
The 160 images from the shadow clients are used to generate the attacker's training set, while the 40 images from the target clients form the test set.
For each target image, we generate a positive sample (membership label = 1) by pairing it with the image synthesized by its owner's local meta-learner.
A negative sample (membership label = 0) is created by pairing the same target image with an image synthesized by a randomly selected client's local meta-learner within the same partition. 
This process results in a balanced dataset for the attacker model, comprising 320 training samples and 80 test samples.

\noindent \textbf{Evaluation.} We assess the attacker model's accuracy in correctly predicting the membership labels.
Higher accuracy signifies a more effective attack and, consequently, a more significant privacy vulnerability.

\subsection{Property Inference Attack}
The objective of the Property Inference Attack (PIA)~\cite{ganju2018pia} is for the server to infer a global property of a local meta-learner's training set.
Specifically, a successful attack allows the server to determine the vehicle category (e.g., Sedan, SUV, or Coupe) of the car that each client owns.

\noindent \textbf{Setting.} The attacker model for the PIA is a classifier, implemented with a simple convolutional network architecture similar to that for MIA.
It is trained to take a synthesized image from an arbitrary client's meta-learner as input and predict the vehicle category of that client's car.

\begin{figure}[ht!]
    \centering
    \includegraphics[width=\linewidth]{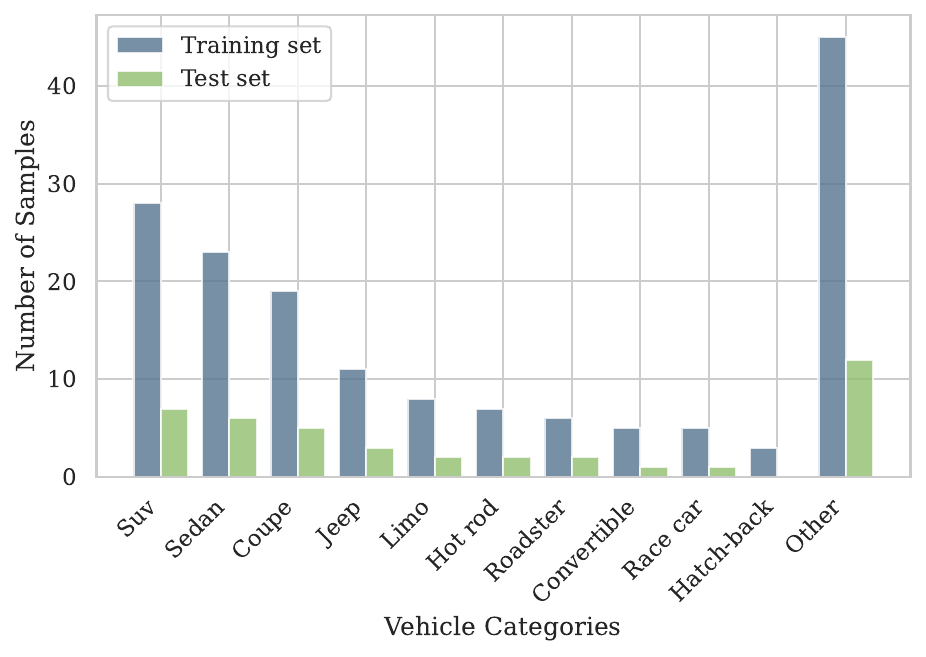}
    \caption{Distribution of the training and test sets for the Privacy Inversion Attack.}
    \label{fig:dist_pia} 
\end{figure}

\noindent \textbf{Dataset.} Similar to the MIA setup, we construct the PIA dataset from the 200 unique target images provided by all clients.
For each target image, we use its owner client's local meta-learner to synthesize a new image from the same viewpoint.
The synthesized image is then labeled with the vehicle category of the client's car, constructing a single data sample for the attacker model.
As a result, we obtain a total of 200 samples: 160 samples derived from the shadow clients, which form the training set, and 40 samples from the target clients, which form the test set.
The distributions of the training and test sets are shown in \cref{fig:dist_pia}. 

\noindent \textbf{Evaluation.} We evaluate the attacker model's classification accuracy on the vehicle categories of the target clients' cars.
High accuracy implies that the local meta-learner leaks information about the private properties (i.e., vehicle category) of a client's training data.

\section{Implementation Details}
\label{sec:implementation_details}
\textbf{Models.} To maintain consistency with previous research~\cite{dupont2022functa,tack2024gradncp,tancik2021learnedit,chen2022transinr}, we employ the widely adopted SIREN~\cite{sitzmann2020siren} architecture for image and video reconstruction and the simplified NeRF model~\cite{mildenhall2021nerf} for novel view synthesis. The SIREN model consists of 6 layers with a hidden dimension of 128. The NeRF model consists of 6 layers with a hidden dimension of 256. We apply gradient clipping to the gradients (max norm value = 5).
All experiments are run on a cluster of 64 NVIDIA TITAN RTX GPUs.
Our code will be publicly available upon publication. 

\noindent \textbf{Hyperparameters.} \Cref{tab:hyperparameters} summarizes the hyperparameters used in our experiments.

\section{Datasets}
\label{sec:dataset_suppl}
We conduct experiments across the datasets of various modalities, including images, videos, and NeRF.
To evaluate performance with limited local training data per client, we select scenarios in which each client has only one or very few tasks.
For images, we use the cat category from the PetFace dataset~\cite{shinoda2025petface}, assuming each client has an average of 3.12 images of a unique cat instance.
For videos, we use the GolfDB dataset~\cite{mcnally2019golfdb}, where each client has an average of 1.56 videos of only one person's golf swings.
Since a task is defined as reconstructing a single image or video, each image or video corresponds to one task.
Note that a client has images or videos of an individual, and the number of images or videos determines the number of tasks.

However, the definition of a task is slightly different for NeRF scenes.
A task is defined as synthesizing a novel view of a 3D object, where the input views for test-time optimization (support set) and the new views for testing (query set) constitute one task.
We use the Cars category from the ShapeNet dataset~\cite{chang2015shapenet}.
The training set for each client contains on average four input views of a single car with an equivalent amount allocated for testing.
We also test the FaceScape dataset~\cite{zhu2023facescape,yang2020facescape} for human faces. Each client has an average of 10 training input views for a single facial expression and the same number of views for testing.
Each client has one task (one 3D object) for NeRF scenes, and the number of input views follows a Dirichlet distribution.

\subsection{Petface}
\begin{figure}[ht!]
    \centering
    \includegraphics[width=\linewidth]{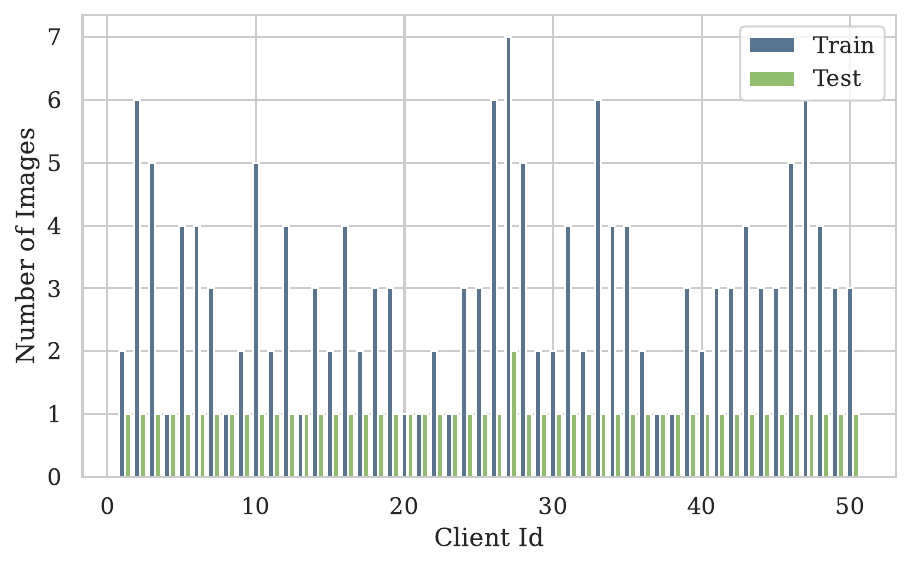}
    \caption{Distribution of the number of images for each client in the PetFace~\cite{shinoda2025petface} dataset.}
    \label{fig:dist_petface} 
\end{figure}

We use the PetFace~\cite{shinoda2025petface} dataset that contains animal face images of 257484 unique individuals across 13 animal families and 319 breed categories.
In this work, we focus exclusively on the cat category within the dataset.
We assume each client represents an individual pet owner and possesses images of a unique pet.
The total number of training images across all clients is 156, and the total number of test images is 51.
On average, each client has 3.12 training images and 1.02 test images.
All images have a resolution of 224 × 224.
The number of images available for each pet (client) varies, as shown in \cref{fig:dist_petface}.
The client's local training dataset is used in a federated meta-learning framework to collaboratively train a global meta-learner.
Once the global meta-learner is trained, each client can use it to rapidly optimize the neural fields of new images, whether of their own pet, another client's pet, or out-of-distribution (OOD) pets.

\subsection{GolfDB}
\begin{figure}[ht!]
    \centering
    \includegraphics[width=\linewidth]{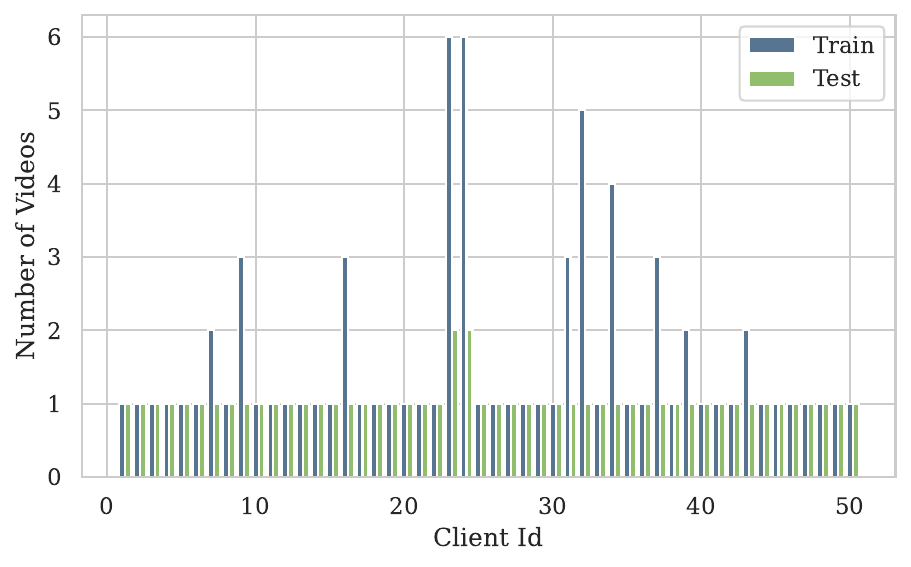}
    \caption{Distribution of the number of videos for each client in the GolfDB~\cite{mcnally2019golfdb} dataset.}
    \label{fig:dist_golfdb} 
\end{figure}

GolfDB~\cite{mcnally2019golfdb} is a video dataset that includes 1400 golf swing videos of professional golfers.
Each video is categorized into one of three view types: face-on, down-the-line, or other. We use only the down-the-line videos with a resolution of 160 x 160 at 30 fps.
We assume that each client represents an individual golfer and possesses only their own golf swing videos.
Across all clients, the total number of training videos is 78, and the total number of test videos is 52. On average, each client has 1.56 training videos and 1.04 test videos. The number of videos available for each client is shown in \cref{fig:dist_golfdb}.
Each client can use the trained global meta-learner to rapidly optimize the neural field for a new golf swing video, regardless of who performed the swing.

\subsection{ShapeNet}
\label{sec:suppl:dataset_shapenet}
We conducted experiments on ShapeNet~\cite{chang2015shapenet}, which is broadly used for Neural Radiance Fields (NeRF). NeRF~\cite{mildenhall2021nerf} is a method for rendering novel views of a 3D scene by predicting the color and density at the arbitrary 3D location and 2-dimensional viewing directions through volume rendering. This approach enables the synthesis of images from new viewpoints, a task referred to as novel view synthesis. In our experiments, we focused on private and personal objects that people typically possess in limited varieties, such as cars. From ShapeNet, we randomly selected a subset of 100 car objects, each represented by 50 images of $128 \times 128$ resolution along with their corresponding camera poses. The 100 car objects are allocated to 50 clients, with each client assigned a total of 2 car objects; one for the local training dataset to train a global meta-learner, and one for the local test dataset to evaluate novel view synthesis performance.

\begin{figure}[ht]
    \centering
    \begin{subfigure}[b]{0.47\textwidth}
        \centering
        \includegraphics[width=\linewidth]{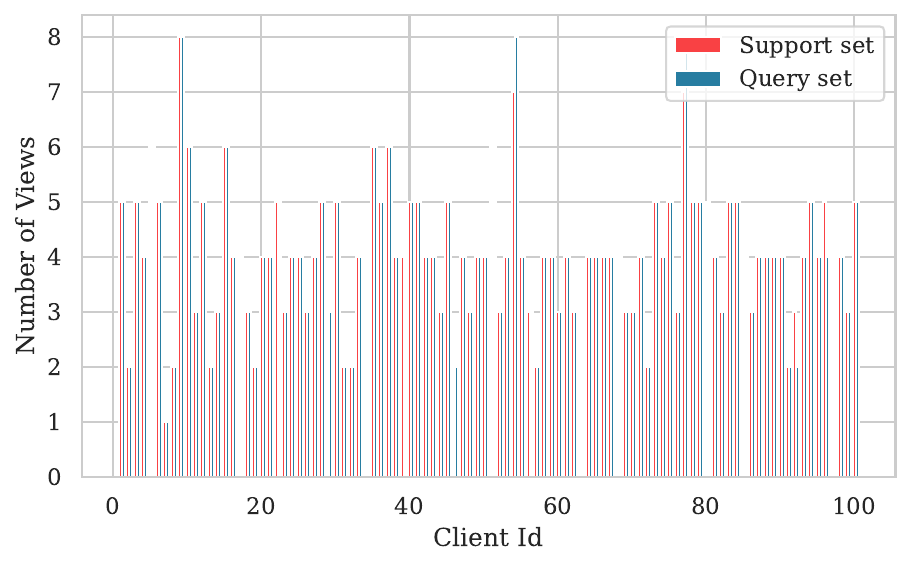}
        \subcaption{$\alpha = 10$}
    \end{subfigure}
    \vfill
    \begin{subfigure}[b]{0.47\textwidth}
        \centering
        \includegraphics[width=\linewidth]{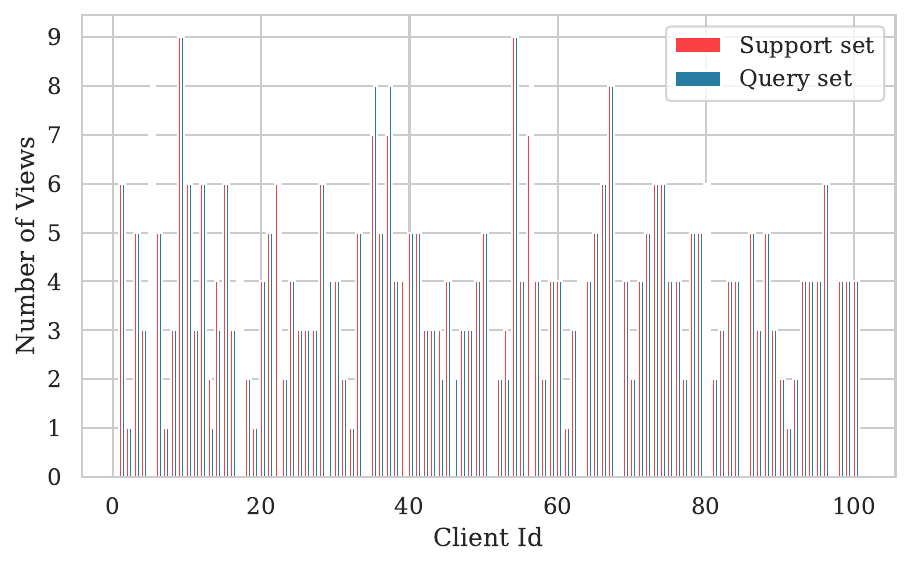}
        \subcaption{$\alpha = 5.0$}
    \end{subfigure}
    \vfill
    \begin{subfigure}[b]{0.47\textwidth}
        \centering
        \includegraphics[width=\linewidth]{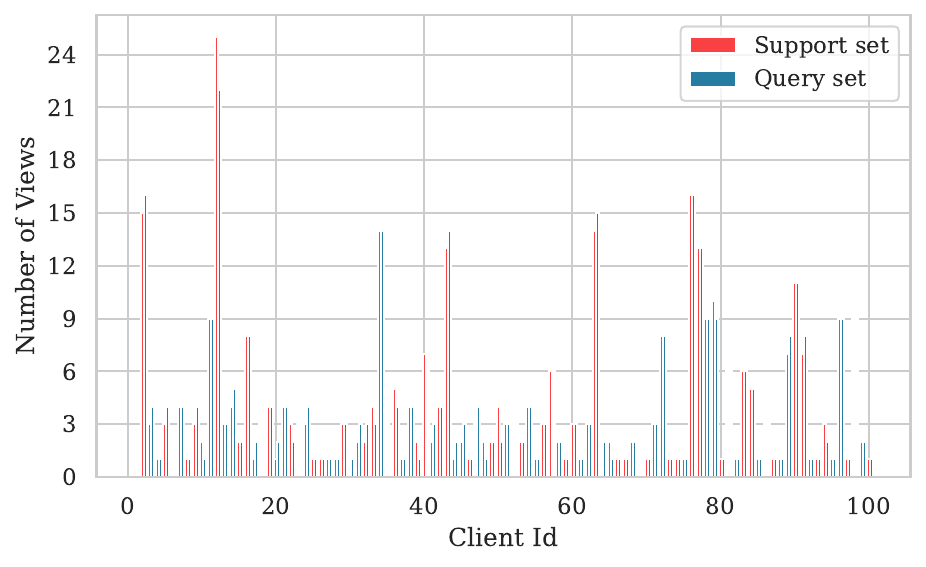}
        \subcaption{$\alpha = 1.0$}
    \end{subfigure}

    \caption{Distribution of the number of views for each car in the Cars~\cite{chang2015shapenet} dataset under various Dirichlet parameters.}
    \label{fig:dist_cars}
\end{figure}

\begin{figure}[ht]
    \centering
    \begin{subfigure}[b]{0.47\textwidth}
        \centering
        \includegraphics[width=\linewidth]{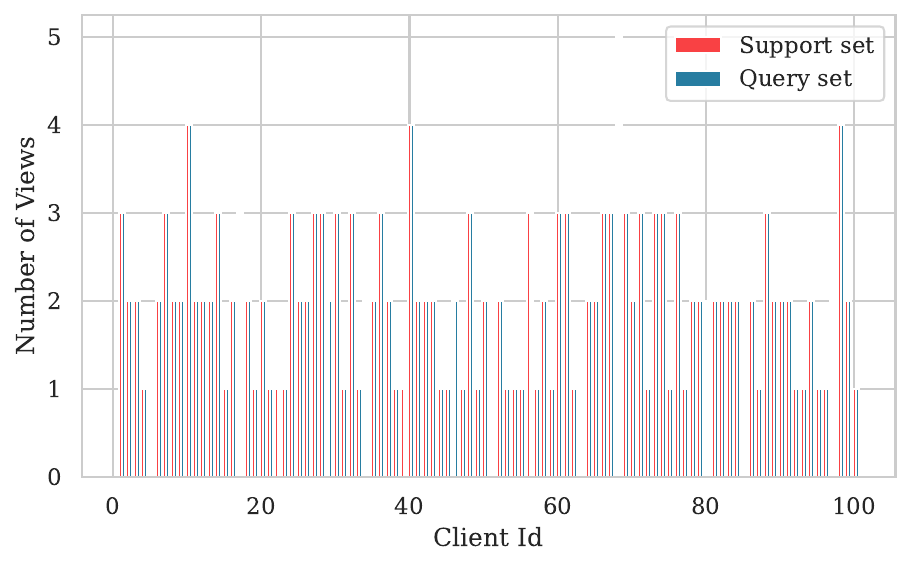}
        \subcaption{Avg. number of views = 2}
    \end{subfigure}
    \vfill
    \begin{subfigure}[b]{0.47\textwidth}
        \centering
        \includegraphics[width=\linewidth]{assets/dist_cars_5.pdf}
        \subcaption{Avg. number of views = 4}
    \end{subfigure}
    \vfill
    \begin{subfigure}[b]{0.47\textwidth}
        \centering
        \includegraphics[width=\linewidth]{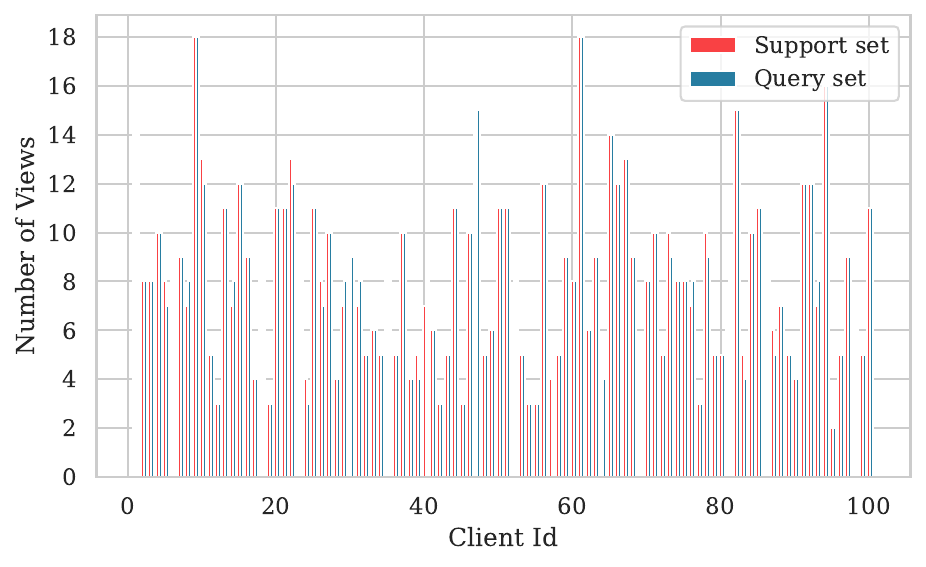}
        \subcaption{Avg. number of views = 8}
    \end{subfigure}

    \caption{Distribution of the number of views for each car in the Cars~\cite{chang2015shapenet} dataset under various avg. number of views ($\alpha = 5.0$).}
    \label{fig:dist_cars_few_views}
\end{figure}

To mimic real-world scenarios where the number of available views per object is limited and varies, we restricted the input views for each car object using a Dirichlet distribution. Each car's data is divided into two parts. The support set is used for inner-loop optimization during training and test-time optimization during testing. The query set is used for outer-loop optimization during training and to evaluate the performance of novel view synthesis during testing.
On average, each car object has four support images and 4 query images. Across all 100 car objects, this results in 400 support images distributed among clients. Using the Dirichlet distribution~\cite{chen2022pflbench,xie2022federatedscope}, we redistributed the number of support images to simulate varying levels of heterogeneity, ensuring the size of the query set is proportional to the support set.
To evaluate the generalization ability of our FedMeNF in non-identically distributed (non-IID) settings, we conduct experiments with different Dirichlet parameters ($\alpha = 10, 5.0, 1.0$). As $\alpha$ decreases, the distribution of support and query set sizes among clients becomes more heterogeneous, as illustrated in \cref{fig:dist_cars}.
Additionally, we test scenarios where the average number of support and query images per car is not fixed at 4. For example, we experimented with average sizes of 2 views and 8 views per car to examine FedMeNF's performance in different scenarios. The distributions of support and query set sizes for these settings are shown in \cref{fig:dist_cars_few_views}.
This comprehensive experimental setup allows us to evaluate FedMeNF's performance and robustness across a wide range of realistic, heterogeneous, and resource-constrained scenarios. The trained global meta-learner enables the rapid optimization of a neural field for any new car object even with a few images, allowing for the synthesis of high-quality novel views of that car.

\subsection{FaceScape}
We extended our evaluation to human faces, a critical area for privacy-preserving applications, using the FaceScape~\cite{zhu2023facescape,yang2020facescape} dataset, a more private and unique data type. This experiment aims to test the ability of our global meta-learner to handle diverse and privacy-sensitive datasets.
The FaceScape dataset contains 3D facial data from 847 subjects, each with 20 expressions. These expressions include emotional states, such as neutral, smile, and anger, and dynamic facial movements, such as blinking or raising eyebrows. There are 120 images for each expression with a resolution of 512 x 512 and corresponding camera poses.

In our work, we randomly select 50 subjects from the dataset and treat each subject as a separate client. Each client is assigned data for two expressions, averaging 20 images per expression at a downsampled resolution of 128 x 128. Data for one expression is used for training, while the other, considered an unseen expression, is employed for testing.

\begin{figure}[ht!]
    \centering
    \includegraphics[width=\linewidth]{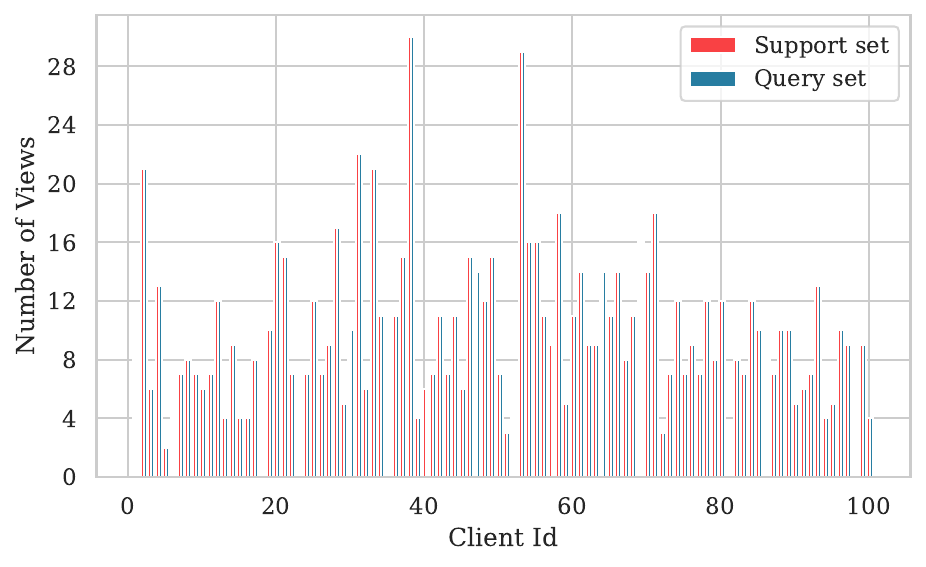}
    \caption{Distribution of the number of views for each expression in the FaceScape~\cite{chen2022pflbench,xie2022federatedscope} dataset ($\alpha = 5.0$).}
    \label{fig:dist_facescape} 
\end{figure}

For each expression, we assume an average of 10 support images used for inner-loop optimization during training or test-time optimization during testing. The number of support images for each expression, a total of $100 \text{ (clients)} \times 10 \text{ (images/client)} = 1000$ images across all expressions, is reallocated using a Dirichlet distribution~\cite{chen2022pflbench,xie2022federatedscope}. Similarly, we assign an average of 10 query images per expression for outer-loop optimization during training or novel view synthesis evaluation during testing, maintaining the same ratio as the number of support images. The support and query images distribution across expressions is shown in \cref{fig:dist_facescape}.

Using the trained global meta-learner, each client can quickly optimize a neural field for a new expression or even an entirely new person with only a few input images. This allows for synthesizing novel views of the face with the desired expression. This capability has broad applications: in AR/VR, it can enable the creation of realistic avatars that dynamically mimic user expressions. In 3D animation, it simplifies the production of dynamic facial models. Personal digital assistants can also use this approach to deliver more engaging and personalized facial interactions.

\begin{table*}[t]
    \centering
    \small
    \resizebox{\linewidth}{!}{
        \begin{tabular}{l|cccc}
            \toprule
            \multicolumn{1}{c|}{Hyperparameter} &  \multicolumn{1}{c}{PetFace~\cite{shinoda2025petface}} &  \multicolumn{1}{c}{GolfDB~\cite{mcnally2019golfdb}} &  \multicolumn{1}{c}{Cars~\cite{chang2015shapenet}} &  \multicolumn{1}{c}{FaceScape~\cite{zhu2023facescape,yang2020facescape}}\\
            \hline

            Number of communication rounds ($R$)& \multicolumn{4}{c}{1000} \\
            
            Number of clients ($N$) & \multicolumn{4}{c}{50} \\
            Number of participating clients in each communication round ($M$) & \multicolumn{4}{c}{5}\\

            FedProx~\cite{fedprox} $\mu$  & 0.1 & 0.1 & \multicolumn{2}{c}{0.1 (Reptile: 0.0001)} \\
            FedExP~\cite{fedexp} $\epsilon$ & \multicolumn{4}{c}{0.001} \\
            FedACG~\cite{kim2024fedacg} $\lambda$ & \multicolumn{4}{c}{0.2} \\
            FedACG~\cite{kim2024fedacg} $\beta$ & 0.1 & 0.1 & \multicolumn{2}{c}{0.1 (Reptile: 0.0001)}\\
            meta-NSGD~\cite{zhou2022metansgd} $\epsilon$ & \multicolumn{4}{c}{10.0} \\
            meta-NSGD~\cite{zhou2022metansgd} $\lambda$ & \multicolumn{4}{c}{0.1} \\
            
            Number of outer steps ($E$) & 32 & 64 & 8 & 40 \\
            Number of inner steps ($K$) & 1 & 1 & 8 & 8 \\
            Outer learning rate ($\lambda_{o}$) & 0.01 (FOMAML: 0.05) & 0.05 & \multicolumn{2}{c}{0.05 (Reptile:10)} \\
            Inner learning rate ($\lambda_{i}$) & 0.005 & 0.01 & 0.001 & 0.0001 \\

            Optimizer & SGD & SGD & AdamW & AdamW \\
            Test-time optimization (TTO) steps & 64 & 8192 & 8192 & 32768 \\

            Batch size & 1024 & 2048 & \multicolumn{2}{c}{128 rays $\times$ 128 points} \\
            \bottomrule
        \end{tabular}
    }
    \caption{
        Hyperparameters.
    }
    \label{tab:hyperparameters}
\end{table*}

\section{Additional Metrics}
\label{sec:additional_metrics}
We used not only PSNR but also SSIM and LPIPS as metrics to evaluate the performance of reconstruction or novel view synthesis for the trained neural field.
Similarly, we extended the privacy metric, $\text{PSNR}_p$, defined in \cref{sec:privacy_metric}, to $\text{SSIM}_p$ and $\text{LPIPS}_p$. These metrics assess how much information about the client's private dataset is included in the global meta-learner. The privacy metric measures the similarity between data inferred by the meta-learner and the client's local data. The closer the similarity, the greater the privacy leakage, as it indicates the meta-learner has retained more specific information about the client's private data. 
Expanded results corresponding to \cref{tab:main} are presented in \cref{tab:main_petface,tab:main_golfdb,tab:main_cars,tab:main_facescape}.

\section{Convergence}

\begin{figure*}[ht]
    \centering
    \captionsetup[subfigure]{font=scriptsize}

    \hfill
    \begin{subfigure}[b]{0.33\textwidth}
        \centering
        \includegraphics[width=\linewidth]{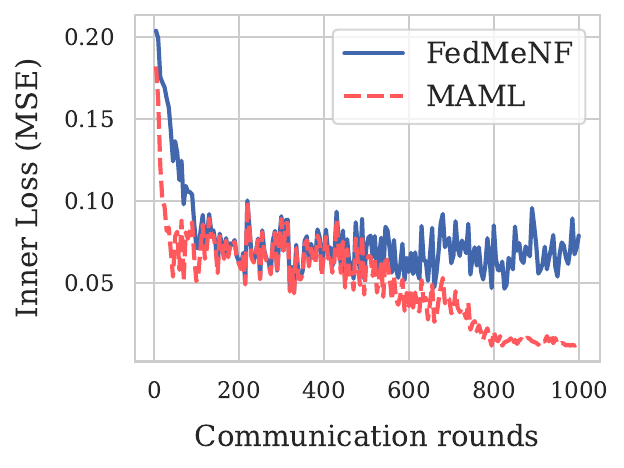}
        \subcaption{Inner loss}
    \end{subfigure}
    \hfill
    \begin{subfigure}[b]{0.33\textwidth}
        \centering
        \includegraphics[width=\linewidth]{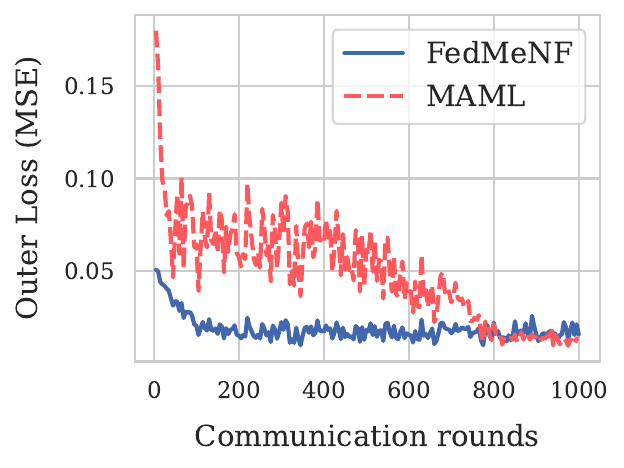}
        \subcaption{Outer loss}
    \end{subfigure}
    \hfill
    \hfill

    \vfill
    
    \begin{subfigure}[b]{0.33\textwidth}
        \centering
        \includegraphics[width=\linewidth]{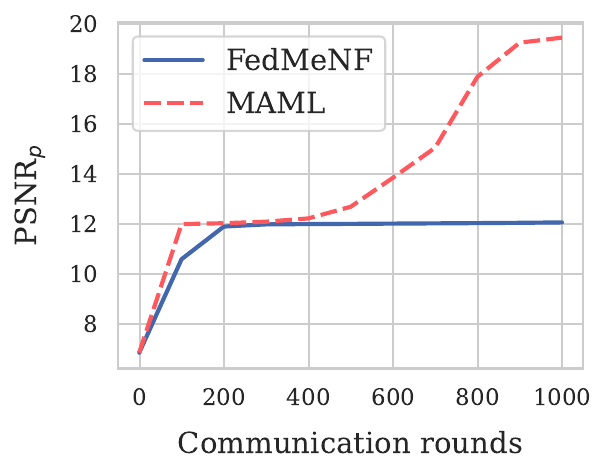}
        \subcaption{$\text{PSNR}_p$}
    \end{subfigure}
    \hfill
    \begin{subfigure}[b]{0.33\textwidth}
        \centering
        \includegraphics[width=\linewidth]{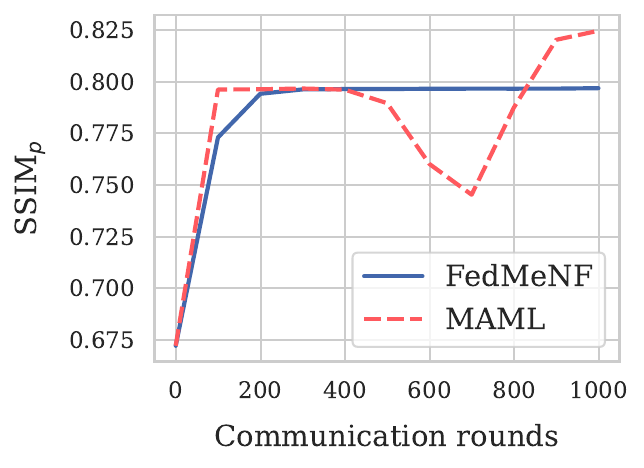}
        \subcaption{$\text{SSIM}_p$}
    \end{subfigure}
    \hfill
    \begin{subfigure}[b]{0.33\textwidth}
        \centering
        \includegraphics[width=\linewidth]{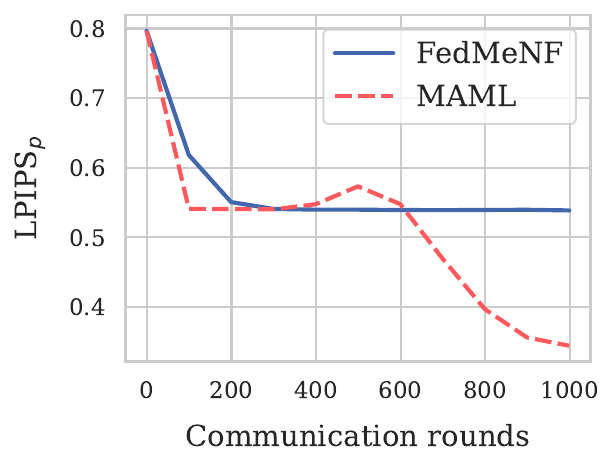}
        \subcaption{$\text{LPIPS}_p$}
    \end{subfigure}

    \vfill

    \begin{subfigure}[b]{0.33\textwidth}
        \centering
        \includegraphics[width=\linewidth]{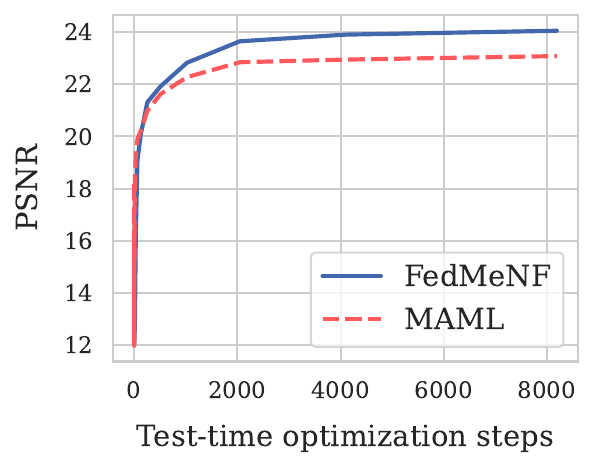}
        \subcaption{PSNR}
    \end{subfigure}
    \hfill
    \begin{subfigure}[b]{0.33\textwidth}
        \centering
        \includegraphics[width=\linewidth]{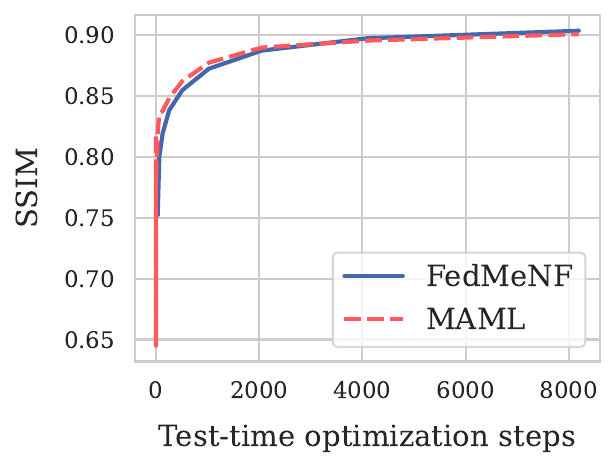}
        \subcaption{SSIM}
    \end{subfigure}
    \hfill
    \begin{subfigure}[b]{0.33\textwidth}
        \centering
        \includegraphics[width=\linewidth]{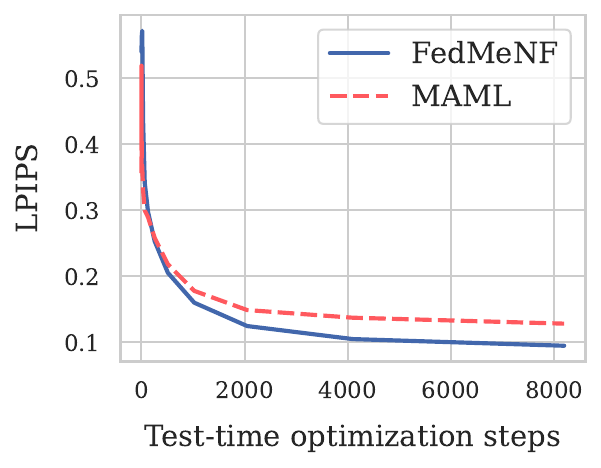}
        \subcaption{LPIPS}
    \end{subfigure}
    \caption{
        Convergence of MAML~\cite{finn2017maml} and our FedMeNF ($\gamma = 0.75$) with FedAvg~\cite{fedavg} on the Cars dataset~\cite{chang2015shapenet}. 
    }
    \label{fig:convergence}
\end{figure*}

\Cref{fig:convergence} shows the convergence behavior of MAML~\cite{finn2017maml} and our FedMeNF ($\gamma = 0.75$) with FedAvg~\cite{fedavg} on the Cars~\cite{chang2015shapenet} dataset. The inner loss represents the MSE loss during the inner loop, as defined in \cref{eq:inner_step}, while the outer loss corresponds to the MSE loss during the outer loop, as described in \cref{eq:outer_step}, for each round.

For the inner loss, MAML converges to a lower value compared to FedMeNF. This happens because the global meta-learner in MAML, used as the initialization for neural field parameters, is closer to the client's data. 

On the other hand, the outer loss of FedMeNF is generally lower than that of MAML. This is because, according to \cref{eq:pp_loss,eq:outer_step}, FedMeNF's loss is calculated as $$L(\varphi_K, B_K) - \gamma L(w_i, B_K),$$ which subtracts the global meta-learner's loss on the client's query set $L(w_i, B_K)$ from MAML's loss $L(\varphi_K, B_K)$. However, the outer loss of FedMeNF converges to a similar level to that of MAML.

One might incorrectly interpret this convergence as indicating that $L(w_i, B_K)$ approaches zero. However, this is not the case. The privacy metric $\text{PSNR}_p$ for FedMeNF converges to a non-zero value, meaning that $L(w_i, B_K)$ also converges to a non-zero value. Instead, the lower outer loss in FedMeNF implies that $L(\varphi_K, B_K)$ is smaller in FedMeNF compared to MAML. This result demonstrates that FedMeNF achieves a lower loss than MAML while preserving client data privacy. This also explains why FedMeNF's novel view synthesis metrics, such as PSNR, SSIM, and LPIPS, not only outperform MAML but also converge faster.

\section{Proof of Proposition 1}
\label{sec:proof_prop_1}

Consider the local meta-optimization gradient in \cref{eq:outer_step} where $\gamma = 0$:

\begin{align}
    g_{M} = \nabla_{w_i}L(\varphi_K, B_K).
\end{align}

\noindent Then, the first-order approximation of the meta-gradient $g_M$ is

\begin{align}
    g_{M} \approx \ g_K - \lambda_{i} \mathcal{I}_{K}, \ 
    \text{where} \ \mathcal{I}_{K} = \sum_{j=0}^{K-1} \nabla_{\varphi_0}\innerproduct{g_K}{g_j}.
\end{align}

\begin{proof}
We use the following definitions:
\begin{align}
    g_i = \nabla_{\varphi_0} L(\varphi_0, B_i) \\
    \label{eq:prop1:def_g_k}
    g_K = \nabla_{\varphi_0} L(\varphi_0, B_K) \\
    H_i = \nabla_{\varphi_0}^2 L(\varphi_0, B_i) \\
    \hat{g}_i = \nabla_{\varphi_i} L(\varphi_i, B_i) \\
    \hat{g}_K = \nabla_{\varphi_K} L(\varphi_K, B_K) \\
    \hat{H}_i = \nabla_{\varphi_i}^2 L(\varphi_i, B_i) 
\end{align}

According to \cref{eq:inner_step},
\begin{align}
    \varphi_{j+1} = \varphi_j - \lambda_{i} \nabla_{\varphi_j} L(\varphi_j, B_j).
\end{align}

Then, we have
\begin{align}
    \varphi_{j} = \varphi_0 - \lambda_{i} \sum_{l=0}^{j-1} \nabla_{\varphi_l} L(\varphi_l, B_l),
\end{align}
\begin{align}
    \varphi_{j} - \varphi_0  &= - \lambda_{i} \sum_{l=0}^{j-1} \nabla_{\varphi_l} L(\varphi_l, B_l) \\
    \label{eq:prop1:diff}
    & = - \lambda_{i} \sum_{l=0}^{j-1} \hat{g}_l.
\end{align}

We assume $L(\varphi_j, B_j)$ is differentiable three times at $\varphi_0$ to apply Taylor theorem.
By Taylor's theorem, we have
\begin{align}
    \hat{g}_j =& \nabla_{\varphi_j} L(\varphi_j, B_j) \\
    =& \nabla_{\varphi_0} L(\varphi_0, B_j) \nonumber \\ 
    &+ \nabla_{\varphi_0}^2 L(\varphi_0, B_j) \cdot (\varphi_j - \varphi_0) \nonumber \\
    &+ \frac{1}{2!}\nabla_{\varphi_0}^3 L(\varphi_0, B_j) \cdot (\varphi_j - \varphi_0)^2 \nonumber \\
    &+ \cdots.
    \label{eq:prop1:taylor}
\end{align}

Combining \cref{eq:prop1:taylor} and \cref{eq:prop1:diff},
\begin{align}
    \hat{g}_j =& \nabla_{\varphi_0} L(\varphi_0, B_j) \nonumber\\
    &+ \nabla_{\varphi_0}^2 L(\varphi_0, B_j) \cdot (\varphi_j - \varphi_0) \nonumber\\
    &+ O(\lambda_{i}^2) \\
    =& g_j - \lambda_{i} H_j \sum_{l=0}^{j-1} \hat{g}_l + O(\lambda_{i}^2).
\end{align}

We assume a learning rate  $\lambda_i$ is small enough for first-order approximation.
Then, the first-order approximation of the $ \hat{g}_j$ is
\begin{equation}
    \hat{g}_j \approx g_j - \lambda_{i} H_j \sum_{l=0}^{j-1} \hat{g}_l.
    \label{eq:prop1:g_first_order_hat}
\end{equation}

Note that 
\begin{equation}
    \hat{g}_j \approx g_j + O(\lambda_{i}).
    \label{eq:prop1:g_hat_g}
\end{equation}

Similarly, we can have
\begin{equation}
    \hat{H}_j \approx H_j + O(\lambda_{i}).
    \label{eq:prop1:h_hat_h}
\end{equation}

Combining \cref{eq:prop1:g_first_order_hat} and \cref{eq:prop1:g_hat_g}, we have
\begin{align}
    \hat{g}_j \approx& g_j - \lambda_{i} H_j \sum_{l=0}^{j-1} g_l + O(\lambda_{i}^2) \\
    \approx& g_j - \lambda_{i} H_j \sum_{l=0}^{j-1} g_l.
    \label{eq:prop1:g_first_order}
\end{align}

Now, let's calculate the meta-gradient $g_{M}$ as follows.
\begin{align}
    g_{M} =& \nabla_{w_i}L(\varphi_K, B_K) \\
    =& \nabla_{\varphi_0}L(\varphi_K, B_K) \quad \quad (\varphi_0 = w_i, \text{ ref. \cref{alg:algo})} \\
    =& \prod_{i=0}^{K-1} \nabla_{\varphi_i} \varphi_{i+1} \cdot \nabla_{\varphi_K} L(\varphi_K, B_K) \\
    =& \prod_{i=0}^{K-1} \nabla_{\varphi_i} (\varphi_{i} - \lambda_{i} \nabla_{\varphi_i} L(\varphi_i, B_i)) \cdot \nabla_{\varphi_K} L(\varphi_K, B_K) \\
    =& \prod_{i=0}^{K-1} (I - \lambda_{i} \nabla_{\varphi_i}^2 L(\varphi_i, B_i)) \cdot \nabla_{\varphi_K} L(\varphi_K, B_K) \\
    =& \prod_{i=0}^{K-1} (I - \lambda_{i} \hat{H}_i) \cdot \hat{g}_K \\
    =& (I - \lambda_{i} \sum_{i=0}^{K-1} \hat{H}_i) \cdot \hat{g}_K + O(\lambda_{i}^2) \\
    =& (I - \lambda_{i} \sum_{i=0}^{K-1} H_i) \cdot \hat{g}_K + O(\lambda_{i}^2) \quad \quad \text{(using \cref{eq:prop1:h_hat_h})} \\
    =& (I - \lambda_{i} \sum_{i=0}^{K-1} H_i) \cdot (g_K - \lambda_{i} H_K \sum_{i=0}^{K-1} g_i) \nonumber \\
    &+ O(\lambda_{i}^2) \quad \quad \text{(using \cref{eq:prop1:g_first_order})} \\
    =& g_K - \lambda_{i} \sum_{i=0}^{K-1} (H_K g_i + g_K H_i) + O(\lambda_{i}^2) \\
    =& g_K - \lambda_{i} \sum_{i=0}^{K-1} \nabla_{\varphi_0} \innerproduct{g_K}{g_i} + O(\lambda_{i}^2)
    \label{eq:prop1:g_m1}
\end{align}

Therefore, the first-order approximation of the $g_{M}$ is
\begin{align}
    g_{M} \approx& g_K - \lambda_{i} \sum_{i=0}^{K-1} \nabla_{\varphi_0} \innerproduct{g_K}{g_i}, \\
    =& g_K - \lambda_{i} \mathcal{I}_{K}.
    \label{eq:prop1:g_m_first_order}
\end{align}

\end{proof}

\section{Proof of Proposition 2}
\label{sec:proof_prop_2}

Let
    $\Delta L_{i+1} = L(w_{i+1}, B_K) - L(w_i, B_K)$.
Then, the first-order approximation of $\Delta L_{i+1}$ is 
\begin{align}
    \label{eq:privacy_metric}
    \Delta L_{i+1} \approx -\lambda_{o} (\nabla_{w_i} L(w_i, B_K))^2 = - \lambda_{o} \cdot {g_K}^2 \leq 0.
\end{align}

\begin{proof}
According to \cref{eq:outer_step} where $\gamma = 0$,
\begin{align}
    w_{i+1} = w_{i} - \lambda_{o} \nabla_{w_i} L(\varphi_K, B_K), \\
    w_{i+1} - w_{i} = - \lambda_{o} \nabla_{w_i} L(\varphi_K, B_K).
    \label{eq:prop2:w_diff}
\end{align}

By Taylor's theorem, we have
\begin{align}
    \label{eq:prop2:l_taylor_raw}
    L(w_{i+1}, B_K) =& L(w_{i}, B_K) \nonumber \\
    &+ L'(w_{i}, B_K) \cdot (w_{i+1}-w_i) \nonumber \\
    &+ \frac{1}{2!} L^{''}(w_{i}, B_K) \cdot (w_{i+1}-w_i)^2 \nonumber \\
    &+ \cdots. \\
    \label{eq:prop2:l_taylor}
    =& L(w_{i}, B_K) \nonumber \\
    &+ L'(w_{i}, B_K) \cdot (- \lambda_{o} \nabla_{w_i} L(\varphi_K, B_K)) \nonumber \\
    &+ O(\lambda_{o}^2) \quad \quad \text{(using \cref{eq:prop2:w_diff}}.
\end{align}

Since $\varphi_0 = w_i$ according to \cref{alg:algo}, 
\begin{align}
     L'(w_{i}, B_K) =  L'(\varphi_0, B_K).
\end{align}

Then, using \cref{eq:prop1:def_g_k},
\begin{align}
    L'(w_{i}, B_K) = g_K.
    \label{eq:prop2:l_to_g_k}
\end{align}

Combining \cref{eq:prop2:l_taylor}, \cref{eq:prop2:l_to_g_k}, and \cref{prop:g_maml}
\begin{align}
    \Delta L_{i+1} =& L(w_{i+1}, B_K) - L(w_{i}, B_K) \\
    =& g_K \cdot (- \lambda_{o} \nabla_{w_i} L(\varphi_K, B_K)) + O(\lambda_{o}^2) \\
    \approx& - \lambda_{o} \cdot g_K \cdot (g_K - \lambda_{i} \mathcal{I}_{K}) + O(\lambda_{o}^2) \\
    =& - \lambda_{o} \cdot {g_K}^2 + O(\lambda_{i} \lambda_{o}) + O(\lambda_{o}^2).
\end{align}

Therefore, the first-order approximation of the $\Delta L_{i+1}$ is
\begin{align}
    \Delta L_{i+1} \approx& - \lambda_{o} \cdot {g_K}^2 \leq 0.
\end{align}

\end{proof}

\section{Proof of Proposition 3}
\label{sec:proof_prop_3}

Consider the gradient of privacy-preserving loss $L_{pp}$:
\begin{align}
    g_{pp} &= \nabla_{w_i} L_{pp}(\gamma, w_i, \varphi_K, B_K) \\
    &= \nabla_{w_i} (L(\varphi_K, B_K) - \gamma L(w_i, B_K)).
\end{align}

\noindent Then, the first-order approximation of $g_{pp}$ and $\Delta L_{i+1}$ are
\begin{align}
    g_{pp} &\approx (1-\gamma) \cdot g_K - \lambda_{i} \mathcal{I}_{K}, \\
    \Delta L_{i+1} &\approx - \lambda_{o} (1 - \gamma) (g_K)^2 \leq 0.
\end{align}

\begin{proof}
Using \cref{prop:g_maml} and \cref{eq:prop1:def_g_k}, we have
\begin{align}
    g_{pp} =& \nabla_{w_i} (L(\varphi_K, B_K) - \gamma L(w_i, B_K)) \\
    \approx& g_M - \gamma g_K \\
    =& (1 - \gamma) \cdot g_K - \lambda_{i} \mathcal{I}_K.
    \label{eq:prop3:g_pp}
\end{align}

According to \cref{eq:outer_step} and using \cref{eq:prop3:g_pp},
\begin{align}
    w_{i+1} = w_{i} - \lambda_{o} \nabla_{w_i} (L(\varphi_K, B_K) - \gamma L(w_i, B_K)),
\end{align}
\begin{align}
    w_{i+1} - w_{i} =& - \lambda_{o} \nabla_{w_i} (L(\varphi_K, B_K) - \gamma L(w_i, B_K)) \\
   \approx& - \lambda_{o} ((1 - \gamma) \cdot g_K - \lambda_{i} \mathcal{I}_K).
    \label{eq:prop3:w_diff}
\end{align}

Combining \cref{eq:prop2:l_taylor_raw} and \cref{eq:prop3:w_diff}, we have
\begin{align}
    L(w_{i+1}, B_K) =& L(w_{i}, B_K) \nonumber \\
    &+ L'(w_{i}, B_K) \cdot (w_{i+1} - w_i) \nonumber \\
    &+ O(\lambda_{o}^2) \\
    \approx& L(w_{i}, B_K) \nonumber \\
    &+ L'(w_{i}, B_K) \cdot ( - \lambda_{o} ((1 - \gamma) \cdot g_K - \lambda_{i} \mathcal{I}_K)) \nonumber \\
    &+ O(\lambda_{o}^2).
    \label{eq:prop3:l_taylor}
\end{align}

Combining \cref{eq:prop3:l_taylor}, \cref{eq:prop2:l_to_g_k}, and \cref{prop:g_maml}
\begin{align}
    \Delta L_{i+1} =& L(w_{i+1}, B_K) - L(w_{i}, B_K) \\
    \approx& g_K \cdot ( - \lambda_{o} ((1 - \gamma) \cdot g_K - \lambda_{i} \mathcal{I}_K)) + O(\lambda_{o}^2) \\
    =& - \lambda_{o} (1 - \gamma)\cdot {g_K}^2 + O(\lambda_{i} \lambda_{o}) + O(\lambda_{o}^2).
\end{align}

Therefore, the first-order approximation of the $\Delta L_{i+1}$ is
\begin{align}
    \Delta L_{i+1} \approx& - \lambda_{o} (1 - \gamma) {g_K}^2 \leq 0.
\end{align}

\end{proof}

\begin{table*}[t]
    \setlength{\tabcolsep}{1.5pt}
    \centering
    \small

        \begin{tabular}{llcccccc}
            \toprule
            
            \multicolumn{2}{c}{\bf{Modality}} & \multicolumn{6}{c}{Image} \\
            
            \multicolumn{2}{c}{\bf{Dataset}} & \multicolumn{6}{c}{PetFace~\cite{shinoda2025petface}} \\
            
            \cmidrule(lr){3-4} \cmidrule(lr){5-6} \cmidrule(lr){7-8}
            
            \multicolumn{2}{c}{\bf{Method \textbackslash \space Metric}} 
            & \multicolumn{1}{c}{$\text{PSNR}_p$(↓)} & \multicolumn{1}{c}{PSNR(↑)}
            & \multicolumn{1}{c}{$\text{SSIM}_p$(↓)} & \multicolumn{1}{c}{SSIM(↑)}
            & \multicolumn{1}{c}{$\text{LPIPS}_p$(↑)} & \multicolumn{1}{c}{LPIPS(↓)}
            \\
            
            \midrule
            \textit{Local} &  & - & 22.29 & - & 0.572 & - & 0.716 \\
            \midrule[0.25pt]
            FedAvg~\cite{fedavg}
            & + MAML          & 16.57   & 27.39 & 0.544   & 0.734  & 0.534   &0.419 \\
            & + FOMAML        & 18.52   & 23.15 & 0.573   & 0.618  & 0.421   &0.415 \\
            & + Reptile       & 17.39   & 22.52 & 0.505   & 0.585  & 0.731   &0.614 \\
            & + meta-NSGD     & 12.49   & 5.15  & 0.395   & 0.003  & 0.923   &1.566 \\
            & + \textbf{Ours} & 14.77   & 27    & 0.526   & 0.723  & 0.606   &0.441 \\
            \midrule[0.25pt]
            FedProx~\cite{fedprox}
            & + MAML          & 16.58   & 27.49 & 0.545   & 0.737  & 0.532   &0.415 \\
            & + FOMAML        & 18.53   & 24.32 & 0.578   & 0.647  & 0.423   &0.4 \\
            & + Reptile       & 17.37   & 22.5  & 0.503   & 0.582  & 0.733   &0.619 \\
            & + meta-NSGD     & 12.49   & 5.15  & 0.395   & 0.003  & 0.923   &1.57 \\
            & + \textbf{Ours} & 14.44   & 27.08 & 0.517   & 0.727  & 0.613   &0.438 \\
            \midrule[0.25pt]
            Scaffold~\cite{scaffold}
            & + MAML          & 16.71   & 27.66 & 0.547   & 0.743  & 0.53    &0.408 \\
            & + FOMAML        & 18.56   & 23.98 & 0.578   & 0.641  & 0.421   &0.401 \\
            & + Reptile       & 17.37   & 22.47 & 0.503   & 0.582  & 0.736   &0.619 \\
            & + meta-NSGD     & 12.49   & 5.13  & 0.395   & 0.003  & 0.923   &1.571 \\
            & + \textbf{Ours} & 14.93   & 27.19 & 0.512   & 0.729  & 0.592   &0.434 \\
            \midrule[0.25pt]
            FedNova~\cite{wang2020fednova}
            & + MAML          & 16.52   & 27.51 & 0.54    & 0.737  & 0.548   &0.414 \\
            & + FOMAML        & 18.5    & 24.1  & 0.571   & 0.635  & 0.432   &0.434 \\
            & + Reptile       & 17.38   & 22.52 & 0.505   & 0.585  & 0.731   &0.615 \\
            & + meta-NSGD     & 12.49   & 5.14  & 0.395   & 0.003  & 0.923   &1.57 \\
            & + \textbf{Ours} & 14.94   & 27.15 & 0.526   & 0.728  & 0.582   &0.433 \\
            \midrule[0.25pt]
            FedExP~\cite{fedexp}
            & + MAML          & 16.57   & 27.39 & 0.544   & 0.734  & 0.534   &0.419 \\
            & + FOMAML        & 18.52   & 23.15 & 0.573   & 0.618  & 0.421   &0.415 \\
            & + Reptile       & 17.39   & 22.64 & 0.505   & 0.586  & 0.731   &0.614 \\
            & + meta-NSGD     & 12.49   & 5.15  & 0.395   & 0.003  & 0.923   &1.566 \\
            & + \textbf{Ours} & 14.65   & 26.86 & 0.504   & 0.72   & 0.619   &0.449 \\
            \midrule[0.25pt]
            FedACG~\cite{kim2024fedacg}
            & + MAML          & 16.63   & 27.5  & 0.547   & 0.738  & 0.526   &0.415 \\
            & + FOMAML        & 18.48   & 24.04 & 0.572   & 0.638  & 0.43    &0.416 \\
            & + Reptile       & 17.38   & 22.63 & 0.505   & 0.587  & 0.727   &0.612 \\
            & + meta-NSGD     & 12.49   & 5.15  & 0.395   & 0.003  & 0.923   &1.571 \\
            & + \textbf{Ours} & 14.71   & 26.97 & 0.524   & 0.723  & 0.606   &0.441 \\
            
            \bottomrule
        \end{tabular}
    \caption{
        Results of various reconstruction quality metrics (PSNR, SSIM, LPIPS) and privacy metrics ($\text{PSNR}_p$, $\text{SSIM}_p$, $\text{LPIPS}_p$) on the PetFace dataset~\cite{shinoda2025petface}.
    }
    \label{tab:main_petface}
\end{table*}

\begin{table*}[t]
    \setlength{\tabcolsep}{1.5pt}
    \centering
    \small

        \begin{tabular}{llcccccc}
            \toprule
            
            \multicolumn{2}{c}{\bf{Modality}} & \multicolumn{6}{c}{Video} \\
            
            \multicolumn{2}{c}{\bf{Dataset}} & \multicolumn{6}{c}{GolfDB~\cite{mcnally2019golfdb}} \\
            
            \cmidrule(lr){3-4} \cmidrule(lr){5-6} \cmidrule(lr){7-8}
            
            \multicolumn{2}{c}{\bf{Method \textbackslash \space Metric}} 
            & \multicolumn{1}{c}{$\text{PSNR}_p$(↓)} & \multicolumn{1}{c}{PSNR(↑)}
            & \multicolumn{1}{c}{$\text{SSIM}_p$(↓)} & \multicolumn{1}{c}{SSIM(↑)}
            & \multicolumn{1}{c}{$\text{LPIPS}_p$(↑)} & \multicolumn{1}{c}{LPIPS(↓)}
            \\
            
            \midrule
            \textit{Local} &  & - & 26.92 & - & 0.796 & - & 0.27 \\
            \midrule[0.25pt]
            FedAvg~\cite{fedavg}
            & + MAML           & 21.21 & 29.68    &   0.657    &  0.867 &  0.385    &0.131     \\
            & + FOMAML         & 20.82 & 28.57    &   0.634    &  0.843 &  0.445    &0.166     \\
            & + Reptile        & 19.89 & 27.22    &   0.562    &  0.811 &  0.66     &0.245     \\
            & + meta-NSGD      & 10.96 & 4.85     &   0.42     &  0.003 &  0.936    &1.485     \\
            & + \textbf{Ours}  & 17.31 & 28.89    &   0.579    &  0.855 &  0.53     &0.167     \\
            \midrule[0.25pt]
            FedProx~\cite{fedprox}
            & + MAML           & 21    & 29.35    &   0.652    &  0.862 &  0.406    &0.146     \\
            & + FOMAML         & 21.28 & 28.79    &   0.642    &  0.85  &  0.434    &0.154     \\
            & + Reptile        & 19.79 & 27.21    &   0.559    &  0.811 &  0.665    &0.242     \\
            & + meta-NSGD      & 10.96 & 4.84     &   0.42     &  0.003 &  0.936    &1.486    \\
            & + \textbf{Ours}  & 15.68 & 28.96    &   0.546    &  0.856 &  0.543    &0.164     \\
            \midrule[0.25pt]
            Scaffold~\cite{scaffold}
            & + MAML           & 21.23 & 29.11    &   0.658    &  0.857 &  0.403    &0.157     \\
            & + FOMAML         & 21.27 & 28.78    &   0.648    &  0.849 &  0.427    &0.161     \\
            & + Reptile        & 20    & 27.23    &   0.568    &  0.812 &  0.651    &0.24      \\
            & + meta-NSGD      & 10.96 & 4.84     &   0.42     &  0.003 &  0.936    &1.486     \\
            & + \textbf{Ours}  & 16.21 & 29.23    &   0.56     &  0.861 &  0.528    &0.149     \\
            \midrule[0.25pt]
            FedNova~\cite{wang2020fednova}
            & + MAML           & 21.56 & 29.63    &   0.669    &  0.868 &  0.378    &0.127     \\
            & + FOMAML         & 20.12 & 28.52    &   0.616    &  0.842 &  0.496    &0.156     \\
            & + Reptile        & 19.92 & 27.27    &   0.563    &  0.814 &  0.662    &0.235     \\
            & + meta-NSGD      & 10.96 & 4.84     &   0.42     &  0.003 &  0.936    &1.485    \\
            & + \textbf{Ours}  & 15.71 & 28.98    &   0.549    &  0.856 &  0.517    &0.161     \\
            \midrule[0.25pt]
            FedExP~\cite{fedexp}
            & + MAML           & 21.21 & 29.68    &   0.657    &  0.867 &  0.385    &0.131     \\
            & + FOMAML         & 20.82 & 28.57    &   0.634    &  0.843 &  0.445    &0.166     \\
            & + Reptile        & 19.89 & 27.22    &   0.562    &  0.811 &  0.66     &0.245     \\
            & + meta-NSGD      & 10.96 & 4.85     &   0.42     &  0.003 &  0.936    &1.485     \\
            & + \textbf{Ours}  & 15.38 & 28.1     &   0.53     &  0.834 &  0.54     &0.204     \\
            \midrule[0.25pt]
            FedACG~\cite{kim2024fedacg}
            & + MAML           & 21.09 & 29.51    &   0.657    &  0.866 &  0.398    &0.132     \\
            & + FOMAML         & 21.12 & 28.7     &   0.637    &  0.847 &  0.445    &0.167    \\
            & + Reptile        & 19.83 & 27.23    &   0.561    &  0.812 &  0.658    &0.24      \\
            & + meta-NSGD      & 10.96 & 4.84     &   0.42     &  0.003 &  0.936    &1.485     \\
            & + \textbf{Ours}  & 15.94 & 28.99    &   0.535    &  0.858 &  0.584    &0.165     \\
            
            \bottomrule
        \end{tabular}
    \caption{
        Results of various reconstruction quality metrics (PSNR, SSIM, LPIPS) and privacy metrics ($\text{PSNR}_p$, $\text{SSIM}_p$, $\text{LPIPS}_p$) on the GolfDB dataset~\cite{mcnally2019golfdb}.
    }
    \label{tab:main_golfdb}
\end{table*}

\begin{table*}[t]
    \setlength{\tabcolsep}{1.5pt}
    \centering
    \small

        \begin{tabular}{llcccccc}
            \toprule
            
            \multicolumn{2}{c}{\bf{Modality}} & \multicolumn{6}{c}{3D (NeRF)} \\
            
            \multicolumn{2}{c}{\bf{Dataset}} & \multicolumn{6}{c}{Cars~\cite{chang2015shapenet}} \\
            
            \cmidrule(lr){3-4} \cmidrule(lr){5-6} \cmidrule(lr){7-8}
            
            \multicolumn{2}{c}{\bf{Method \textbackslash \space Metric}} 
            & \multicolumn{1}{c}{$\text{PSNR}_p$(↓)} & \multicolumn{1}{c}{PSNR(↑)}
            & \multicolumn{1}{c}{$\text{SSIM}_p$(↓)} & \multicolumn{1}{c}{SSIM(↑)}
            & \multicolumn{1}{c}{$\text{LPIPS}_p$(↑)} & \multicolumn{1}{c}{LPIPS(↓)}
            \\
            
            \midrule
            \textit{Local} &  & - & 17.13 & - & 0.844 & - & 0.319 \\
            \midrule[0.25pt]
            FedAvg~\cite{fedavg}
            & + MAML          & 19.73    &23.08 & 0.827 &  0.901 &  0.343    & 0.128    \\
            & + FOMAML        & 19.73    &23.66 & 0.827 &  0.905 &  0.343    & 0.112    \\
            & + Reptile       & 19.96    &21.98 & 0.839 &  0.892 &  0.311    & 0.154    \\
            & + meta-NSGD     & 6.85     &10.62 & 0.672 &  0.767 &  0.798    & 0.566    \\
            & + \textbf{Ours} & 12.15    &24.05 & 0.798 &  0.904 &  0.537    & 0.095    \\
            \midrule[0.25pt]
            FedProx~\cite{fedprox}
            & + MAML          & 19.67    &22.94 & 0.826 &  0.9   &  0.346    & 0.129    \\
            & + FOMAML        & 19.67    &23.73 & 0.826 &  0.907 &  0.346    & 0.11     \\
            & + Reptile       & 19.95    &22.11 & 0.839 &  0.891 &  0.311    & 0.155    \\
            & + meta-NSGD     & 6.85     &10.91 & 0.672 &  0.779 &  0.798    & 0.547   \\
            & + \textbf{Ours} & 12.14    &23.98 & 0.798 &  0.902 &  0.537    & 0.096    \\
            \midrule[0.25pt]
            Scaffold~\cite{scaffold}
            & + MAML          & 19.07    &24.21 & 0.811 &  0.911 &  0.372    & 0.092    \\
            & + FOMAML        & 19.07    &24.47 & 0.811 &  0.912 &  0.373    & 0.086    \\
            & + Reptile       & 19.81    &22.6  & 0.936 &  0.896 &  0.322    & 0.137    \\
            & + meta-NSGD     & 6.85     &10.63 & 0.672 &  0.762 &  0.798    & 0.559    \\
            & + \textbf{Ours} & 14.4     &24.34 & 0.798 &  0.91  &  0.514    & 0.086    \\
            \midrule[0.25pt]
            FedNova~\cite{wang2020fednova}
            & + MAML          & 19.72    &23.63 & 0.827 &  0.907 &  0.344    & 0.109    \\
            & + FOMAML        & 19.72    &23.27 & 0.827 &  0.903 &  0.344    & 0.121    \\
            & + Reptile       & 19.96    &22.7  & 0.838 &  0.897 &  0.311    & 0.135    \\
            & + meta-NSGD     & 6.85     &10.6  & 0.672 &  0.763 &  0.798    & 0.561   \\
            & + \textbf{Ours} & 12.14    &24.12 & 0.798 &  0.903 &  0.537    & 0.096    \\
            \midrule[0.25pt]
            FedExP~\cite{fedexp}
            & + MAML          & 19.81    &22.87 & 0.829 &  0.899 &  0.339    & 0.14     \\
            & + FOMAML        & 19.81    &22.81 & 0.829 &  0.899 &  0.339    & 0.141    \\
            & + Reptile       & 20.91    &22.03 & 0.851 &  0.891 &  0.267    & 0.161    \\
            & + meta-NSGD     & 6.85     &10.61 & 0.672 &  0.765 &  0.798    & 0.571    \\
            & + \textbf{Ours} & 12.17    &24.05 & 0.798 &  0.905 &  0.537    & 0.093    \\
            \midrule[0.25pt]
            FedACG~\cite{kim2024fedacg}
            & + MAML          & 19.9     &22    & 0.832 &  0.891 &  0.328    & 0.155    \\
            & + FOMAML        & 19.9     &21.95 & 0.832 &  0.89  &  0.328    & 0.155   \\
            & + Reptile       & 20.13    &22.26 & 0.842 &  0.894 &  0.299    & 0.148    \\
            & + meta-NSGD     & 6.85     &10.59 & 0.672 &  0.761 &  0.798    & 0.57     \\
            & + \textbf{Ours} & 10.93    &22.45 & 0.779 &  0.881 &  0.601    & 0.133    \\

            \bottomrule
        \end{tabular}
    \caption{
        Results of various reconstruction quality metrics (PSNR, SSIM, LPIPS) and privacy metrics ($\text{PSNR}_p$, $\text{SSIM}_p$, $\text{LPIPS}_p$) on the Cars dataset~\cite{chang2015shapenet}.
    }
    \label{tab:main_cars}
\end{table*}

\begin{table*}[t]
    \setlength{\tabcolsep}{1.5pt}
    \centering
    \small

        \begin{tabular}{llcccccc}
            \toprule
            
            \multicolumn{2}{c}{\bf{Modality}} & \multicolumn{6}{c}{3D (NeRF)} \\
            
            \multicolumn{2}{c}{\bf{Dataset}} & \multicolumn{6}{c}{FaceScape~\cite{zhu2023facescape,yang2020facescape}} \\
            
            \cmidrule(lr){3-4} \cmidrule(lr){5-6} \cmidrule(lr){7-8}
            
            \multicolumn{2}{c}{\bf{Method \textbackslash \space Metric}} 
            & \multicolumn{1}{c}{$\text{PSNR}_p$(↓)} & \multicolumn{1}{c}{PSNR(↑)}
            & \multicolumn{1}{c}{$\text{SSIM}_p$(↓)} & \multicolumn{1}{c}{SSIM(↑)}
            & \multicolumn{1}{c}{$\text{LPIPS}_p$(↑)} & \multicolumn{1}{c}{LPIPS(↓)}
            \\
            
            \midrule
            \textit{Local} &  & - & 23.67 & - & 0.772 & - & 0.178 \\
            \midrule[0.25pt]
            FedAvg~\cite{fedavg}
            & + MAML          &  21.31  &28.59   &0.658    &0.904     &0.421    & 0.053 \\
            & + FOMAML        &  21.24  &28.65   &0.658    &0.905     &0.416    & 0.051 \\
            & + Reptile       &  21.92  &28.24   &0.684    &0.895     &0.368    & 0.056 \\
            & + meta-NSGD     &  7.72   &11.29   &0.191    &0.461     &0.934    & 0.742 \\
            & + \textbf{Ours} &  15.16  &27.88   &0.268    &0.894     &0.665    & 0.061 \\
            \midrule[0.25pt]
            FedProx~\cite{fedprox}
            & + MAML          &  21.31  &28.59   &0.658    &0.904     &0.421    & 0.053 \\
            & + FOMAML        &  21.24  &28.65   &0.658    &0.905     &0.416    & 0.051 \\
            & + Reptile       &  21.89  &28.21   &0.683    &0.895     &0.37     & 0.056 \\
            & + meta-NSGD     &  7.72   &11.29   &0.191    &0.461     &0.934    & 0.742\\
            & + \textbf{Ours} &  15.16  &27.88   &0.268    &0.894     &0.665    & 0.061 \\
            \midrule[0.25pt]
            Scaffold~\cite{scaffold}
            & + MAML          &  21.09  &28.51   &0.651    &0.902     &0.44    & 0.053 \\
            & + FOMAML        &  21.01  &28.51   &0.65     &0.903     &0.435    & 0.053 \\
            & + Reptile       &  21.79  &28.14   &0.677    &0.893     &0.384    & 0.058 \\
            & + meta-NSGD     &  7.72   &11.29   &0.191    &0.461     &0.933    & 0.741 \\
            & + \textbf{Ours} &  13.94  &27.53   &0.195    &0.888     &0.748    & 0.065 \\
            \midrule[0.25pt]
            FedNova~\cite{wang2020fednova}
            & + MAML          &  21.37  &28.59   &0.661    &0.903     &0.421    & 0.053 \\
            & + FOMAML        &  21.3   &28.62   &0.66     &0.905     &0.416    & 0.052 \\
            & + Reptile       &  21.98  &28.17   &0.686    &0.894     &0.369    & 0.057 \\
            & + meta-NSGD     &  7.72   &11.29   &0.191    &0.461     &0.933    & 0.743\\
            & + \textbf{Ours} &  15.45  &27.86   &0.282    &0.894     &0.648    & 0.061 \\
            \midrule[0.25pt]
            FedExP~\cite{fedexp}
            & + MAML          &  21.33  &27.64   &0.659    &0.886     &0.419    & 0.065 \\
            & + FOMAML        &  21.26  &27.66   &0.659    &0.887     &0.415    & 0.064 \\
            & + Reptile       &  21.93  &28.27   &0.685    &0.896     &0.367    & 0.058 \\
            & + meta-NSGD     &  7.72   &11.29   &0.191    &0.461     &0.933    & 0.742 \\
            & + \textbf{Ours} &  13.69  &26.12   &0.191    &0.843     &0.766    & 0.103 \\
            \midrule[0.25pt]
            FedACG~\cite{kim2024fedacg}
            & + MAML          &  22.28  &28.02   &0.7      &0.897     &0.327    & 0.055 \\
            & + FOMAML        &  22.24  &28.03   &0.699    &0.9       &0.327    & 0.052\\
            & + Reptile       &  22.07  &28.32   &0.691    &0.896     &0.356    & 0.055 \\
            & + meta-NSGD     &  7.72   &11.29   &0.191    &0.461     &0.934    & 0.742 \\
            & + \textbf{Ours} &  14.83  &27.46   &0.365    &0.876     &0.669    & 0.074 \\
            \bottomrule
        \end{tabular}
    \caption{
        Results of various reconstruction quality metrics (PSNR, SSIM, LPIPS) and privacy metrics ($\text{PSNR}_p$, $\text{SSIM}_p$, $\text{LPIPS}_p$) on the FaceScape dataset~\cite{zhu2023facescape,yang2020facescape}.
    }
    \label{tab:main_facescape}
\end{table*}

\end{document}